\documentclass[11pt]{article} 
\usepackage{acl}

\usepackage{times}
\usepackage{latexsym}

\usepackage[T1]{fontenc}

\usepackage[utf8]{inputenc}

\usepackage{inconsolata}

\usepackage{microtype}

\usepackage{graphicx}

\usepackage{amsmath,amsfonts,bm}

\def\eqref#1{equation~\ref{#1}}

\def\1{\bm{1}}

\DeclareMathAlphabet{\mathsfit}{\encodingdefault}{\sfdefault}{m}{sl}
\SetMathAlphabet{\mathsfit}{bold}{\encodingdefault}{\sfdefault}{bx}{n}

\usepackage{listings}
\usepackage{dsfont}

\definecolor{positional}{HTML}{7777F9}
\definecolor{nonpositional}{HTML}{777777}
\definecolor{lightgray}{HTML}{D3D3D3}

\usepackage{hyperref}
\usepackage{algorithm}
\usepackage{algpseudocode}
\usepackage{url}
\usepackage{graphicx}
\usepackage{booktabs}
\usepackage{colortbl}
\usepackage{enumitem}
\usepackage{linguex}
\usepackage{multirow}

\usepackage[compact]{titlesec}

\title{Position-aware Automatic Circuit Discovery}

\author{
Tal Haklay$^1$  \quad Hadas Orgad$^1$  \quad David Bau$^2$  \quad Aaron Mueller$^{1,2}$  \quad Yonatan Belinkov$^1$  \\
$^1$Technion -- Israel Institute of Technology \quad $^2$Northeastern University \\[1ex]
\{\href{mailto:tal.ha@campus.ac.il}{\texttt{tal.ha}},
\href{mailto:orgad.hadas@campus.technion.ac.il}{\texttt{orgad.hadas}}\}\texttt{@campus.ac.il},
\{\href{mailto:d.bau@northeastern.edu}{\texttt{d.bau}},
\href{mailto:aa.mueller@northeastern.edu}{\texttt{aa.mueller}}\}\texttt{@northeastern.edu} \\
\href{mailto:belinkov@technion.ac.il}{\texttt{belinkov@technion.ac.il}}
}

\begin{document}

\maketitle

\begin{abstract}
A widely used strategy to discover and understand language model mechanisms is circuit analysis. A circuit is a minimal subgraph of a model’s computation graph that executes a specific task. 
We identify a gap in existing circuit discovery methods:
they assume circuits are position-invariant, treating model components as equally relevant across input positions.
This limits their ability to capture cross-positional interactions or mechanisms that vary across positions.
To address this gap, we propose two improvements to incorporate positionality into circuits, even on tasks containing variable-length examples.
First, we extend edge attribution patching, a gradient-based method for circuit discovery, to differentiate between token positions.
Second, we introduce the concept of a dataset schema, which defines token spans with similar semantics across examples, enabling position-aware circuit discovery in datasets with variable length examples.
We additionally develop an automated pipeline for schema generation and application using large language models.
Our approach enables fully automated discovery of position-sensitive circuits, yielding better trade-offs between circuit size and faithfulness compared to prior work.%
\footnote{Our code is available in \url{https://github.com/technion-cs-nlp/PEAP}.}

\end{abstract}
\begin{figure}[ht]
    \centering
    \includegraphics[width=0.8\linewidth]{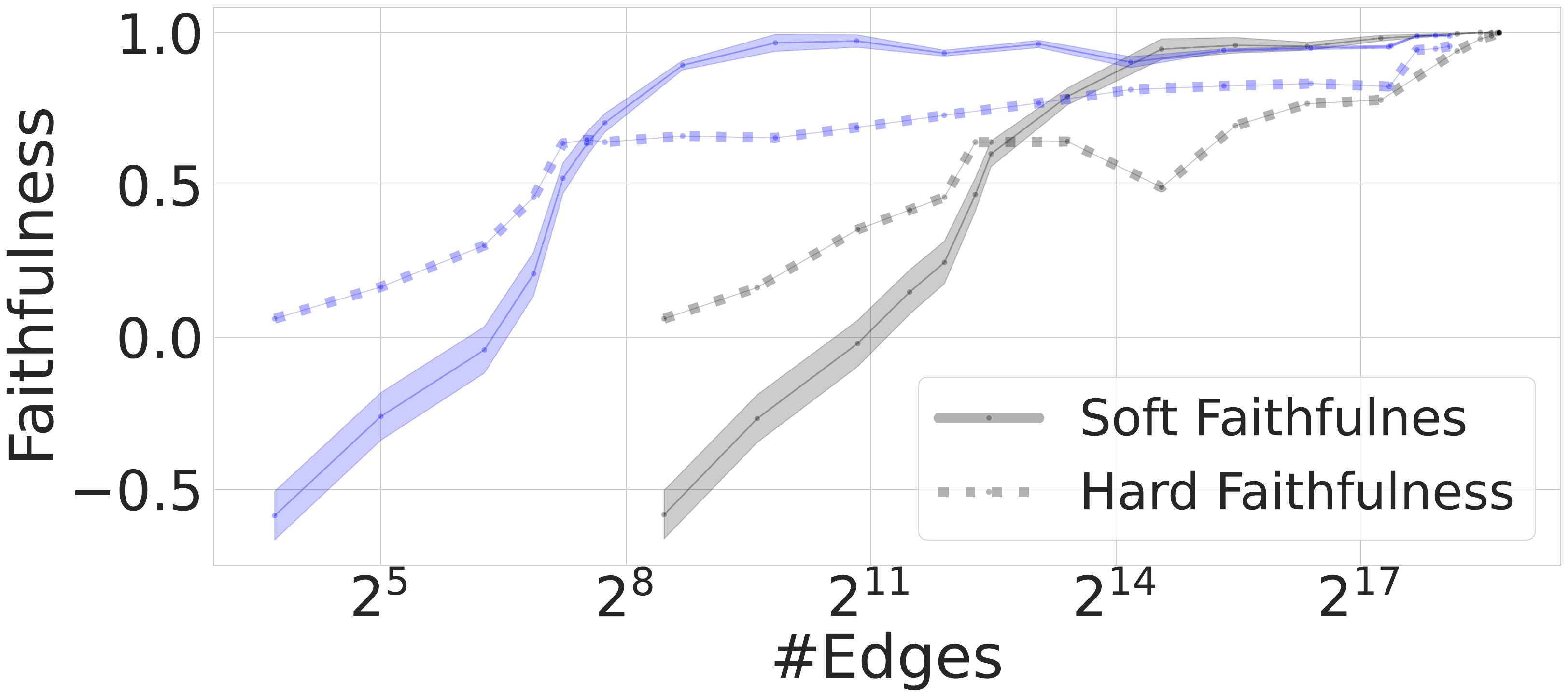}
    \vspace{0.5cm}
    \includegraphics[width=0.8\linewidth]{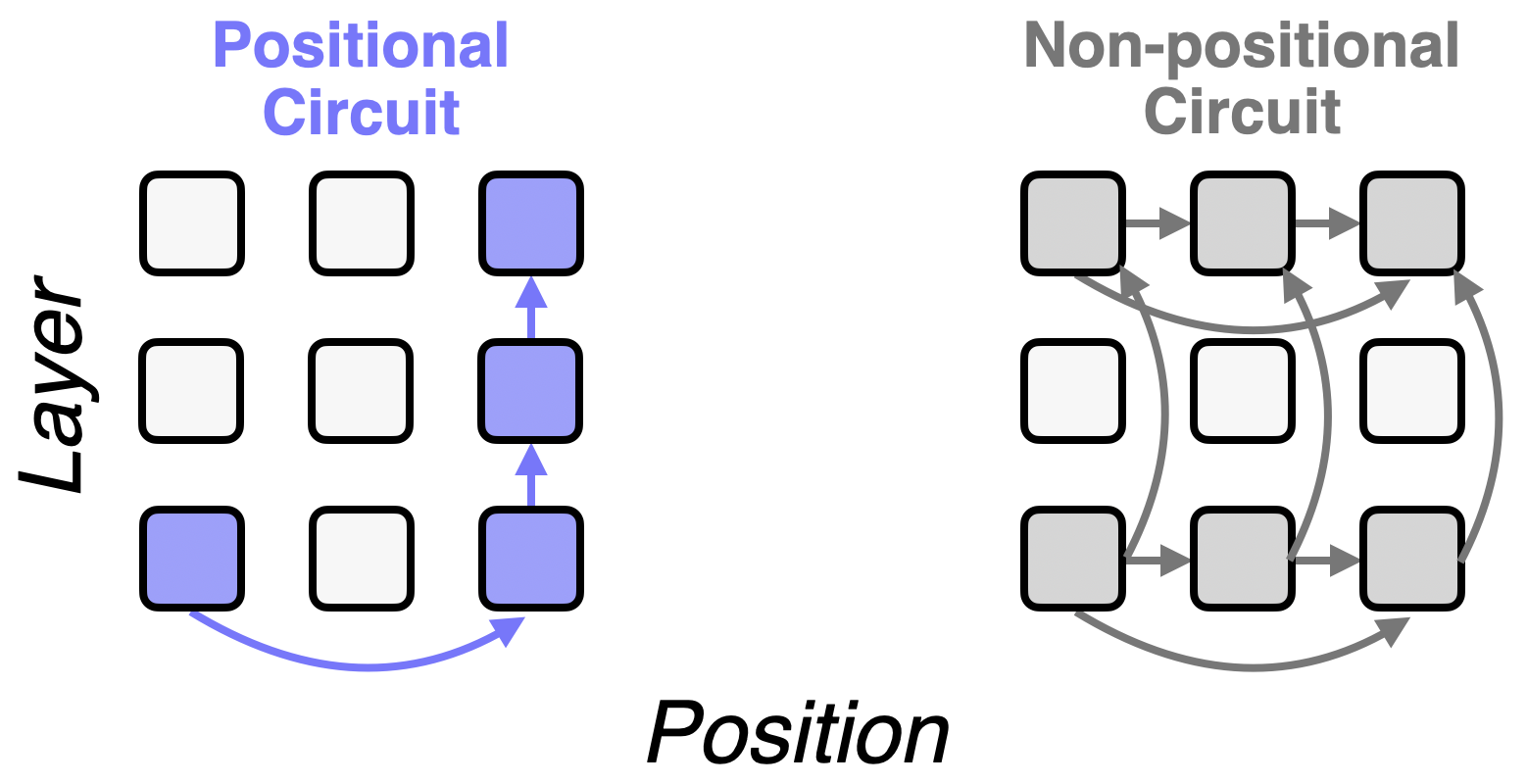}
    \vspace{-5pt}
    \caption{\textcolor{positional}{Positional} vs.\ \textcolor{nonpositional}{non-positional} circuits. In a \textcolor{nonpositional}{non-positional} circuit, the same edges must be included at all positions. A \textcolor{positional}{positional} circuit can distinguish between the same edge at different positions. This specificity yields better trade-offs between circuit size and faithfulness. It can also increase both precision and recall.}
    \label{fig:p1}
    \vspace{-5pt}
\end{figure}

\section{Introduction}

\looseness=-1
A primary goal of interpretability research is to characterize the internal mechanisms in language models (LMs) and other NLP models. 
A core approach in this area is \textbf{circuit discovery}---identifying the minimal subgraph within the model's computation graph that performs a specific task \citep{olah2021framework,olah-mech}.
Typically, the nodes of a circuit represent model components (e.g., attention heads, neurons, or layers).
While manual circuit discovery methods can yield position-specific insights \citep{wanginterpretability,goldowskydill2023localizingmodelbehaviorpath}, \emph{automatic methods often overlook positional information}, treating components as uniformly relevant across all input token positions \citep{conmytowards,syed2023attribution}. 
For instance, if an attention head is included in a circuit, it is assumed to contribute equally to the computation for every position in the input sequence.
The assumption that circuits are position-invariant ignores the fact that different positions often require distinct computations.
By ignoring positions, current methods limit their ability to capture mechanisms that operate across positions, such as interactions between attention heads across positions.

In this study, we start by demonstrating that positional agnosticism is a significant limitation (\S\ref{sec:motivating}). Then, to address these limitations, we introduce a new approach: position-aware edge attribution patching (PEAP; \S\ref{sec:full_circ_discovery}; Figure~\ref{fig:p1}). Current approaches  assume that if an edge is in a circuit, then the same edge will be in the circuit at all positions, thus leading to low precision. It is also assumed that an edge's importance should be aggregated across positions before deciding whether it should be included in the circuit; this can lead to cancellation effects, and thus low recall. PEAP instead allows us to compute the importance of cross-positional edges, and separately evaluates edge importance at each position. We show that this leads to smaller and more accurate circuits; see Figure~\ref{fig:p1}.

Incorporating positional information into circuit discovery is straightforward when inputs have the same length and structure across examples.

However, realistic datasets are not nearly this templatic.
How, then, can we incorporate positional information into automatic circuit discovery?
To address this challenge, we propose \textbf{schemas} (\S\ref{sec:schema}). 
Schemas assign semantic labels to spans of tokens, enabling information aggregation across examples even when the spans differ in length.

For example, in the input ``The \textcolor{positional}{war} lasted from 1453 to 14\underline{\hspace{1em}},'' the span ``\textcolor{positional}{war}'' could be labeled as ``\emph{Subject}''.
This enables handling spans with varying lengths: the phrase ``\textcolor{positional}{Black Plague}'' in another example can be treated as a single positional span with the same role as ``\textcolor{positional}{war}''.
In experiments with two LMs and three tasks, we find that circuits discovered using schemas achieve a better trade-off between circuit size and faithfulness to the model's behavior than position-agnostic circuits.
Importantly, position-aware circuits offer a more precise representation of the underlying mechanisms, providing a more concise foundation for mechanistic explanations.

We also present a fully automated pipeline for schema generation and application (\S\ref{sec:schema-generation}) using large language models (LLMs). 
We evaluate the quality of the generated schemas and their utility in discovering position-aware circuits (\S\ref{sec:schema-eval}).
Notably, circuits derived using automatically generated and applied schemas achieve comparable faithfulness scores to circuits discovered with human-designed and manually applied schemas.

We summarize our contributions as follows:
\begin{itemize}[noitemsep,leftmargin=*,topsep=1pt,parsep=1pt]
    \item Introduce a position-aware circuit discovery method, which obtains better faithfulness than position-agnostic discovery.  
    \item Introduce dataset schemas,  facilitating positional circuit discovery in more naturalistic settings. 
    \item Develop an automated schema generation and application pipeline with LLMs, yielding schemas that are comparable to manually-annotated ones.
\end{itemize}

\section{Background and Motivation }\label{sec:motivating}
\begin{figure*}
    \centering
    \includegraphics[width=0.45\linewidth]{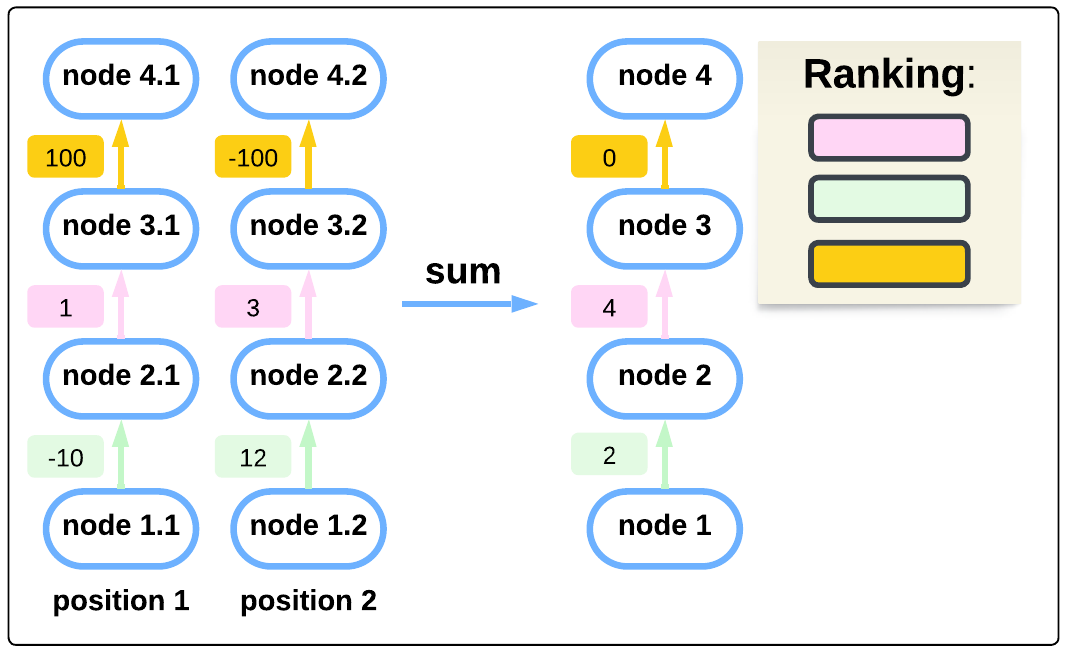}\hspace{1em} 
\includegraphics[width=0.45\linewidth]{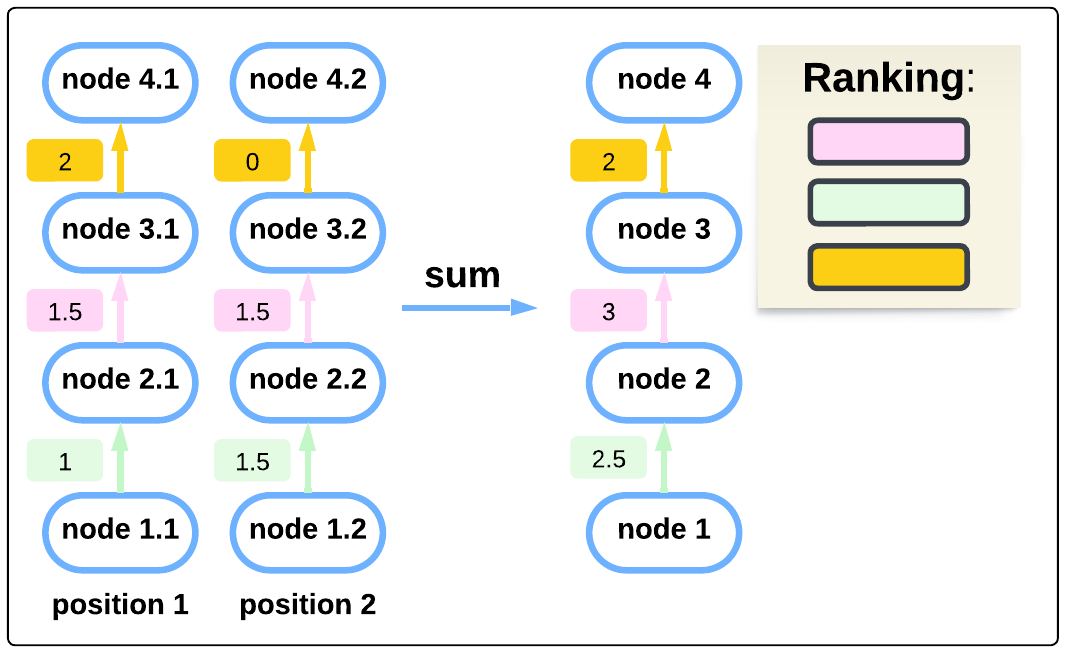}
    \caption{\textbf{Left}: The yellow edge at position 1 has the highest score of 100, indicating it is the most important edge. However, aggregating across positions causes scores of opposite signs to cancel. This causes the yellow edge to be incorrectly ranked as the \emph{least} important. \textbf{Right}: The yellow edge at position 1 has the highest score; the scores of other edges are consistently high (but lower) at many positions. After summing  across positions, the non-yellow edges have higher scores. Thus, the yellow edge is incorrectly ranked as the least important.}
    \label{fig:failure-exp}
\end{figure*}

A circuit is a subgraph of the model's computation graph;
it can be conceptualized as a binary mask $B(V,E,\mathcal{T})$ over all components $V$ and edges $E$ in the graph, selecting the components and edges that have the strongest effect on the model's behavior on a target task $\mathcal{T}$.
There are many methods for computing the influence of a component on the model's behavior on $\mathcal{T}$, including activation patching \citep{vig-2020-gender,finlayson-etal-2021-causal,geiger2021causal}, path patching \citep{wanginterpretability,goldowskydill2023localizingmodelbehaviorpath}, and edge patching \citep{hanna2024have,marks2024sparsefeaturecircuitsdiscovering}, with 
 attribution patching to approximate direct patching \cite{nanda2023attribution,syed2023attribution}.  
We focus  on edge patching, which aims to identify edges in $E$ that are causally important for $\mathcal{T}$.
For each such edge $(u,v)$, the nodes $u$ and $v$ are included in the circuit.

\emph{Manual} circuit discovery methods can distinguish between components at different token positions; examples include the IOI circuit \citep{wanginterpretability}, the Greater-Than circuit \citep{hanna2024does}, and the Attribute-Binding circuit \citep{prakashfine}.
The authors determined connections between attention heads by examining attention patterns and establishing connections if a head at one position strongly attended to a head at another.
However, this approach has three key limitations: (1) it is not scalable, (2) it is prone to human bias, and (3) it is unclear whether strong attention scores reliably indicate the a causal connection to the downstream metric \citep{jain2019attention}.

In contrast, \emph{automatic} approaches \citep{syed2023attribution, hanna2024have} systematically examine every connection and evaluate them \emph{quantitatively} via their causal effect on the downstream metric. However, when using automatic methods it is common to aggregate across token positions,\footnote{Cf.\ \citet{kramar2024atp}, propose a variant of attribution patching and perform position-sensitive node attribution. They do not use it to discover positional circuits.} which causes specific problems that we now define.

\paragraph{Cancellations across positions (low recall).}
If a component has scores with opposite signs across different positions, summing these scores can partially cancel out the component's overall effect, potentially resulting in a near-zero score (Figure~\ref{fig:failure-exp}, left).
\citet{kramar2024atp} note that cancellation can occur when aggregating across examples in the dataset. We observe that the extent of this phenomenon is larger than previously assumed: it can occur \emph{within a single sample} across positions. 
To measure cancellation effects across positions, we compare importance rankings from edge attribution patching (EAP; \citealp{syed2023attribution}) under two positional aggregation methods: (i) summing the absolute scores across both positions and examples (unaffected by cancellation effects); and (ii) summing scores across positions and then summing the absolute scores across different examples (affected by cancellation effects). 
We observe (Table~\ref{tab:intersections}, Top)  that the two rankings differ significantly at the most important components. 

\paragraph{Importance overestimation (low precision).} 
Circuits that do not consider positional information may favor edges that have small impacts at many positions over edges that have large impact in one or few positions (Figure \ref{fig:failure-exp}, Right). To measure overestimation effects we compare importance rankings derived from two aggregation methods: (i) summing the absolute scores across both positions and examples; and (ii) taking the max of the absolute across positions and then summing scores across different examples. Table \ref{tab:intersections} (Bottom) provides evidence for this phenomenon.

These problems motivate a circuit discovery method that takes position into account. We introduce this method in \S\ref{sec:full_circ_discovery}.

\begin{table}[t!]
\vspace{-10pt}
    \centering   
     \begin{tabular}{lcccc}
    \toprule
    \rowcolor{lightgray} \multicolumn{5}{c}{Cancellation} \\
    $K\%$ & Diff & Diff$_{\text{Control}}$ & $\rho$ & $\rho_{\text{Control}}$ \\
    \midrule
    1 & 17.1\% & 3.9\% & 0.760 & 0.985 \\
    5 & 13.4\% & 2.4\% & 0.831 & 0.991 \\
    10 & 12.1\% & 2.3\% & 0.877 & 0.992 \\
    \midrule
    \rowcolor{lightgray} \multicolumn{5}{c}{Overestimation} \\
    $K\%$ & Diff & Diff$_{\text{Control}}$ & $\rho$ & $\rho_{\text{Control}}$ \\
    \midrule
    1 & 17.5\% & 3.6\% & 0.772 & 0.984 \\
    5 & 14.6\% & 2.1\% & 0.811 & 0.993 \\
    10 & 12.4\% & 2.2\% & 0.864 & 0.993 \\
    \bottomrule
    \end{tabular}
   \caption{Cancellation and overestimation effects when ignoring positions. We rank edges by their importance scores (IOI task, GPT2-small), and take the top $K\%$. We compute the set difference (Diff) and rank correlations ($\rho$) between rankings produced by the two aggregation methods described in \S\ref{sec:motivating}. We define the difference of two ranking lists $R_1$,$R_2$ at length L as $1-\frac{|R_1 \bigcap R_2|}{L}$. As a control, we also compute the mean pairwise set difference (Diff$_\text{Control}$) and rank similarities ($\rho_\text{Control}$) produced by the \emph{same} aggregation method across 3 data subsets. Differences with respect to control are all significant ($p<.01$).
   }
    \label{tab:intersections}
    \vspace{-5pt}
\end{table}

\section{Position-aware Edge Attribution Patching (PEAP)} \label{sec:full_circ_discovery}

The importance of an edge $e$ is typically measured with the indirect effect (IE) of the edge on some target metric $M$.
In direct activation patching, also known as causal mediation analysis \cite{Pearl:2001:DIE:2074022.2074073,vig-2020-gender, mueller2024quest}, 
the IE is the change in the metric $M$ when the edge is `patched' to some counterfactual value, e.g., the edge value in a run on a different input $x'$:  $M(x|e=e_{x'}) - M(x)$. Performing this intervention at every edge is costly, prompting approximate algorithms. 
Edge attribution patching (EAP; \citealp{syed2023attribution})
linearly approximates the IE, $g(e)$, of edge $e=(u, v)$ as follows:
\begin{equation}
    \resizebox{\linewidth}{!}{$g(e) = M(x|e=e_{x'}) - M(x) \approx (z^*_u - z_u)^\top \nabla_v M(x) \label{eq:eap}$}
\end{equation}

The target metric $M$ can vary depending on the task.
Typically, $M$ is the logit difference between a correct completion and a minimally different incorrect completion.
$z_{u}$ and $z^*_{u}$ are the clean and counterfactual activations at the output of $u$, and $\nabla_v M(x)$ is the gradient of $M(x)$ w.r.t the input of $v$. 
\citet{syed2023attribution} showed EAP  to outperform direct activation patching with a greedy approach \citep{conmytowards}.
However, 
\citeauthor{syed2023attribution}\ only discovered circuits that do not consider positions. 

\subsection{Method} Equation~\ref{eq:eap} holds only when $u$ and $v$ are at the same position.
To include token positions in the circuit, attention edges that cross positions must be included in the discovey process. 
In autoregressive Transformer-based models,
these edges exist between nodes representing a given attention head that operates at different positions. 
Let $h^i_{t,l}$ denote the node corresponding to the $i$-th attention head at token position $t$ in layer $l$. 
Following \citet{olah2021framework}, we view the contribution of head $h^i_{t}$ to the residual stream 
as:
\begin{equation}
z_{h^i_{t}} = W^{i}_{O} (\text{softmax} ( \frac{q^i_{t}{K^i_{t}}^\top}{\sqrt{d_k}}  ) V^i_{t} ) \in \mathbb{R}^{d_{\text{model}}}
\label{eq:attn}
\end{equation}
Here, \( W_{O}^{i} \) represents the columns of the projection matrix \( W_{O} \) that specifically project the output of head \( h^i \). \( K^i_{t} \in \mathbb{R}^{t \times d_{\text{head}}} \) is the key matrix, and \( V^i_{t} \in \mathbb{R}^{t \times d_{\text{head}}} \) is the value matrix.

$h^i_{t}$  is connected to 
 every node  $h^i_{t',l}$ at position $t' \le t$, via \textbf{3 edges}: the value vector $v^i_{t',l}$, the key vector $k^i_{t',l}$, and the query vector $q^i_{t,l}$. As direct communication between heads occurs only within the same layer, 
we omit henceforth the layer notation and assume that all attention edges connect attention heads within the same layer.

To approximate the attribution scores of attention edges, we first calculate \( z^*_{h^i_t} \), the corrupted output of the head \( h^i_t \) caused by patching \( v^i_{t'} \), \( k^i_{t'} \), or \( q^i_t \). We then approximate the attribution as follows:
\begin{equation}
    M(x| e = e_{x'}) - M(x) \approx (z^*_{h^i_{t}} - z_{h^i_{t}})^\top \nabla_{z_{h^i_{t}}} M(x) \label{eq:eap-attn}
\end{equation}

Based on Eq.~\ref{eq:attn}, we define the corrupted vector \( z^*_{h^i_t} \) for patching $v_{t'}^i$ (Eq.~\ref{eq:patch-v}), patching $k_{t'}^i$ (Eq.~\ref{eq:patch-k}), and patching $q_{t}^i$ (Eq.~\ref{eq:patch-q}):

\begin{align}
\label{eq:patch-v} \resizebox{\linewidth}{!}{$z^*_{h^i_{t}} = W^{i}_{O} (\text{softmax} \left( \frac{q^i_{t}{K^i_{t}}^\top}{\sqrt{d_k}}\right) \left[ v_1^i, ..., {v_{t’}}^*, ..., v_t^i\right])$}\\
\label{eq:patch-k} \resizebox{\linewidth}{!}{$z^*_{h^i_{t}} = W^{i}_{O} (\text{softmax} \left( \frac{q^i_{t}{\left[ k_1^i, ..., {k_{t’}}^*, ..., k_t^i \right]}^\top}{\sqrt{d_k}}\right)V^i_{t} )$}\\
\label{eq:patch-q} \resizebox{\linewidth}{!}{$z^*_{h^i_{t}} = W^{i}_{O} (\text{softmax} \left( \frac{{\left[ q^i_{t} {k_1^i}^\top, ..., {q^i_{t}}^* {k_{t’}}^\top, ..., q^i_{t} {k_t^i}^\top \right]}}{\sqrt{d_k}}\right)V^i_{t} )$}
\end{align}

\begin{figure}[t]
    \centering
    \includegraphics[scale=0.2]{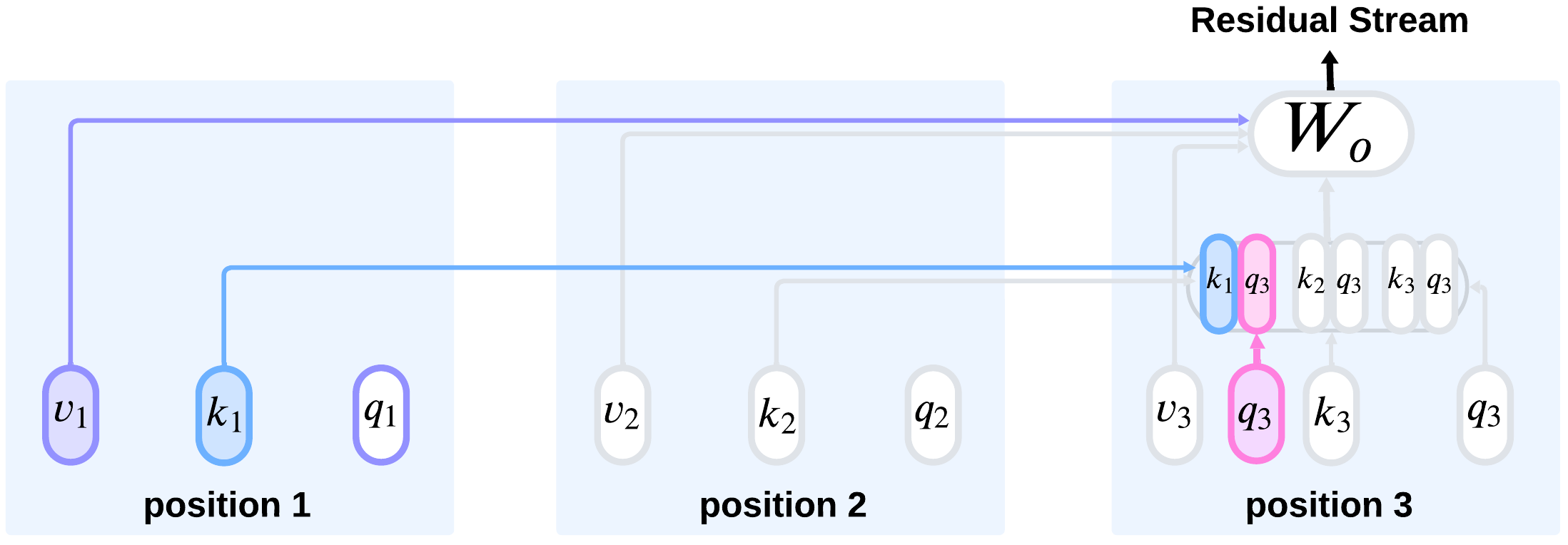}
    \caption{\vspace{-2pt}
    Illustration of the attention mechanism from the perspective of position 3. We approximate how patching \textcolor[RGB]{147,145,255}{$v_1$}, \textcolor[RGB]{109,177,255}{$k_1$} or \textcolor[RGB]{255,128,223}{$q_3$} impacts the downstream metric via the output of the attention head at position 3.}
    \label{fig:attention_patching}
    \vspace{-6pt}
\end{figure}

Figure \ref{fig:attention_patching} provides an illustration of each type of patching. 
By using PEAP to approximate attention edges, we can now approximate both within-position edges and cross-position edges.

Once the attribution scores for all edges have been computed, we construct the circuit using an adapted version of the greedy algorithm proposed by \citet{hanna2024have}. See App.~\ref{ap:circuit construcion} for details. 

\vspace{-2pt}
\subsection{Preliminary Demonstration}
\vspace{-2pt}
We now compare PEAP to the position-agnostic approach of \citet{syed2023attribution} using the Greater-Than task \citep{hanna2024does} on GPT2-small \citep{radford2019language}.
The dataset includes prompts like: ``The war lasted from the year 1741 to the year 17\underline{\hspace{1em}}'' and counterfactual variants with ``01'' as the starting year (e.g., ``The war lasted from the year 1701 to the year 17\underline{\hspace{1em}}'').
The downstream metric $M$ measures the probability difference between valid and invalid year answers.
We use 500 examples each for circuit discovery and evaluation, considering only prompts with valid model predictions.
Circuit evaluation is based on two metrics: (1) \textbf{Soft Faithfulness} ($F_S(C) = \frac{M(C)}{M(\mathcal{M})}$), comparing the circuit's performance to the full model's, and (2) \textbf{Hard Faithfulness} ($F_H(C) = \mathds{1}\{C_T = \mathcal{M}_T\}$), assessing token agreement at the final position $T$.
While $F_S$ is more commonly used, we see $F_H$ as a more behaviorally grounded metric.

\begin{figure*}[t]
    \centering
    \includegraphics[width=0.95\linewidth]{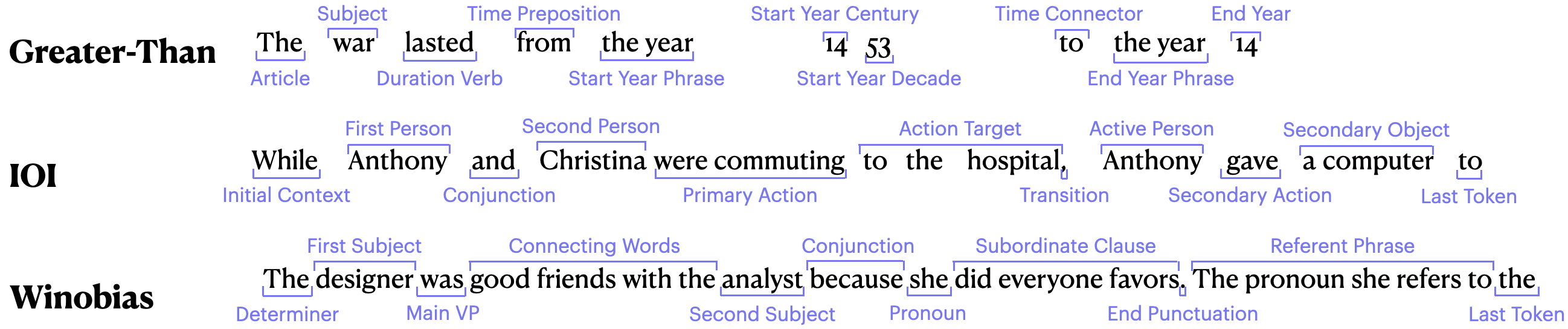}
    \caption{Example schema for each task. We show examples from the LLM+Mask method. See \S\ref{ap:task details} for examples of human-designed schemas.}
    \label{tab:schema-example}
    \vspace{-5pt}
\end{figure*}

Figure~\ref{fig:p1} presents the faithfulness scores of the Greater-Than task for both methods as a function of circuit size. PEAP enables the discovery of circuits that improve the trade-off between circuit size and faithfulness: \textbf{position-aware circuits are smaller, and yet achieve similar faithfulness with orders-of-magnitude fewer edges.}

\subsection{Aggregating Scores Across Examples}

In the Greater-Than dataset, we can simply aggregate position-specific scores across examples.
This naive approach works because all examples in the Greater-Than dataset consist of exactly the same number of tokens, and each position has the same meaning across all examples.
In other words, this approach requires all examples in the dataset to be \emph{fully position-aligned}.
This raises a key challenge for non-templatic datasets: the same token position may not have the same meaning across examples, and examples may vary in length.

Prior methods addressing positionality typically follow one of two strategies: 
(1) \textbf{full alignment}, where the dataset is generated from a single template---as in the Greater-Than dataset---and (2) \textbf{partial alignment}, where specific token position roles are consistent across examples.
For instance, in the IOI dataset \citep{wanginterpretability}, the authors \emph{manually} identified five key single-token roles (IO, S1, S1+1, S2, End) shared across all prompt templates, which are sufficient for constructing a faithful circuit. 
In the next section, we describe an automatic approach inspired by partial alignment that  enables us to include positional information in tasks with variable-length inputs.

\section{Schemas for Variable-length Inputs} 
\label{sec:schema}

Discovering circuits requires aggregating edge scores across examples. 
However, because edges correspond to specific positions in the computation graph, naive aggregation assumes perfect positional alignment across examples---an impractical assumption for most datasets.
To address this challenge, we relax this assumption and only assume that examples share a similar high-level structure, which is represented by a \textbf{schema}.
A dataset schema identifies \textit{spans} within input examples, where each span covers consecutive tokens grouped under a meaningful category.
For instance, in the input ``The \textcolor{positional}{war} lasted from 1453 to 14\underline{\hspace{1em}}'', the span ``\textcolor{positional}{war}'' could be labeled \emph{Subject}.
This allows us to handle spans of varying lengths, such as treating ``\textcolor{positional}{Black Plague}'' in another example as a single position with the same role as ``\textcolor{positional}{war}''.
Examples of schemas for specific datasets are shown in Figure~\ref{tab:schema-example}.
Schemas are defined based on semantic, syntactic, or other patterns in the data, and may be guided by knowledge of how the model processes examples.
Spans are ordered sequentially within the input, covering all parts of a prompt.\footnote{Future work may relax the sequential order assumption to support even greater variation across examples.}

\subsection{Discovering Circuits at the Schema Level}
When all examples share the same schema-defined structure, we can leverage this consistency to create an abstract computation graph for all examples.
For now, we assume spans in the schema can be automatically mapped to corresponding tokens in any dataset sample. We discuss automating this process later.

Let $ \smash{ G_{x}=(E_x,V_x) }$ represent the computation graph derived from example $x \in \mathcal{D}$.
Given schema $\mathcal{S}$ with $k$ spans, we define the \textit{abstract} computation graph $G_\mathcal{S}=(E_\mathcal{S},V_\mathcal{S})$, which is structurally equivalent to a computation graph of $\mathcal{M}$ on an input of length $k$. Intuitively, each span is represented by a single position.

At a high level, given an example, we (i) compute edge scores on the true computation graph $G_x$;
(ii) map from edges in $G_x$ to edges in $G_\mathcal{S}$, and sum edge scores in $G_x$ to compute edge scores in $G_\mathcal{S}$;
(iii) construct a circuit in $G_\mathcal{S}$.

To this end, we define a mapping $f_{x}: E_{\mathcal{S}} \rightarrow 2^{E_x}$ from an edge $e=(u_{s_1},v_{s_2})$ to a set of edges in $E_x$:
\begin{equation}
    f_{\mathcal{S}}^x(e)=\{ e'\in G_x \mid e=(u_i, u_j), i \in s_1, j\in s_2 \}
\end{equation}
where $u_{s_1},v_{s_2}$ represent components in the computation graph at spans $s_1, s_2$. 

Given an attribution function $g_x$ (defined at the token position level), the attribution score $g_\mathcal{S}$ (defined at the segment level) of the edge $e\in G_{\mathcal{S}}$ is the sum of all the edge effects mapped to this edge, averaged over all examples in the task dataset:

\begin{equation}
    g_{\mathcal{S}}(e) = \frac{1}{|\mathcal{D}|}\sum_{x \in \mathcal{D}}\sum_{e' \in f^x_\mathcal{S}(e)} g_{x}(e')
\end{equation}

\begin{figure}[t]
    \centering
    \includegraphics[width=0.97\linewidth]{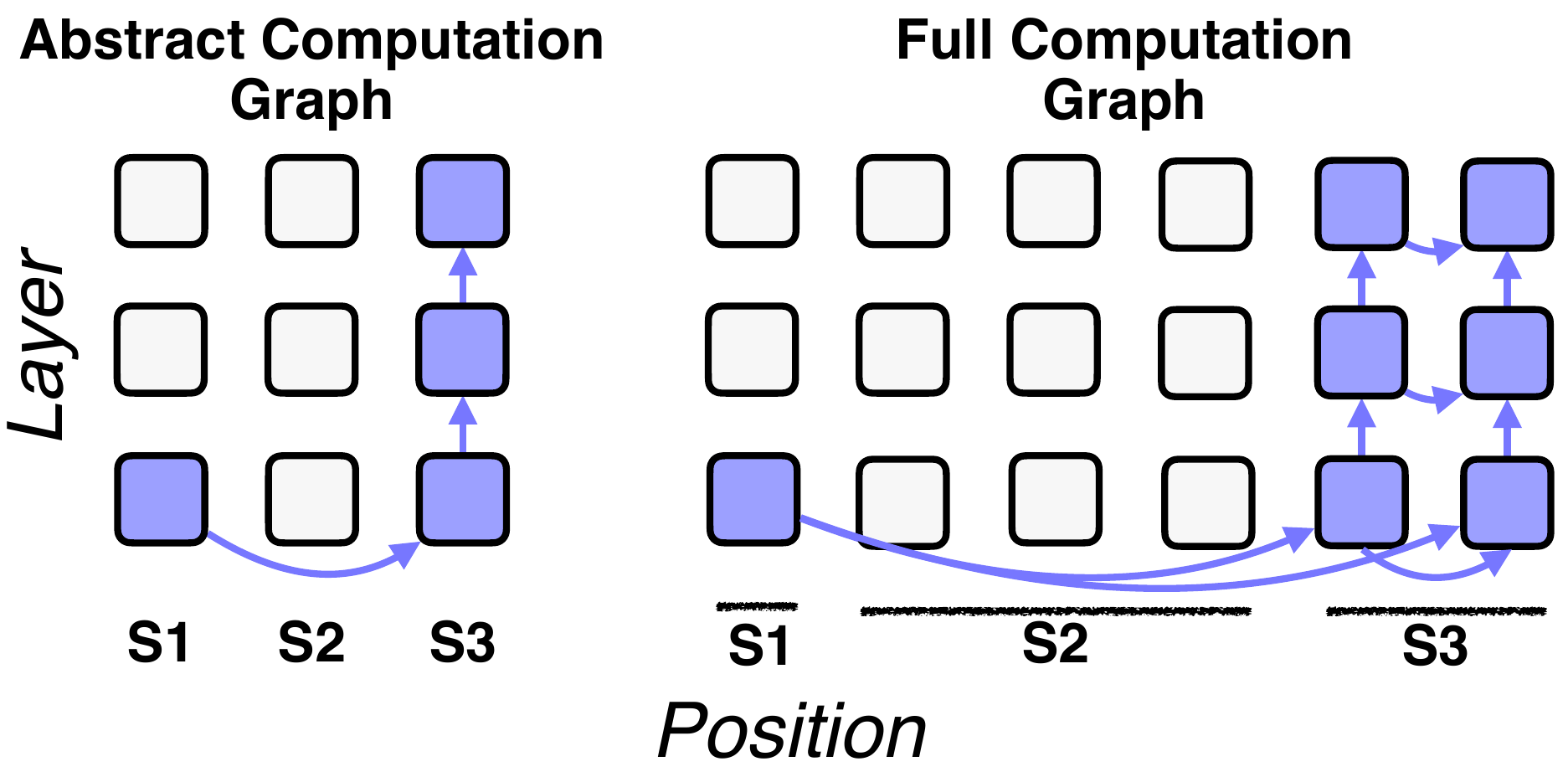}
    \caption{Circuits defined over schemas. Every node/edge at position \( s \) in the abstract computation graph is mapped to a set of nodes/edges in the full computation graph within the span \( s \).}
    \label{fig:abs_graph}
\end{figure}

After computing the attribution score for each edge in $G_{\mathcal{S}}$, we construct the abstract circuit  $\mathcal{C}_{\mathcal{S}} \subseteq G_{\mathcal{S}}$ with the same greedy algorithm used in the previous section (see App.~\ref{ap:circuit construcion}).

\paragraph{Faithfulness evaluation.}
The process of faithfulness evaluation involves ablating edges that are not included in the circuit. 
To evaluate an abstract circuit on a sample $x\in\mathcal{D}$, we convert back to the computational graph $G_x$ and construct $\mathcal{C}_x \subseteq G_x$:
\begin{equation}
\mathcal{C}_x = \{ e \mid e \in f^x_\mathcal{S}(e'), \forall e' \in C_{\mathcal{S}} \}
\end{equation}
In other words, for every edge $e'$ in the abstract circuit $C_{\mathcal{S}}$, the corresponding edges in $f_x(e')$ form the circuit $\mathcal{C}_x$.
Figure \ref{fig:abs_graph} depicts this process.

\subsection{Automating Schema Generation and Application} \label{sec:schema-generation}
Given a schema $ \mathcal{S} $ and a function $ f_\mathcal{S} $ to apply it to every sample $ x \in \mathcal{D} $, we can automatically discover position-aware circuits, even for tasks involving variable-length examples. 
However, as shown in Figure \ref{tab:schema-example}, schema definitions are dataset-specific, requiring tedious manual work and intricate knowledge of the task at hand as well as knowledge of the analyzed model's computations.
Applying the schemas may also require deep knowledge on the target dataset.
To generate interpretable circuits, schemas must be both faithful to the model and meaningful to humans.

In this section, we propose an automated process for schema generation and application to streamline circuit discovery.
Inspired by recent work on LLM agents \cite{wang2024survey} for automated interpretability \cite{schwettmann2023find,shaham2024multimodal}, we investigate the use of LLMs for generating and applying schemas.

\paragraph{Schema Application.}
Applying a schema entails mapping each token to a specific span.
After defining the schema, we utilize an LLM to perform the application process.
We provide the prompt for applying the schema in App.~\ref{ap:schema application}.

\paragraph{Schema Generation.}
Creating a schema requires specifying span types while two conditions: (1) spans must follow the same order across all examples, and (2) each prompt must be fully covered by the spans.
These criteria are incorporated into the LLM's prompt (details in App.~\ref{ap:schema generation}).
Given a dataset, we use an LLM to create three schema versions based on distinct subsamples, then have the LLM unify these versions into a final schema.
The schema is validated by confirming it applies to at least 80\% of the subsampled data;
otherwise, the process is repeated.
Examples of LLM-generated schemas are shown in Figure~\ref{tab:schema-example}.

\paragraph{Saliency scores: A model-based approach for schema generation.}

The schema generation described above does not account for the computations performed by the target model $\mathcal{M}$ on the given dataset $\mathcal{D}$, potentially producing unfaithful schemas (as we will show in \S\ref{sec:results}).
To address this, we incorporate the importance of each token position to the model's computation into the schema generation.

Our key idea is to inform the LLM which positions significantly influence the model's decisions.
While many feature attribution methods can be explored \cite{danilevsky-etal-2020-survey,wiegreffe2021teach,wallace-etal-2020-interpreting}, we employ a simple saliency score, inputXgradient \cite{shrikumar2017learning}.
The score of a token in position $t$ is defined as 
$ s(t) = \| \mathbf{e_t} \cdot \nabla_{\mathbf{e_t}} M(x) \| $, where $ \mathbf{e_t} $ is the token embedding at position $ t $.
We compute a softmax over these scores and define a mask for each example as follows:
\begin{equation}
m(t) =
\begin{cases}
1 & \text{if } \frac{e^{s(t)}}{\sum_{i=1}^{n} e^{s(i)}} > \frac{1}{n}, \\
0 & \text{otherwise},
\end{cases}
\end{equation}
Where $n$ is the prompt length.
This mask is then attached to each example, and the LLM is instructed to use it when designing the schema.
Token position which is important across many examples should be placed in its own span. Further information on mask construction and alternative attribution methods can be found in Appendix~\ref{ap:mask-creation}.

\paragraph{Schema Evaluation.}\label{sec:schema-eval}
We propose two intrinsic metrics and one extrinsic metric to evaluate the entire schema pipeline.
Intrinsic metrics assess the LLM schema application. An application is \textbf{valid} if span labels are ordered correctly and every token is assigned to a single span, and \textbf{correct} if it matches a human application for the same schema.
Extrinsic metrics evaluate schema design and application through circuit discovery.
A good schema definition and application should achieve better trade-offs between circuit size and faithfulness.

Invalid schema applications are filtered out for both the discovery and evaluation datasets, while incorrect applications are retained since automating their filtering is infeasible in general datasets.
If an application is valid but incorrect, we expect it to affect the faithfulness of the discovered circuit.
To ensure minimal distribution shift in the dataset, we consider a generation and an application of a schema on an entire dataset as successful if at least 90\% of the examples are valid.
This means that each circuit is discovered using a slightly different set of examples (up to 10\%), but we ensure that all circuits are compared using the exact same evaluation set, which is the intersection of the examples for all runs. In practice this intersection includes 90\% of the total dataset examples.
In our experiments, three full pipeline runs were usually sufficient to achieve at least one successful run.

We found Claude 3.5 Sonnet \citep{claude3} to perform well in both schema generation and application, achieving high validity and correctness scores (Table~\ref{tab:schema-eval}, Appendix~\ref{ap:schema-validation}).
We also experimented with Llama-3-70B \citep{grattafiori2024llama3} and GPT-4o \citep{openai2024gpt4ocard}, but they failed to meet our thresholds for valid applications.
In \S\ref{sec:results}, we show that LLM-generated schemas score well on extrinsic quality measures, with saliency-enhanced schemas proving comparable to human-designed ones.

\section{Experiments}
In all experiments, we use Llama-3-8B\footnote{We use Llama-3 with BF16 precision.} and GPT2-small. The experiments are implemented using the Transformerlens library \citep{nanda2022transformerlens}.

\subsection{Tasks}
For all tasks, we uniformly sample 500 examples for circuit discovery and another 500 examples for evaluating faithfulness.

\begin{figure*}[t!]
    \centering
    \includegraphics[width=0.245\linewidth]{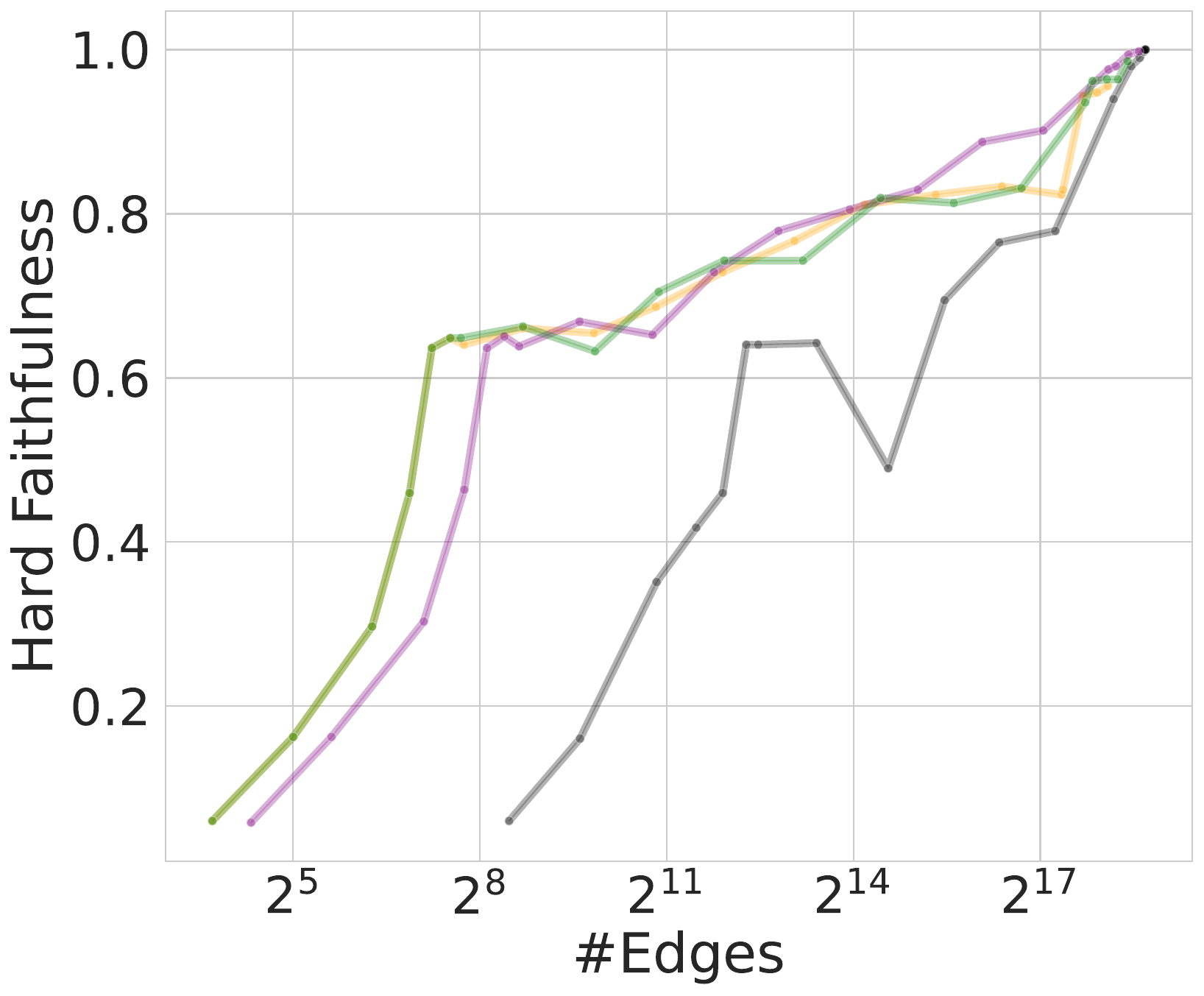} \hfill \includegraphics[width=0.245\linewidth]{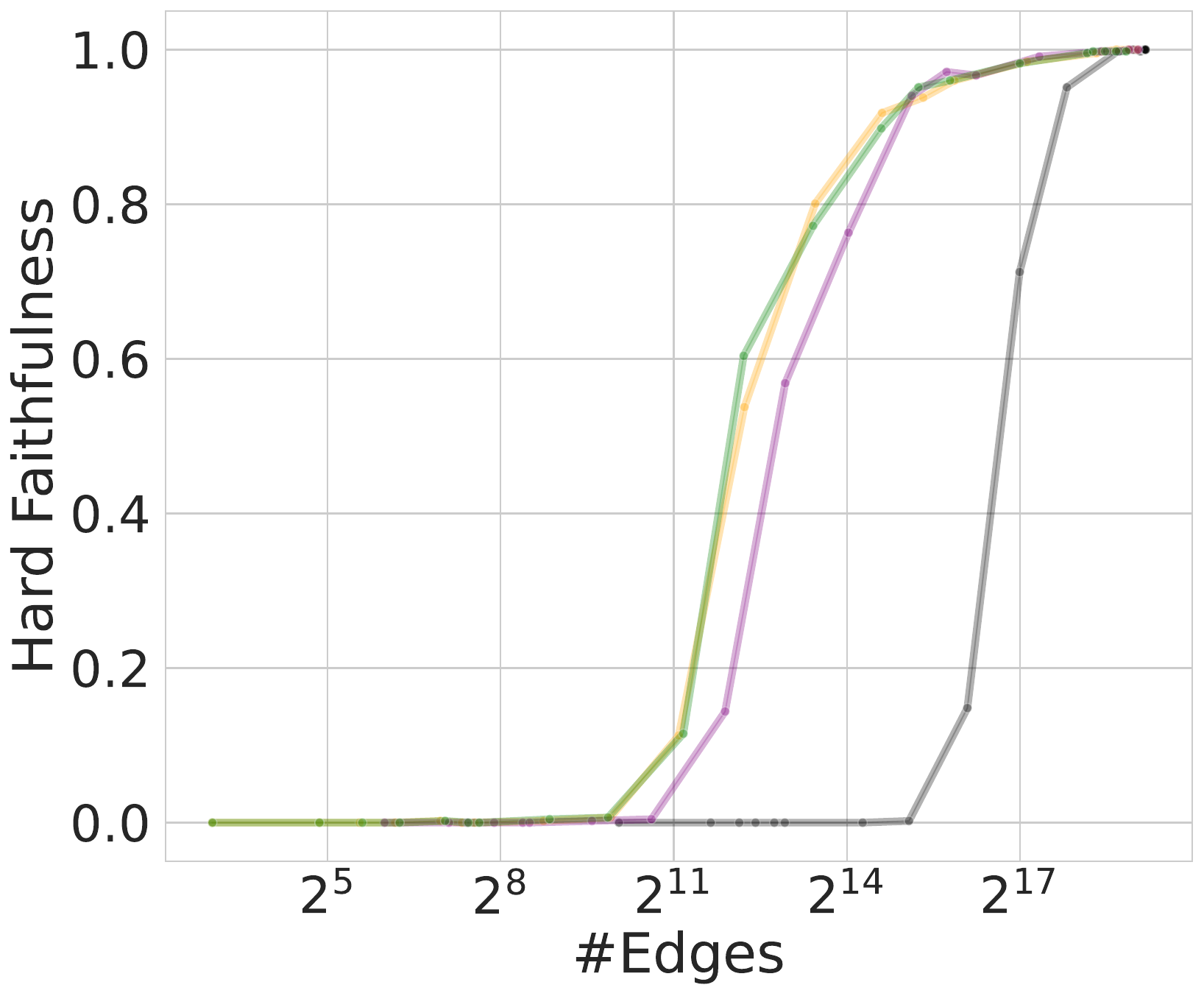} \hfill \includegraphics[width=0.245\linewidth]{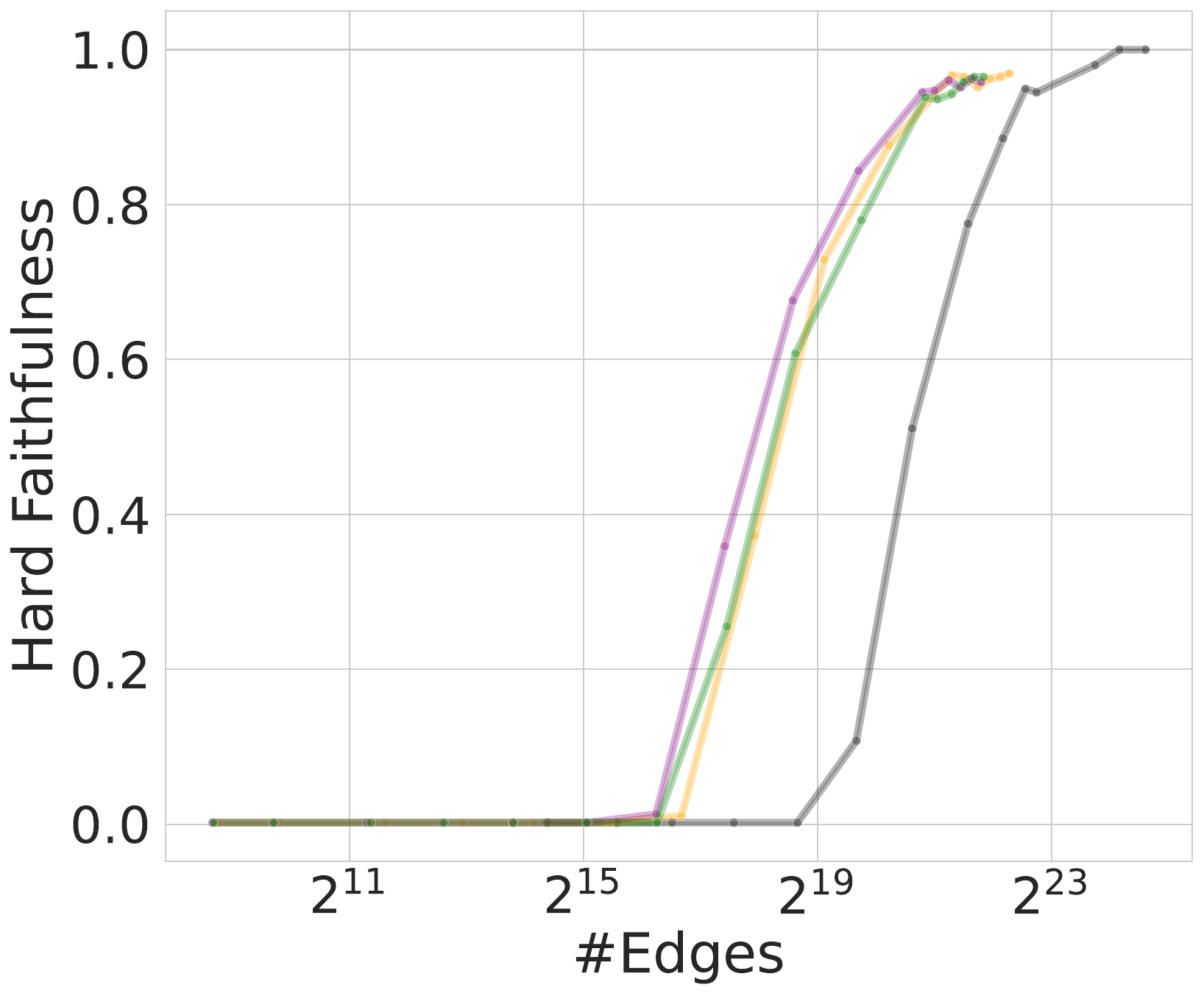} \hfill \includegraphics[width=0.245\linewidth]{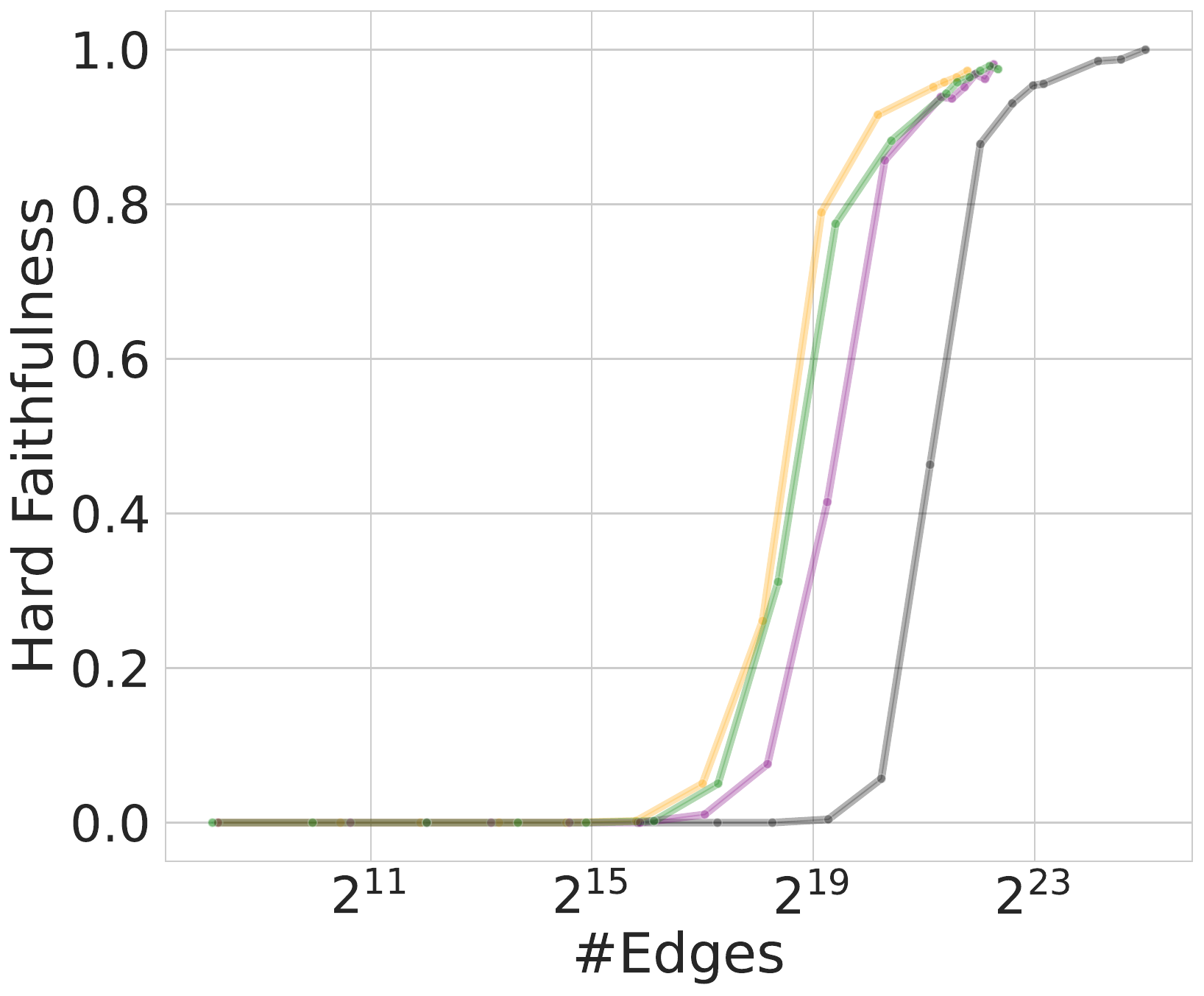}
    \vspace{0.05cm}
     \includegraphics[width=0.55\linewidth]{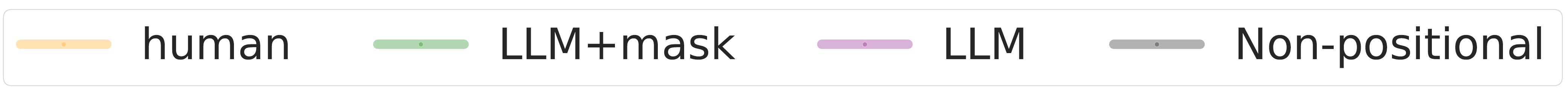} \hfill 
    \caption{Hard faithfulness 
    curves for GPT-2-small on Greater-Than (left) and IOI (mid-left), and for Llama-3-8b on IOI (mid-right) and Winobias (right).}
    \label{fig:faithfulness_all_main}
    \vspace{-5pt}
\end{figure*}

\textbf{Indirect Object Identification} (IOI; \citealp{wanginterpretability}): This task consists of prompts like ``When Mary and John went to the store, John gave a drink to'', and the model should predict the indirect object token `Mary'. The counterfactual prompts for this task are prompt of the same structure but with 3 other unrelated names, for example: ``When Dan and David went to the store, Sarah gave a drink to''. The metric that is being measured here is the logit difference between the token `Mary' and the token `John'. We evaluate with both GPT2-small and Llama-3-8B. For each model, we construct a dataset based on only examples where the model can predict the correct answer.

\textbf{Greater-Than} \citep{hanna2024does}: We use the same setting as described in \S\ref{sec:motivating}. We evaluate this task only on GPT2-small, as Llama-3-8B's tokenizer is not compatible with the task setup; see App.~\ref{ap:task_details_gt} for details.

\textbf{Winobias} \citep{zhao2018gender}: 
A benchmark designed to evaluate gender bias in coreference resolution.
We collect 33 template from the dataset where professions are irrelevant to the coreference decision (e.g., ``The doctor offered apples to the nurse because she had too many of them'').
For each sample, we append the suffix: ``The pronoun \texttt{\{\}} refers to the'', where \texttt{\{\}} is a placeholder for the pronoun.
Each template can be used to construct four types of prompts: Anti-Female, Anti-Male, Pro-Female, Pro-Male. For example of each prompt see Table~\ref{tab:bias_examples}. We focus on the Anti-Female prompts, using only examples where the model predicts the \emph{incorrect} answer due to bias. This approach aims to identify components responsible for biased predictions. For Winobias, counterfactual prompts can be designed in multiple ways, each affecting the kinds of components one would recover; see Appendix~\ref{ap:task details} for further discussion. To avoid counterfactual-specific biases, we use mean ablation with examples from all four types during circuit discovery and faithfulness evaluation. The downstream metric $M$ is the logit difference between the correct profession and incorrect profession.
For further details on all tasks, see App.~\ref{ap:task details}.

\subsection{Circuit Evaluation}  
We measure faithfulness as a function of circuit size. Since different examples may produce circuits of varying sizes (due to differences in span lengths across examples), at each point we report the average circuit size across all examples. We extend the approach of \citet{hanna2024have} for ablating edges to also include attention edges.

\section{Results} \label{sec:results}

Figure~\ref{fig:faithfulness_all_main} shows hard faithfulness for multiple tasks and models. 
\textbf{The positional circuits reach high faithfulness at much smaller circuit sizes compared to the non-positional circuits.}

Using LLM-generated schema works well, and adding mask information yields an additional significant boost. Thus, providing the LLM with information about the target models' computation aids in generating effective schemas. 
Discovering circuits with automatic LLM+mask schemas leads to faithulness results that are as good as---and sometimes better than---human-designed schemas. Thus, \textbf{our automated LLM-based schema pipeline discovers circuits with faithfulness comparable to those identified by human experts, even for tasks containing variable-length inputs.}

We now discuss task-specific patterns. In the Greater-Than task, the circuit discovered with the schema via LLM+mask achieves a faithfulness not significantly different from the human-designed schema. The circuit generated solely by the LLM demonstrates lower faithfulness for smaller circuit sizes but achieves higher faithfulness as the circuit size increases. Comparing the  schemas reveals that the schema derived using saliency scores aligns more closely with the human-designed schema. Specifically, both the human-crafted schema and the LLM+mask schema partition the start year to two spans: the first two digits and the last two digits. However, in the LLM-only schema, all four digits are grouped in a single span.

In the IOI task using GPT2-small, we observe that the circuits identified by our automated pipeline closely match the human-designed circuits in faithfulness. However, in the case of Llama-3-8B, the LLM-generated circuits show slightly superior faithfulness compared to human-designed circuits. One plausible explanation is that the IOI task has not been extensively investigated in this larger model, meaning the schema defined for GPT2-small may not optimally capture the nuances of this task in Llama-3-8B. This highlights the importance of tailoring schemas to the specific combination of task and model, rather than extrapolating from results obtained with a different model.

For the Winobias task dataset, similar trends emerge: using the importance mask consistently improves faithfulness scores, making it comparable to the human-defined schema-based circuit.

\section{Discussion and Conclusions}
In mechanistic investigations, \textbf{position matters.} Our results suggest it does not make sense in practice to create circuits without considering how distinct the circuit at each position might be.
Theoretical results suggest that it also does not make sense in principle to ignore positionality: \citet{merrill2024expressivepowertransformerschain} show that transformers' expressive power increases with multiple generation steps.
Similarly, accounting for positionality in interpretability methods can enhance their expressive power by capturing the distinct mechanisms processing each token, rather than assuming a single pathway for the entire sequence.

Other interpretability methods such as distributed alignment search (DAS; \citealp{geiger2024das}) already support testing hypotheses about the position of particular causal variables. It would be interesting to directly compare the efficacy of DAS methods when separating results by position versus when aggregating information across positions. Stronger results when separating positional information could help generalize our conclusions to a wider array of causal interpretability methods.

\section*{Limitations}
A key limitation we have discussed is that it is not trivial to handle positional information in tasks where the length of inputs vary. We have proposed an automatic pipeline for generating \emph{and} applying schemas, but future work should explore this further. In particular, because there is no single gold standard for schemas, it is not clear \emph{a priori} what kinds of schemas are generally likely to obtain better trade-offs between faithfulness and circuit size. Devising general principles for effective schema design therefore represents a fruitful avenue for future work. It would also be interesting to observe whether human-generated schemas tend to satisfy these principles, or whether the most effective schemas are not necessarily those that humans are likely to design.

Another key limitation is that a schema requires the same spans to appear in the same order across all examples, such that the edges' direction remains correct across examples. Consequently, two schemas with the same span types but in different orders cannot be evaluated together, as these produce different abstract computation graphs. 

\section*{Acknowledgments}
 This research was supported by the Israel Science
Foundation (grant No.\ 448/20), an Azrieli Foundation
Early Career Faculty Fellowship,  an AI Alignment grant from Open Philanthropy, and a Google gift. HO is supported by the Apple AIML PhD fellowship. DB is supported by a grant from Open Philanthropy. AM is supported by a postdoctoral fellowship under the Zuckerman STEM Leadership Program.
This research was funded by the European Union (ERC, Control-LM, 101165402). Views and opinions expressed are however those of the author(s) only and do not necessarily reflect those of the European Union or the European Research Council Executive Agency. Neither the European Union nor the granting authority can be held responsible for them.

\bibliography{references}

\appendix

\section{Tasks Details} \label{ap:task details}

\subsection{IOI}
\label{ap:task details:ioi}

We use the dataset of \citet{wanginterpretability}. The data is generated using 15 templates. For the human-defined schema (provided below), we extend the partial schema provided by the authors to fully cover all spans in the prompt. The original dataset includes two types of prompts: ABBA prompts, where the indirect object (IO) token is the first name in the prompt, and BABA prompts, where the IO token appears as the second name. Because the ABBA and BABA prompts swap the order of important spans, we cannot aggregate across these two prompt types. Thus, we designed two distinct schema, resulting in the definition of two separate datasets: 

\noindent \textbf{IOI ABBA}:   
The human-defined schema and its application:

\begin{itemize}[itemsep=0pt, topsep=0pt]
    \item \textbf{Prefix}: [When]  
    \item \textbf{IO}: [Mary]  
    \item \textbf{and}: [and]  
    \item \textbf{S1}: [John]  
    \item \textbf{S1+1}: [went]  
    \item \textbf{action1}: [to the store,]  
    \item \textbf{S2}: [John]  
    \item \textbf{action2}: [gave a drink]  
    \item \textbf{to}: [to]  
\end{itemize}

\noindent \textbf{IOI BABA}: The human-defined schema and its application:  
\begin{itemize}[itemsep=0pt, topsep=0pt]
    \item \textbf{Prefix}: [When]  
    \item \textbf{S1}: [John]  
    \item \textbf{S1+1}: [and]  
    \item \textbf{IO}: [Mary]  
    \item \textbf{S1+1}: [went]  
    \item \textbf{action1}: [to the store,]  
    \item \textbf{S2}: [John]  
    \item \textbf{action2}: [gave a drink]  
    \item \textbf{to}: [to]  
\end{itemize}

Table \ref{tab:model_accuracy} summarizes the performance of GPT2-small and Llama-3-8B for this task.
For the results in \S\ref{sec:results}, we use the ABBA datasets. Results for both datasets can be found in \S\ref{ap:faithfulness_curves}.
Note that there exist schemas that can handle both datasets simultaneously, eliminating the need for separation. However, these schemas require grouping the IO token and the S1 token into the same span, which mixes signals from both token positions and introduces new drawbacks.

\subsection{Greater-Than}\label{ap:task_details_gt}
We use the dataset of \citet{hanna2024does}. All examples in this task are generated using a single template:
``The \{\} lasted from the year \{\} to the year \{\}''. Because the event span (the first non-terminal) and the years have the same token length for all of our models, all examples in the dataset are fully token-aligned.

For the human-designed schema, we adopted the word-level schema used by \citet{hanna2024does}: \begin{itemize}[itemsep=0pt, topsep=0pt]
    \item \textbf{The}: [The]
    \item \textbf{Noun}: [war]
    \item \textbf{lasted}: [lasted]
    \item \textbf{from}: [from]
    \item \textbf{the}: [the]
    \item \textbf{year}: [year]
    \item \textbf{XX1}: [16]
    \item \textbf{YY}: [45]
    \item \textbf{to}: [to]
    \item \textbf{the}: [the]
    \item \textbf{year}: [year]
    \item \textbf{XX2}: [16]
\end{itemize}

Table \ref{tab:model_accuracy} summarizes the performance of GPT2-small for this task.

\begin{table}[!ht]
\centering
\begin{tabular}{lrr}
 \toprule
 Dataset & GPT2-small & Llama-3-8B \\
 \midrule
 IOI-ABBA   & 92.5\%  & 71.7\% \\
 IOI-BABA   & 87.1\%  & 73.4\% \\
 Greater-Than & 100\% & 0.0\% \\
 \bottomrule
\end{tabular}
\caption{Accuracy on different datasets. Note that Llama-3 performs poorly on Greater-Than due to tokenization: it tokenizes four-digit years as a token of three digits, followed by a token of one digit. The Greater-Than prompts end with a two-digit token, which the model has presumably not seen followed by another two-digit number token; thus, it instead generates whitespaces, and assigns a low probability to the true answer.}
\label{tab:model_accuracy}
\end{table}

\subsection{Winobias}
\label{ap:task details:wb}

We use the dataset templates of \citet{zhao2018gender}. We constructed the dataset based on 33 pairs of prompt templates and 32 single-token professions. Each profession is associated with gender stereotypes.

\textbf{Stereotypical female professions}: teacher, editor, cashier, writer, counselor, counselors, nurse, sewer, baker, auditor, designer, clerk, cleaner, librarian, secretary, assistant, attendant, accountant.

\textbf{Stereotypical male professions}: lawyer, farmer, physician, developer, manager, guard, mechanic, chief, driver, mover, supervisor, analyst, constructor, cook, sheriff.

The original dataset consists of pairs of templates, each sharing the same prefix but having different suffixes. This variation causes the position of the correct answer to change between templates (e.g., ``The doctor offered apples to the nurse because she had too many of them'' and ``The nurse offered apples to the doctor because she might like them''). We separate templates where the first subject is the correct answer from those where the second subject is the correct answer. This separation ensures we do not mix biased signals with non-biased signals during the circuit discovery process.

Moreover, for each template, we construct four types of prompts: Anti-Female, Anti-Male, Pro-Female, and Pro-Male. In total, starting from 33 pairs of templates, we construct eight distinct datasets. Table \ref{tab:bias_examples} provides examples for each type of dataset. Intuitively, ``Anti-'' means that the profession-pronoun relationship goes against conventional gender biases; ``Pro-'' means that the the profession-pronoun relationship conforms to conventional gender biases. For example, if the correct answer is ``nurse'' and the pronoun is ``she'', we would say that this is a Pro-Female example; compare to the case where the pronoun is ``he''; this changes it to an Anti-Male example.

For the main results we used the Anti-Female-I dataset. Results for the Anti-Female-II dataset can be found in \S\ref{ap:faithfulness_curves_wb}.

\begin{table*}[ht]
\centering
\resizebox{\linewidth}{!}{
\begin{tabular}{ ll }
 \toprule
 Dataset & Example\\
 \midrule
 Anti-Female-I   & The \textbf{doctor} offered apples to the \textbf{nurse} because \textbf{she} had too many of them. The pronoun \textbf{she} refers to the" \\
 Anti-Female-II  & The \textbf{nurse} offered apples to the \textbf{doctor} because \textbf{she} might like them. The pronoun \textbf{she} refers to the" \\
 Pro-Female-I    & The \textbf{nurse} offered apples to the \textbf{doctor} because \textbf{she} had too many of them. The pronoun \textbf{she} refers to the"\\
 Pro-Female-II   & The \textbf{doctor} offered apples to the \textbf{nurse} because \textbf{she} might like them. The pronoun \textbf{she} refers to the" \\
 Anti-Male-I     & The \textbf{nurse} offered apples to the \textbf{doctor} because \textbf{he} had too many of them. The pronoun \textbf{he} refers to the"  \\
 Anti-Male-II    & The \textbf{doctor} offered apples to the \textbf{nurse} because \textbf{he} might like them. The pronoun \textbf{he} refers to the" \\
 Pro-Male-I      & The \textbf{doctor} offered apples to the \textbf{nurse} because \textbf{he} had too many of them. The pronoun \textbf{he} refers to the"\\
 Pro-Male-II     & The \textbf{nurse} offered apples to the \textbf{doctor} because \textbf{he} might like them. The pronoun \textbf{he} refers to the"\\
 \bottomrule
\end{tabular}
}
\caption{An example for each dataset. Each entry demonstrates a pronoun resolution scenario, with variations designed to reflect anti-female, pro-female, anti-male, and pro-male biases.}
\label{tab:bias_examples}
\end{table*}

Table \ref{tab:bias_assessment} presents the performance of Llama-3-8B on each type of dataset. We evaluated how often the model responded with the plausible answer and how often it chose the biased answer. Note that the model is not forced to select either of these options, and therefore, the sum of the percentages in each row does not necessarily equal 100\%.

\begin{table}[ht]
\centering
\resizebox{\linewidth}{!}{ 
\begin{tabular}{ lrrr }
 \toprule
 Dataset & Correct Answer & Biased Answer & \emph{Sum} \\
 \midrule
 Anti-Female-I   & 34.5\%  & \textbf{64.8\%}& 99.3\% \\
 Anti-Female-II  & 29.2\%  & \textbf{69.5\%}& 98.7\% \\
 Pro-Female-I    & \textbf{81.6\%} & 17\%&  98.6\%\\
 Pro-Female-II   & \textbf{75.9\%} & 23.2\%& 99.1\%\\
 Anti-Male-I     & \textbf{51.9\%} & 47.4\%&  99.3\%\\
 Anti-Male-II    & 35.8\%& \textbf{63\%}& 98.8\% \\
 Pro-Male-I      & \textbf{79.2\%} & 19.3\%& 98.5\% \\
 Pro-Male-II     & \textbf{61.7\%} & 37.8\%&  99.5\%\\
 \bottomrule
\end{tabular}
}
\caption{Bias measurement across the different datasets. The sum indicates the proportion of examples for which neither the correct nor the biased answer was the top token according to the model.}
\label{tab:bias_assessment}
\end{table}

For the human-defined schema, we used a schema defined by \citet{zhao2018gender} with minor adjustments: 
\begin{itemize}[itemsep=0pt, topsep=0pt]
    \item \textbf{correct answer}: [The doctor]
    \item \textbf{interacts with}: [offered apples to]
    \item \textbf{wrong answer}: [the nurse]
    \item \textbf{conjunction}: [because]
    \item \textbf{pronoun1}: [she]
    \item \textbf{circumstances}: [had too many of them]
    \item \textbf{dot}: [.]
    \item \textbf{The}: [The]
    \item \textbf{pronoun}: [pronoun]
    \item \textbf{pronoun2}: [she]
    \item \textbf{refers}: [refers]
    \item \textbf{to}: [to]
    \item \textbf{the}: [the]
\end{itemize}

For datasets where the wrong answer appears as the first subject, we swap the order of the answers in the schema.

Note that counterfactuals can be defined in many ways for this task; this complicates locating and interpreting circuits. For example, one could define counterfactuals from Anti-Male to Anti-Female, Anti-Male to Pro-Female, among others; each of these would isolate only some bias-specific subcircuit of the full coreference resolution circuit. To overcome biases that would result from using counterfactual prompts, we instead use mean ablations constructed from 16 examples spanning all examples (Anti-Female, Anti-Male, Pro-Female, Pro-Male); this is more likely to recover the full coreference resolution circuit.\footnote{As compared to, for example, the subcircuit that encodes gender bias.} 

\section{The Computation Graph} \label{ap:computation_graph}

The computation graph consists of the following node types: MLPs, attention heads, embeddings, and logits. To account for token positions, each node type has a separate instance at every position. Following \citet{wanginterpretability}, the input edge to an attention head is divided into three components: \texttt{v\_input}, \texttt{k\_input}, and \texttt{q\_input}. Consequently, three distinct edges connect every node \(v\) to a downstream attention head \(u\). Additionally, each attention head is connected to all attention heads at subsequent token positions via three types of connections: \(v\), \(k\), and \(q\).

The size of the computation graph varies depending on the model size \emph{and} prompt length. Table \ref{tab:graph_size} summarizes the average computation graph size for each dataset and model.

\begin{table}[!ht]
\centering
\begin{tabular}{lrr}
 \toprule
 Dataset & GPT2-small & Llama-3-8B \\
 \midrule
 IOI-ABBA   & 593,473.55  & 25,746,710.46 \\
 IOI-BABA   & 584,783.47 & 25,654,744.33 \\
 Greater-Than & 423,59.0 & - \\
 Winobias-I & - & 33,769,270.68 \\
 Winobias-II & - & 32,951,977.84 \\
 \bottomrule
\end{tabular}
\caption{Average number of edges in the computation graph per task.}
\label{tab:graph_size}
\end{table}

\section{Circuit  Construction} \label{ap:circuit construcion}

Once the attribution scores for all edges in the graph are approximated, there are several ways to construct a circuit. A straightforward approach might involve selecting components with the highest scores to construct the circuit. However, this naive method often results in a circuit that lacks proper connectivity between embeddings and logits. To ensure connectivity, we adopt a slightly modified version of the algorithm proposed by \citet{hanna2024have}.

As \citet{hanna2024have} states, this algorithm is a greedy method, similar to a maximizing version of Dijkstra’s algorithm. The process begins with a circuit containing only the logits node at the final token position. At each step, the edge with the highest absolute attribution score that connects to a child node already in the circuit is added. If the corresponding parent node is not yet part of the circuit, it is also included. This iterative process continues for $N$ steps, where $N$ represents the desired circuit size.

Due to the presence of attention edges, parent and child nodes are not always located at the same token position.

At the end of the process, it is guaranteed that there is a path from the logits node at the final position to every node in the graph. To ensure full connectivity, we iterate over each node in the circuit and remove any nodes, along with their corresponding edges, that are not connected to any embedding node through a path in the graph. 

\section{Schema: Further Details}\label{ap:schema generation}
To generate a schema, we sample 3 groups of 5 examples each from the dataset. For each group, we ask the LLM to generate a separate schema.\footnote{We do not provide the LLM with any few-shot examples to avoid influencing its decisions on  defining the spans.} This process produces 3 candidate schema.
Next, we present the LLM with all 15 examples and the 3 candidate schemas, asking it to create a single unified schema. We test the unified schema using the LLM by iterating over all examples, and checking whether it can apply the schema in a valid manner to each.
If the output is invalid for a given example, we point out why to the LLM and ask it to try again. After three failed attempts, we move to the next example. While this process can identify most errors, it is not infallible and may provide false positives.

If the LLM fails to apply the schema correctly to $\geq 20\%$ of the examples, we consider the schema invalid, inform the model of the issues, and restart the schema generation process. The process ends once the schema can be successfully applied to at least $80\%$ of the examples.
See Appendices~\ref{ap:schema application}~and~\ref{ap:schema-validation} for more details on the process of applying and validating schemas, as well as an error analysis.

\subsection{Saliency scores} \label{ap:mask-creation}
We have also explored the following methods to determine the importance of each token position:

\begin{enumerate}
    \item \textbf{Input Attribution:}  
    This method involves patching the embedding of each input token individually and measuring the importance of each position based on its impact on a downstream metric.

    \item \textbf{Aggregated Node Attribution Scores:}  
    The importance of a position is derived from the significance of its components. While edge attribution patching could theoretically be used for each example to identify important components, this approach is computationally expensive. Instead, we propose using \textit{Node Attribution Patching}, which uses a linear approximation to estimate node importance rather than edge importance, significantly reducing computation time. This method efficiently calculates the importance of each node at every layer and position. By aggregating attribution scores for all nodes at each position, we estimate the overall importance of every position. 
\end{enumerate}

For both methods we used mean ablation values derived from "The Pile" dataset \cite{gaopile}. We evaluated both methods for schema generation and observed that the resulting schemas closely resembled those produced by the gradient-based method. However, a significant drawback of these approaches is their reliance on a counterfactual dataset, which adds complexity. For this reason, we ultimately chose the gradient-based method as our preferred approach.

\begin{figure}[ht]
    \centering
    \includegraphics[width=\linewidth]{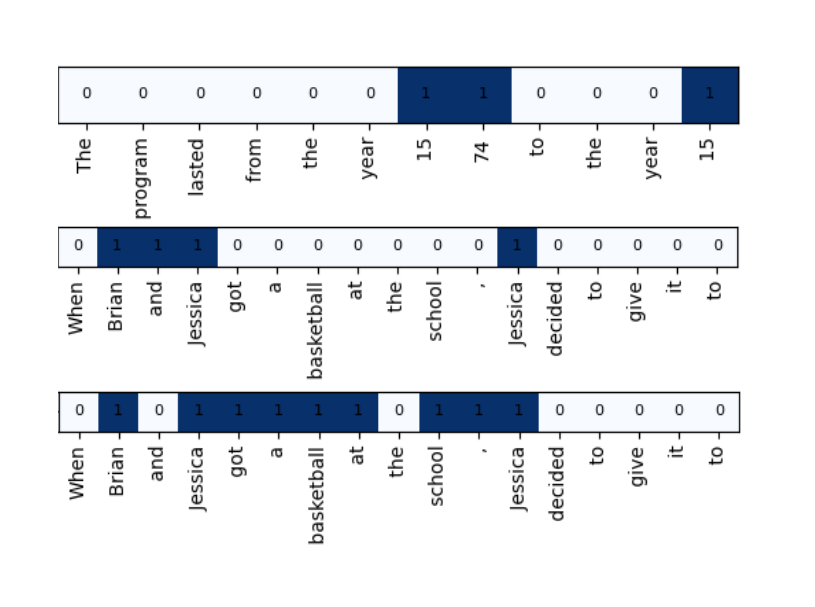}
    \caption{
The first example in Figure~\ref{fig:masks} is taken from the Greater-Than task and is generated using GPT2-small. Both the second and third examples are from the IOI dataset. The second mask is generated with GPT2-small, while the third is generated with LLaMA-3-8b. The highlighted positions are intended to capture the most influential positions that affect the model's predictions. }
    \label{fig:masks}
\end{figure}

\paragraph{Example.}  Figure~\ref{fig:masks} presents examples of the masks provided to the model. While not all masks highlight exactly the same token positions or token roles, we observe a consistent overall pattern across masks within the same dataset and model.

\subsection{Schema Evaluation} 
\label{ap:schema-validation}

We define the application of a schema to an example as valid if:

\begin{itemize}
    \item All spans specified by the schema are included, no extra spans are included, and the spans appear in the exact order defined in the schema.
    \item Each token in the prompt is assigned to only a single span, and the tokens within each span are a continuous sequence from the original prompt.
\end{itemize}
    
Note that empty spans are valid. To ensure that empty spans are justifiable, we initially treat them as invalid during the first iterations. If, after several attempts, a valid result cannot be obtained, we relax this requirement and allow for empty spans.

We automatically test all the above requirements. If an application is found to be invalid, the next attempt includes details in the prompt about the specific failures in the previous attempt.

\begin{table}[t]
    \centering
    \resizebox{\linewidth}{!}{
    \begin{tabular}{lllrr}
    \toprule
    & Task & Method & Valid & Correct \\
    \midrule
    \parbox[t]{2mm}{\multirow{6}{*}{\rotatebox[origin=c]{90}{\small{GPT2-small}}}} & \multirow{2}{*}{IOI ABBA} & LLM &  92.4\% &  88.3\%\\
    & & + Mask & 98.7\% &  86.8\%\\
    & \multirow{2}{*}{IOI BABA} & LLM & 98.0\% &  91.7\%\\
    & & + Mask & 93.7\% & 88.5\% \\
    & \multirow{2}{*}{Greater-Than} & LLM & 100\% & 100\% \\
    & &  + Mask & 100\% & 100\% \\
    \midrule
    \parbox[t]{2mm}{\multirow{8}{*}{\rotatebox[origin=c]{90}{\small{Llama-3-8B}}}} & \multirow{2}{*}{IOI ABBA} & LLM & 99.9\% & 96.0\% \\
    & & + Mask & 96.1\% & 96.3\% \\
    & \multirow{2}{*}{IOI BABA}  & LLM & 95.1\% & 81.5\% \\
    & & + Mask & 98.2\% & 92.3\% \\
    & \multirow{2}{*}{Winobias-$I$} & LLM & 98.5\% &  89.0\% \\
    & & + Mask & 96.5\% &  98.6\%\\
    & \multirow{2}{*}{Winobias-$II$} & LLM & 99.9\% & 97.9\% \\
    & & + Mask & 98.2\% & 95.4\% \\
    \bottomrule
    \end{tabular}}
    \caption{\textbf{Validity} is an automatic evaluation metric that tells us how many examples are usable for circuit discovery. \textbf{Correctness} is a human evalation metric that tells us whether the schema were applied in a way that a human agrees with. By definition, the human schema have 100\% correctness.}
    \label{tab:schema-eval}
\end{table}

We observe (Table~\ref{tab:schema-eval}) that the generated schemas are largely valid, indicating that most examples can be used for circuit discovery.

Recall that we additionally define a correctness metric, which measures to what extent a human annotator agrees with the application of the schema. To measure this, we have a human manually apply the LLM-generated schema to each template; we then compare to what extent the LLM application matches that of the human. Correctness is measured partially: that is, for each example, we compute the fraction of spans that are labeled identically to the human, and take this fraction as the correctness score. We then average these fractions across examples. We observe that a human tends to agree with how the schema were applied, as indicated by high correctness scores in Table~\ref{tab:schema-eval}.\footnote{Note, however, that we use the run with the maximum validity across 3 runs. We show scores across random trials in Appendix~\ref{ap:schema-validation}.} Thus, the schemas score high on intrinsic measures of quality.

\begin{table}
    \centering
    \resizebox{\linewidth}{!}{
    \begin{tabular}{cllrrr}
        \toprule
        & Task & Method & Validity \#1 & Validity \#2 & Validity \#3 \\
        \midrule
        \parbox[t]{2mm}{\multirow{6}{*}{\rotatebox[origin=c]{90}{\small{GPT2-small}}}} & \multirow{2}{*}{IOI ABBA} & LLM & 48.1 \%& \textbf{92.4\%} & 88.8\%\\
        & & + Mask &88.5\% & \textbf{98.7\%} & 65.2\% \\
        & \multirow{2}{*}{IOI BABA} & LLM &87.2\% & 55.1\% & \textbf{98.0\%}\\
        & & + Mask & \textbf{98.4\%}&  87.5\%& 93.7\% \\
        & \multirow{2}{*}{Greater-Than} & LLM & 98.8\%& \textbf{100\%} &  \textbf{100\%} \\
        & & + Mask & \textbf{100\%} & \textbf{100\%} & \textbf{100\%}\\
        \midrule
        \parbox[t]{2mm}{\multirow{8}{*}{\rotatebox[origin=c]{90}{\small{Llama-3-8B}}}} & \multirow{2}{*}{IOI ABBA} & LLM & \textbf{99.9\%}& 96.7\% & 99.8\% \\
        & & + Mask &\textbf{95.1\%} & 88.1\% & 92.9\% \\
        & \multirow{2}{*}{IOI BABA} & LLM & \textbf{96.1\%}& 62.6\% & 77.9\% \\
        & & + Mask & 98.2\% & 96.9\% & 95.4\% \\
        & \multirow{2}{*}{Winobias-$I$} & LLM & \textbf{98.5\%} & 96.0\% &  77.6\% \\
        & & + Mask & \textbf{96.5\%}& 71.4\% & 95.3\%\\
        & \multirow{2}{*}{Winobias-$II$} & LLM & 76.6\%& \textbf{99.9\%}& 93.5\% \\
        & & + Mask & 58.7\% & \textbf{98.2\%} & 93.5\% \\
        \bottomrule
        \end{tabular}
    }
    \caption{In our main experiments, we run schema generation and application three times per method, and take the run with the highest validity score. Here, we show the validities for all three runs for each method. (Validity \#1 corresponds to the run used in the main paper.)}
    \label{tab:schema-eval-seeds}
\end{table}

\section{LLM Prompts} \label{ap:schema prompts}

\subsection{Schema Generation}
 
\lstset{
    basicstyle=\ttfamily\footnotesize, 
    breaklines=true,                  
    frame=single                      
}
Here is an example prompt we use to generate the schema:

\begin{lstlisting}
You are a precise AI researcher, and your goal is to understand how a language model processes a dataset by analyzing its behavior across different segments of prompts.  

To do this, you need to divide all prompts in the dataset into spans, where each span represents a meaningful part of the sentence.  

The aim is to split the prompts in the dataset systematically, allowing you to analyze the relationships between various parts of the sentence and support different types of model analysis.  

### Task: ###  

Your task is to define a schema---a structure that defines how to split all the examples in the dataset into meaningful spans.  

The schema defines how to divide all examples into the same set of spans! Even though the examples do not have the exact same tokens, they share a similar structure.  

All parts of each prompt should be assigned to a span, meaning the schema must provide a complete division of every prompt.  

### Input Format: ###  

1. **Tokens**: A list of tokens representing the example. Your task is to find a schema that defines how to divide this list into meaningful spans.  

2. **Mask**: A list of pairs in the format `[(token, value)]`, where a value of `1` indicates that the token is important and should be placed in its own span, separated from other tokens.  

### Instructions: ###  

1. Use syntactic and semantic rules to create a schema that defines how to divide all the examples in the dataset into meaningful spans.  

2. Use the Masks to create additional spans for any token marked as significant (`value = 1`). Each of these tokens should be placed in its own span.  

   **Note**: Apply this rule only if a specific token or token role is marked as important across many examples.  

3. If you think certain parts or tokens are crucial for the model's processing of the prompt, assign them to a separate span to highlight their importance.  

4. The spans should provide a complete division of the prompt, ensuring that every token is assigned to a span, and the spans should reflect the chronological structure of the prompt.  

5. The examples may vary, so you must define a schema that is not tailored to any specific example but can be applied consistently across all examples.  

### Goal: ###  

Given a set of examples, your goal is to define a schema---a structure that divides all examples into the same set of sub-spans.  

#### Return Format: ####  

Return a JSON object describing the schema.  

Each key in the dictionary should represent a span title (1-3 words), and the corresponding value should describe the tokens or segments assigned to that span.  

Provide a brief description of each span's role based on syntax, semantics, or another relevant aspect, but do not reference the Mask in the description.  

Provide a variety of examples in the descriptions to clarify the scope of each span.  

Assign a descriptive and unique span title (1-3 words) to each span. Avoid mentioning the Mask in the title (e.g., "Significant Token").  

Example format:  

```json
{
    "title": "description and examples"
}
### I will now provide you with 5 pairs of Tokens and a Mask.  

Follow the steps carefully, and return a JSON file in the correct format.  

---

**Example 0:**  

**Tokens:**  
`['Then', ',', ' Michael', ' and', ' Matthew', ' had', ' a', ' long', ' argument', ',', ' and', ' afterwards', ' Michael', ' said', ' to']`  

**Mask:**  
`[('Then', 0), (',', 0), (' Michael', 1), (' and', 0), (' Matthew', 1), (' had', 0), (' a', 0), (' long', 0), (' argument', 0), (',', 0), (' and', 0), (' afterwards', 1), (' Michael', 1), (' said', 0), (' to', 0)]`  

---

**Example 1:**  

**Tokens:**  
`['Then', ',', ' Jennifer', ' and', ' John', ' had', ' a', ' long', ' argument', ',', ' and', ' afterwards', ' Jennifer', ' said', ' to']`  

**Mask:**  
`[('Then', 0), (',', 1), (' Jennifer', 1), (' and', 1), (' John', 1), (' had', 0), (' a', 0), (' long', 0), (' argument', 0), (',', 0), (' and', 0), (' afterwards', 1), (' Jennifer', 1), (' said', 0), (' to', 0)]`  

---

**Example 2:**  

**Tokens:**  
`['Then', ',', ' Michael', ' and', ' William', ' had', ' a', ' long', ' argument', ',', ' and', ' afterwards', ' Michael', ' said', ' to']`  

**Mask:**  
`[('Then', 0), (',', 1), (' Michael', 1), (' and', 0), (' William', 1), (' had', 0), (' a', 0), (' long', 0), (' argument', 0), (',', 0), (' and', 0), (' afterwards', 1), (' Michael', 1), (' said', 0), (' to', 0)]`  

---

**Example 3:**  

**Tokens:**  
`['Then', ',', ' Jessica', ' and', ' Elizabeth', ' went', ' to', ' the', ' office', '.', ' Jessica', ' gave', ' a', ' drink', ' to']`  

**Mask:**  
`[('Then', 0), (',', 0), (' Jessica', 1), (' and', 1), (' Elizabeth', 1), (' went', 0), (' to', 0), (' the', 0), (' office', 0), ('.', 0), (' Jessica', 1), (' gave', 0), (' a', 0), (' drink', 0), (' to', 0)]`  

---

**Example 4:**  

**Tokens:**  
`['Then', ',', ' Kimberly', ' and', ' Michelle', ' had', ' a', ' long', ' argument', '.', ' Afterwards', ' Kimberly', ' said', ' to']`  

**Mask:**  
`[('Then', 0), (',', 0), (' Kimberly', 1), (' and', 0), (' Michelle', 1), (' had', 0), (' a', 0), (' long', 0), (' argument', 0), ('.', 0), (' Afterwards', 0), (' Kimberly', 1), (' said', 0), (' to', 0)]`  

---

### Guidelines: ###  

1. Ensure the schema assigns each part in the prompt to a span.  

2. The final token in every prompt must always be placed in its own span.  

3. The final token is always placed in the last span.  

4. Any token marked with a value of `1` is considered important and should be placed in its own span in the schema.  

5. Create spans that can be applied to all the examples.  

6. Ensure spans are relevant and consistent across the entire dataset.  
\end{lstlisting}

An example of the models' response:

\begin{lstlisting}
Based on the given examples and instructions, I've created a schema that divides the prompts into meaningful spans. 

Here's the JSON object describing the schema:

{
  "Initial Time Marker": "Represents the opening temporal indicator, e.g., 'Then'",

  "First Person": "Contains the name of the first person mentioned in the prompt",

  "Conjunction": "Represents the coordinating conjunction connecting the two people, typically 'and'",

  "Second Person": "Contains the name of the second person mentioned in the prompt",

  "Main Event": "Describes the primary action or event involving both people, e.g., 'had a long argument', 'went to the office'",

  "Punctuation": "Represents the punctuation separating the main event from the following action",

  "Time Transition": "Indicates a shift in time, typically 'afterwards' or similar phrases",

  "Active Person": "Contains the name of the person performing the subsequent action",

  "Secondary Action": "Describes the action taken by the active person after the main event, e.g., 'said', 'gave'",

  "Final Preposition": "The last word of the prompt, typically a preposition like 'to'"
}

This schema provides a consistent structure that can be applied to all the given examples and similar prompts. It accounts for the important tokens (marked with 1 in the mask) by assigning them to individual spans. The schema also ensures that every part of the prompt is assigned to a span, and the final token is always placed in its own span.
\end{lstlisting}

\subsection{Schema Application} \label{ap:schema application}

Given a schema, we call Claude-3.5-Sonnet to apply the schema on each example separately. Here is an example call:

\begin{lstlisting}
You are an assistant that splits tokens into spans based on a given schema.
Schema: {
    'Temporal Context': 'Words or phrases that set the initial time frame or sequence of events',
    'Primary Subjects': "Names of the main characters involved in the first action, typically connected by 'and'",
    'Initial Action': 'Describes the primary action or state of the subjects, including any relevant location',
    'Transition': 'Punctuation or words that separate the initial context from the subsequent action',
    'Active Subject': 'The name of the character performing the secondary action',
    'Secondary Action': 'The verb phrase describing the action performed by the active subject',
    'Object': 'The item or concept involved in the secondary action',
    'Final Preposition': "The last word in the prompt, typically 'to'"
    } 
        
Tokens: [
'While', ' Jason', ' and', ' Crystal', ' were', ' commuting', ' to', ' the', ' hospital', ',', ' Jason', ' gave', ' a', ' kiss', ' to'
]
    
Please split the tokens into the spans defined by the schema.
Return the spans as a JSON object where each key is a span name and the value is the list of tokens in that span.
Format: 
```json {"span title": []} ``` 

Ensure that:
    - All the spans are present.
    - Every token is assigned to a span.
    - No new spans are added.
    - Punctuation marks should be included in the spans. If no specific span is assigned to punctuation mark, it should be grouped with the preceding token.
    - Ensure that the last token is placed exclusively in the final span. 
    - The spans are in the correct order as in the schema.
    - The tokens in each span are a continuous segment of the full prompt.
    - The tokens are kept in the same order as they appear in the original prompt.
    - Don't remove spaces from tokens inside the list.
    - If a span has no tokens in a specific example, leave it empty, but still include the span for consistency across all examples.
\end{lstlisting}

An example of a response from Claude:

\begin{lstlisting}
Here's the JSON object with the tokens split into spans according to the given schema:

{
  "Temporal Context": ["While"],
  "Primary Subjects": [" Jason", " and", " Crystal"],
  "Initial Action": [" were", " commuting", " to", " the", " hospital"],
  "Transition": [","],
  "Active Subject": [" Jason"],
  "Secondary Action": [" gave"],
  "Object": [" a", " kiss"],
  "Final Preposition": [" to"]
}
\end{lstlisting}

We do not provide Claude with any few-shot examples, as we aim to keep this process fully automated. However, using human-crafted few-shot examples could improve the correctness of the application.

\section{Faithfulness Curves}\label{ap:faithfulness_curves}
\subsection{Results Across Schema Generation Trials}\label{ap:faitfulness-trials}
As mentioned in \S\ref{sec:schema-eval}, we run the entire pipeline three times for each task to ensure at least 90\% of the examples are valid. In \S\ref{sec:results}, we present the results from the trials with the highest validation scores. Here, we report results for all trials. 

\subsubsection{Greater-Than}
\begin{figure*}
    \centering
    \textbf{Greater-Than GPT2-small}

    \includegraphics[width=0.32\linewidth]{graphs/hard_faithfulness/main/greater_than_2_500_accuracy.pdf} \hfill 
    \includegraphics[width=0.32\linewidth]{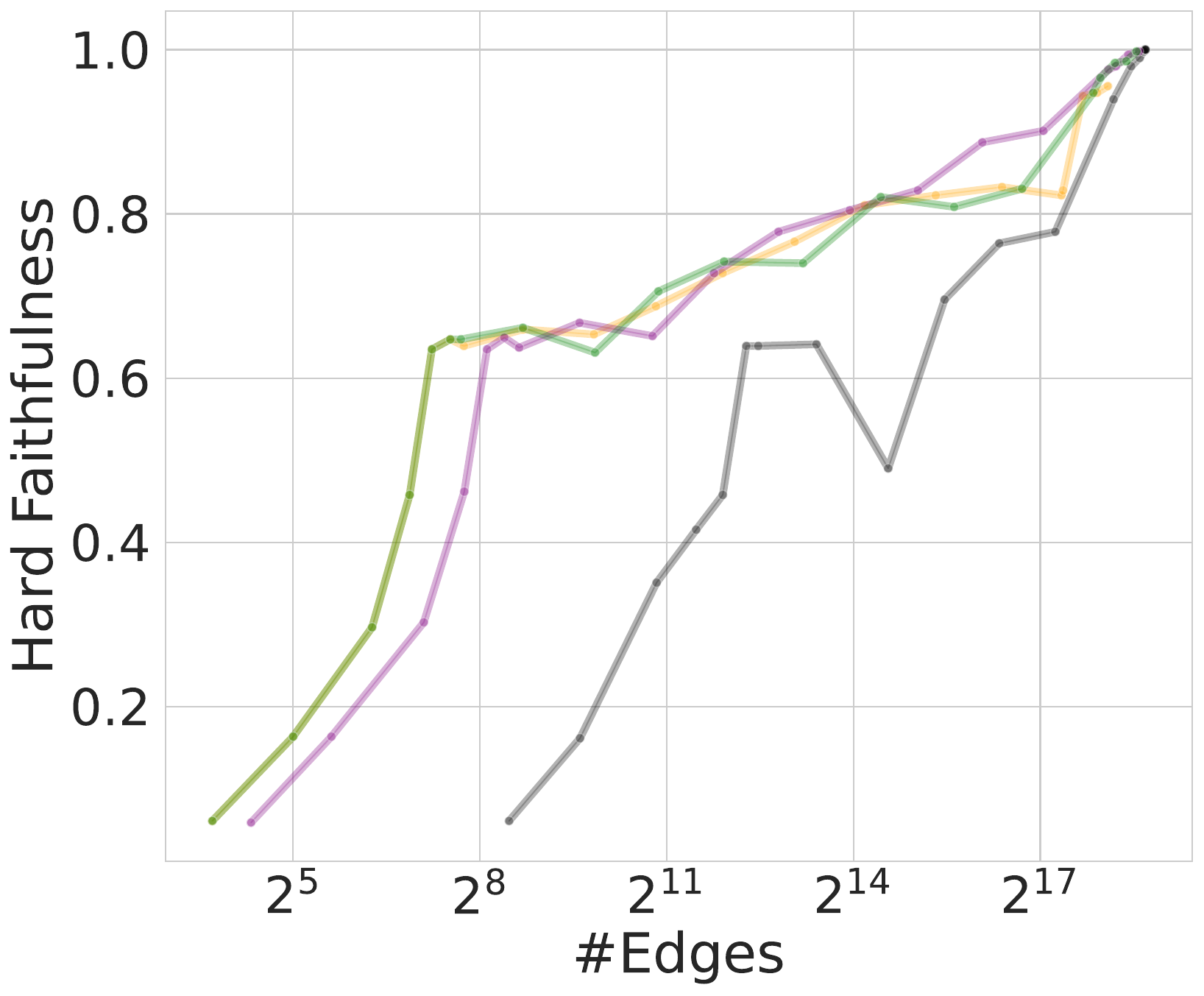} \hfill 
    \includegraphics[width=0.32\linewidth]{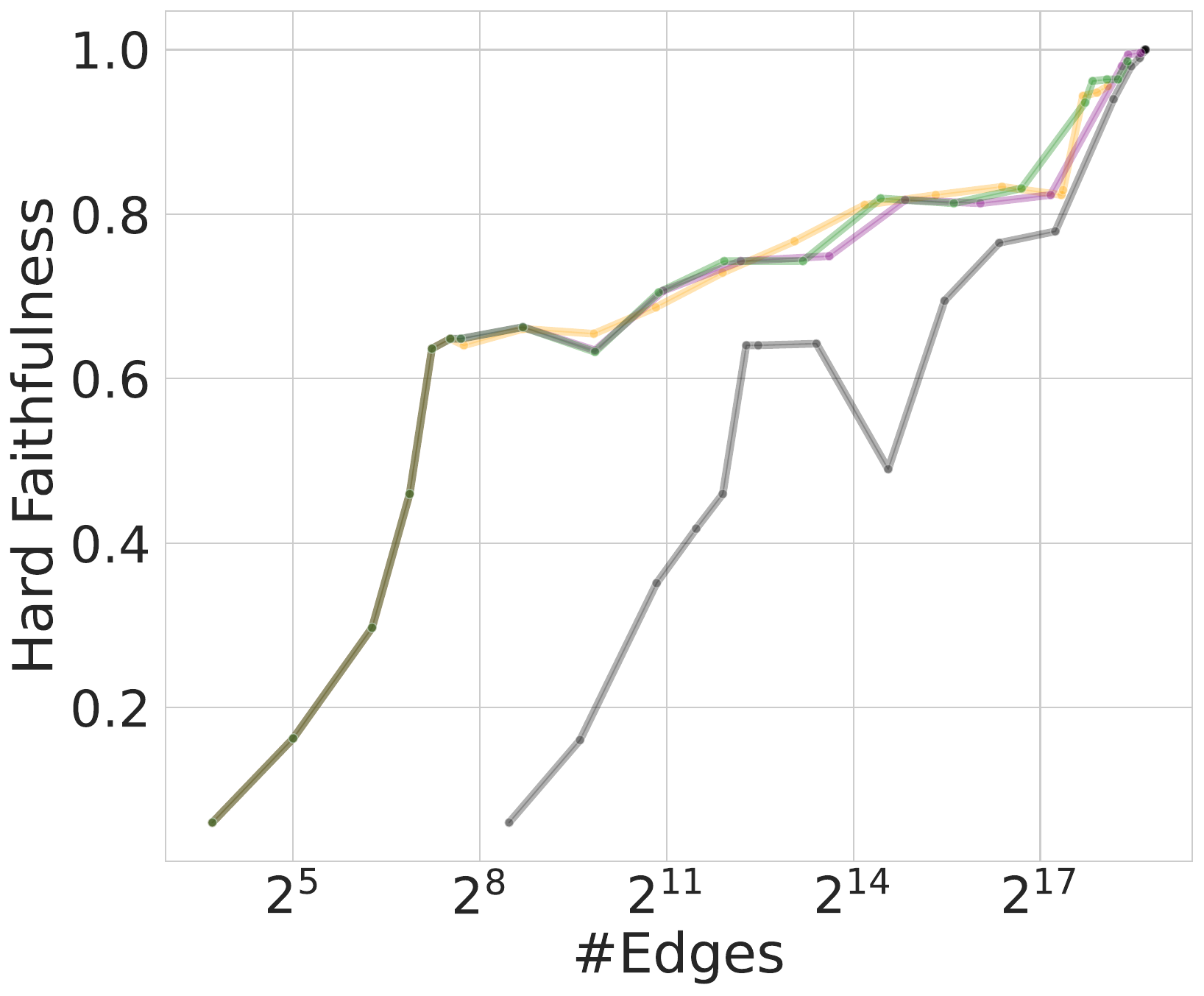} \hfill

    \vspace{0.05cm}
    \includegraphics[width=0.32\linewidth]{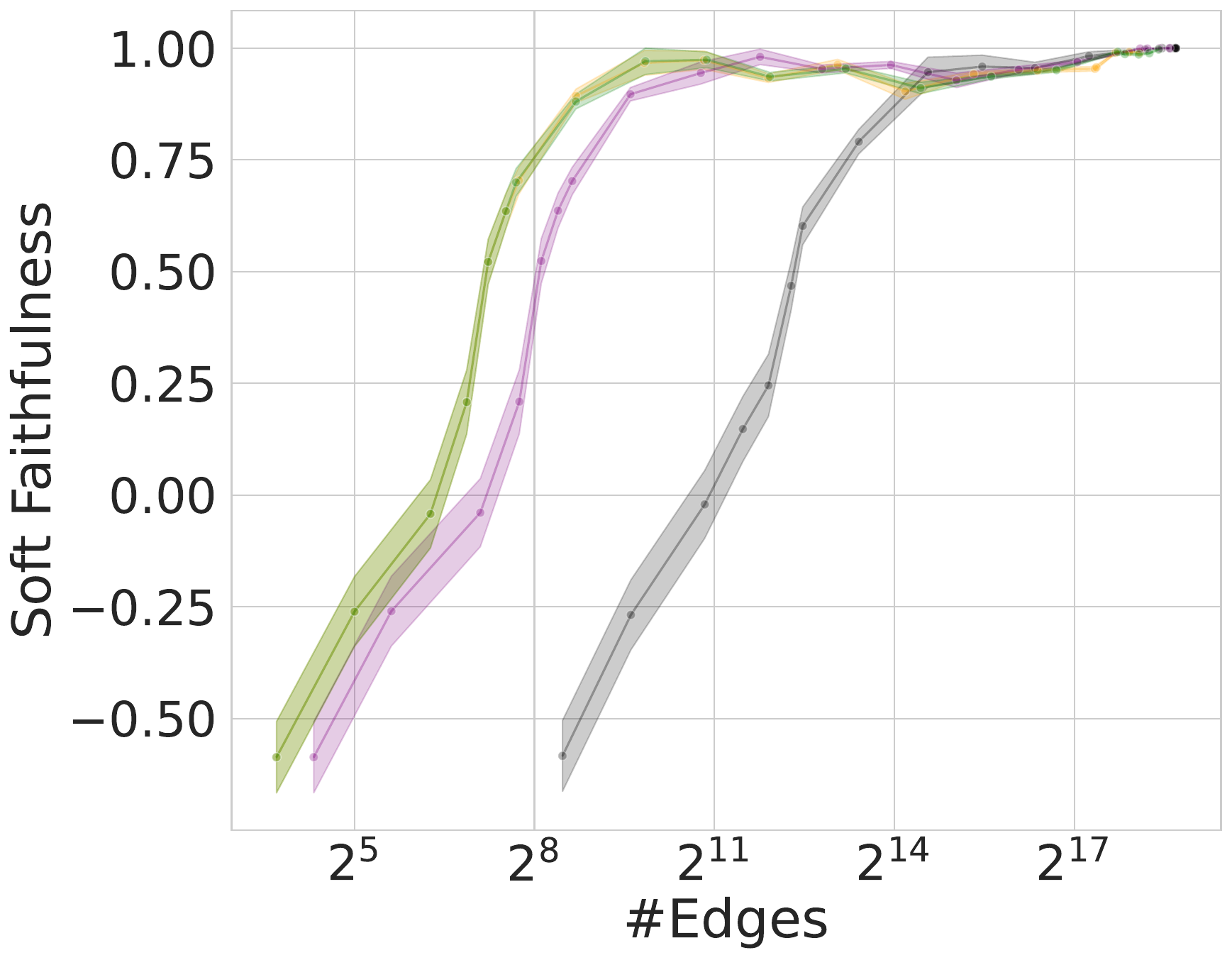} \hfill 
    \includegraphics[width=0.32\linewidth]{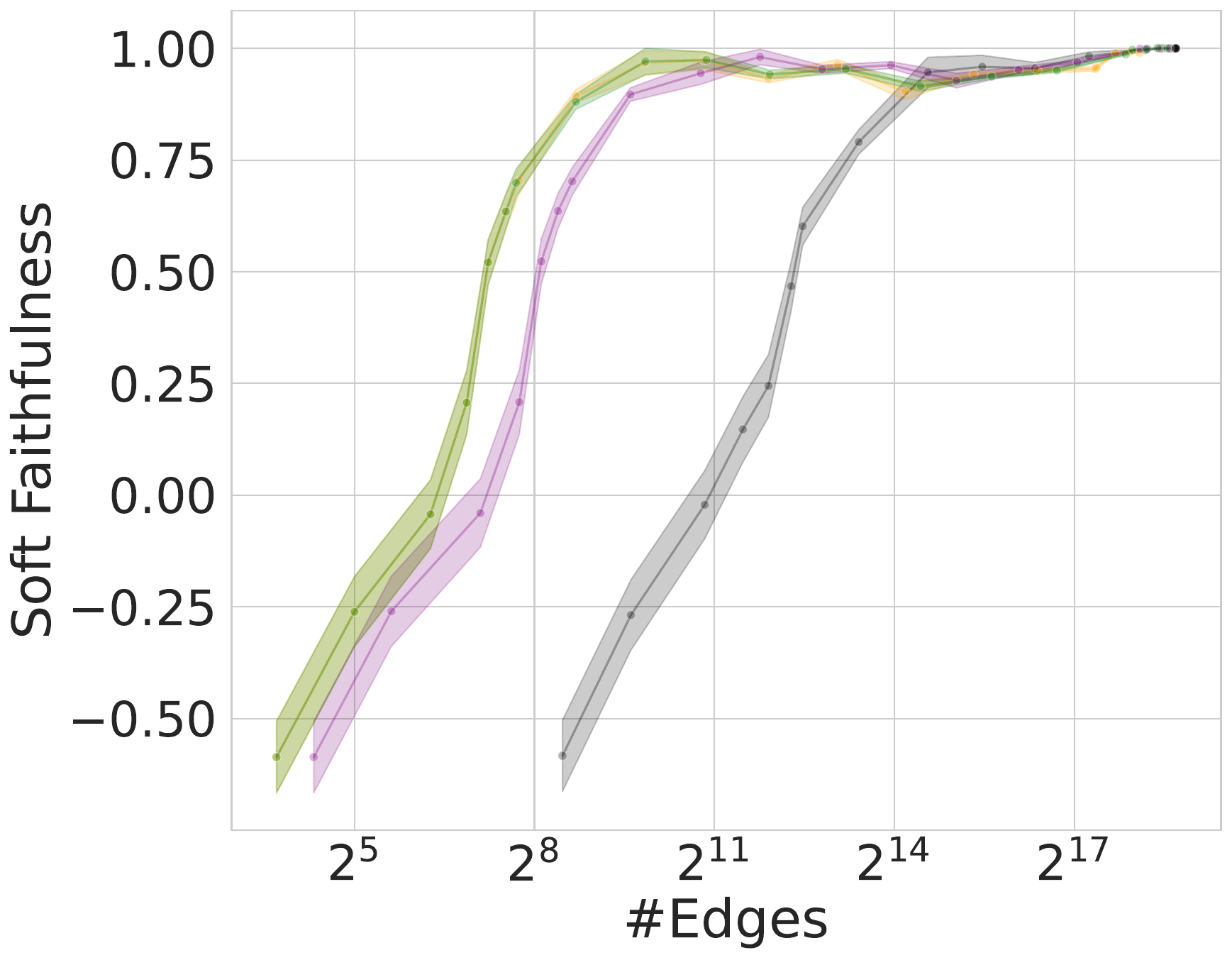} \hfill 
    \includegraphics[width=0.32\linewidth]{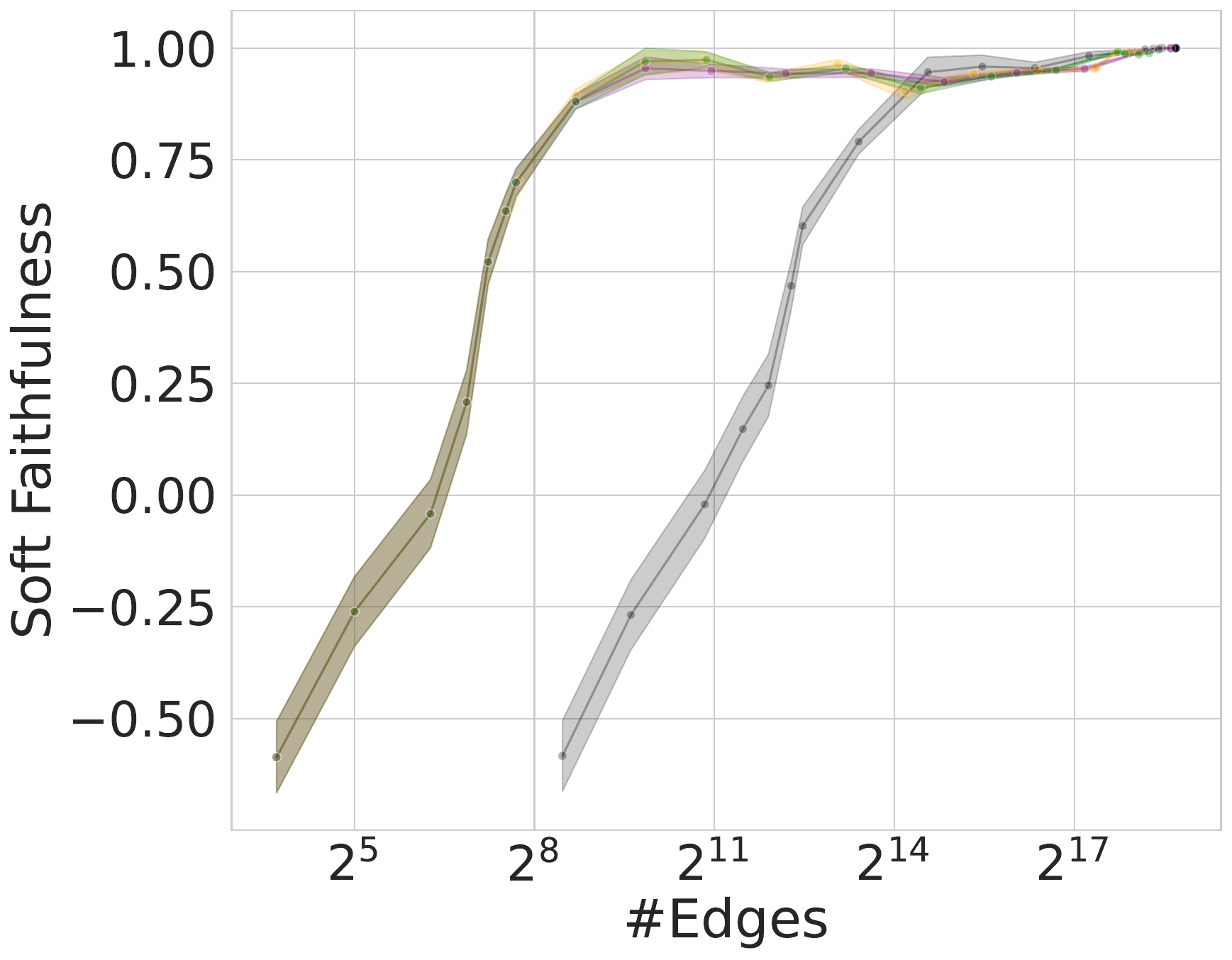} \hfill 
    \vspace{0.05cm}
    \includegraphics[width=0.6\linewidth]{graphs/legend_horizontal.pdf} \hfill 
\caption{Each column shows results for a single trial.}
\label{fig:faithfulness_all_gt}
\end{figure*}

Figures~\ref{fig:faithfulness_all_gt}, \ref{fig:faithfulness_all_ioi_gpt2}, and \ref{fig:faithfulness_all_ioi_llama} display the results for the Greater-Than and IOI tasks. Trends are highly consistent across trials, and are all similar to what we observe in the main paper. This is a trivial case where each word could reasonably be assigned its own span, however, so the following sections are more representative of the variance of this method on more realistic datasets.

\subsubsection{Winobias}
\label{ap:faithfulness_curves_wb}

\begin{figure*}
    \centering
    \textbf{Winobias I Llama-3-8b}

    \includegraphics[width=0.32\linewidth]{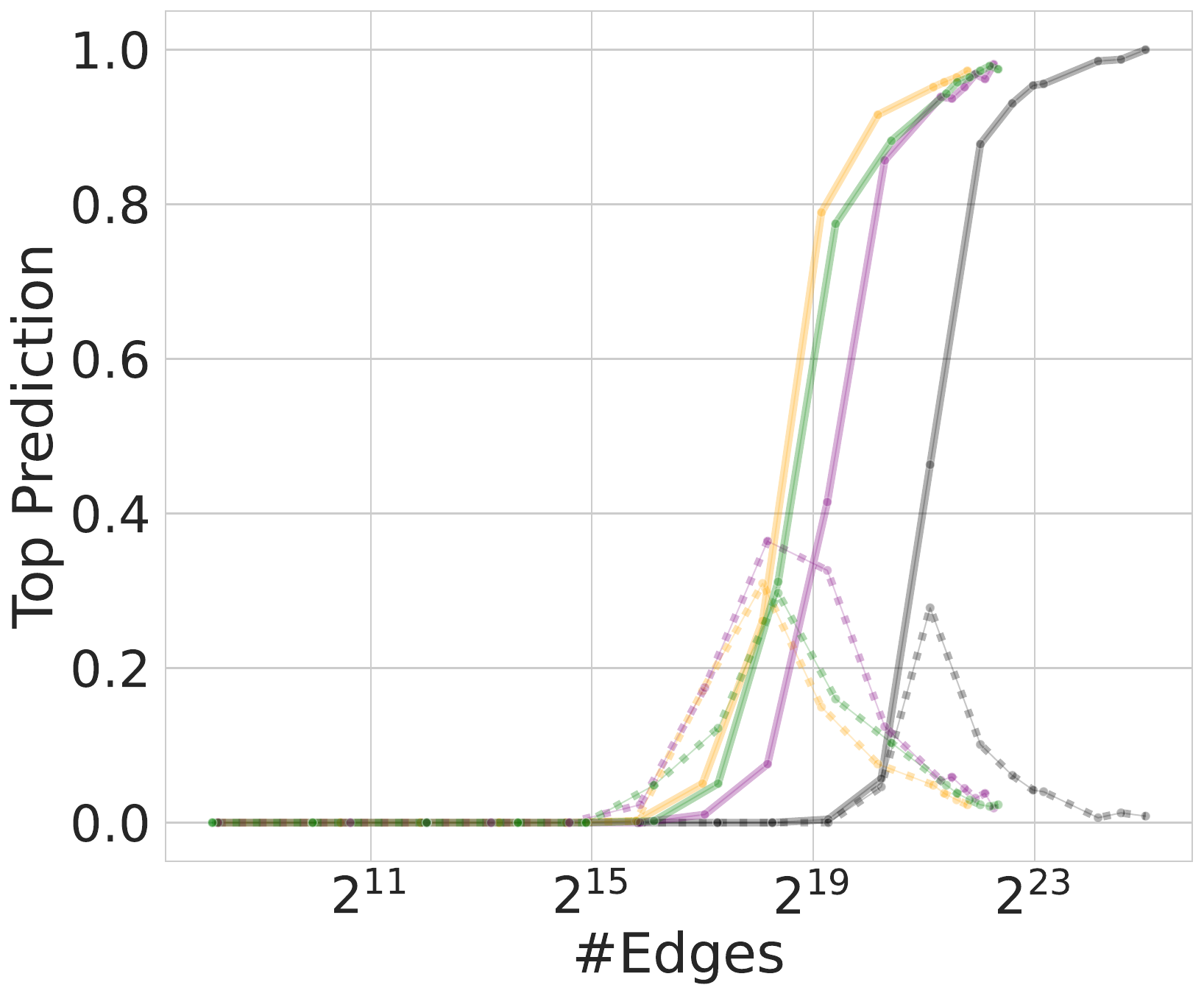} \hfill 
    \includegraphics[width=0.32\linewidth]{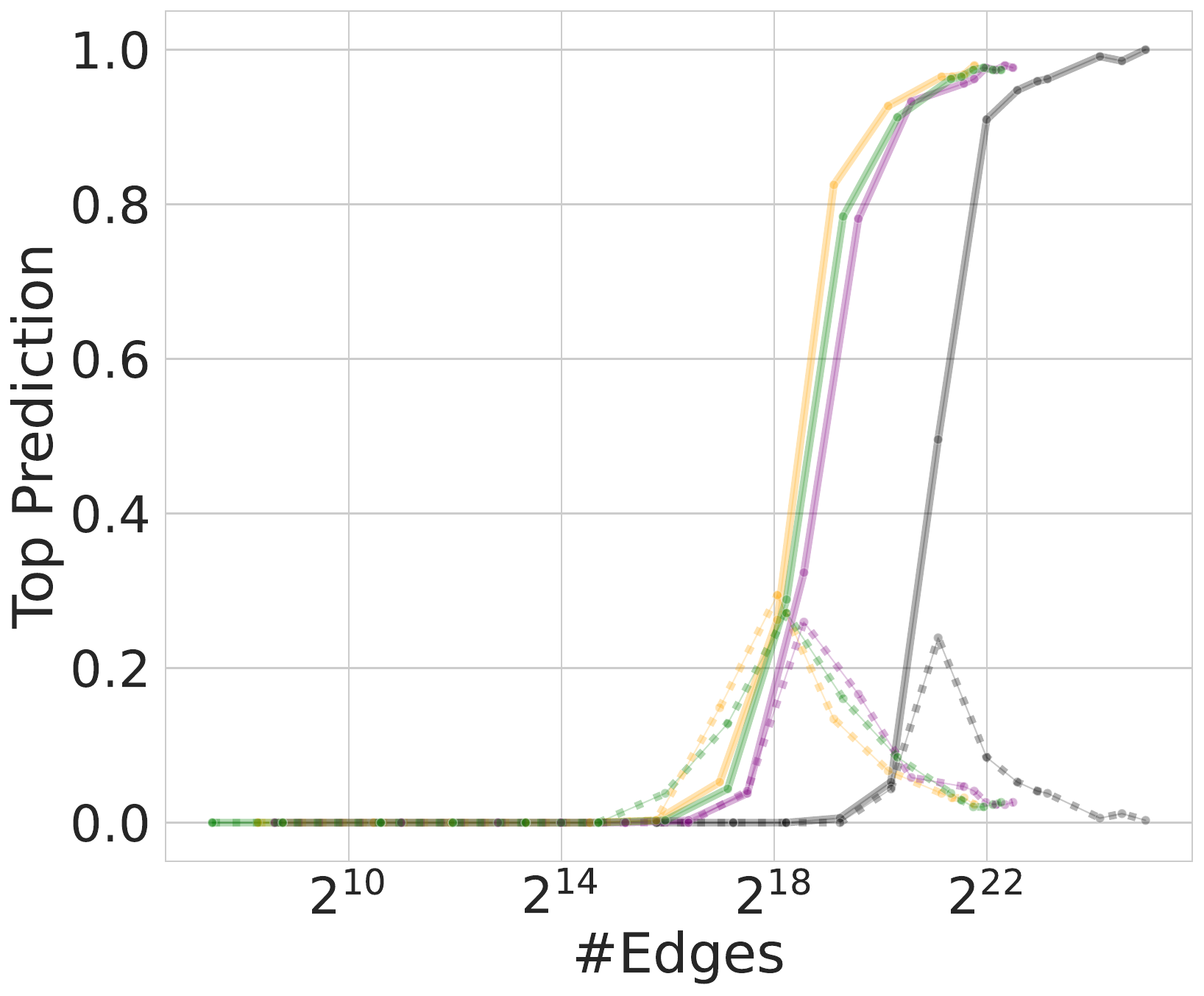} \hfill 
    \includegraphics[width=0.32\linewidth]{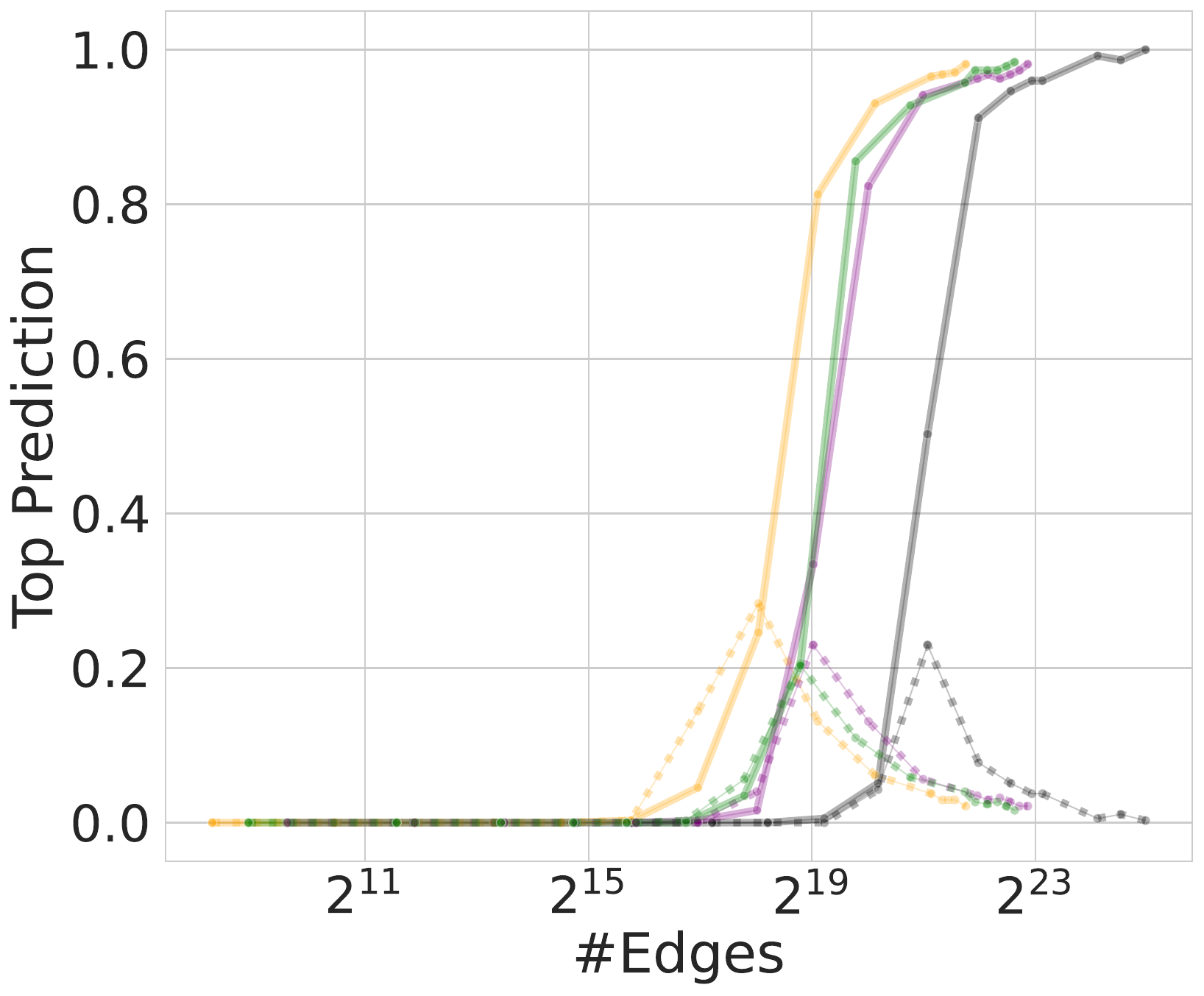} \hfill 

    \vspace{0.05cm}
     \includegraphics[width=0.5\linewidth]{graphs/legend_horizontal.pdf} \hfill

    \vspace{0.05cm}
    
    \includegraphics[width=0.32\linewidth]{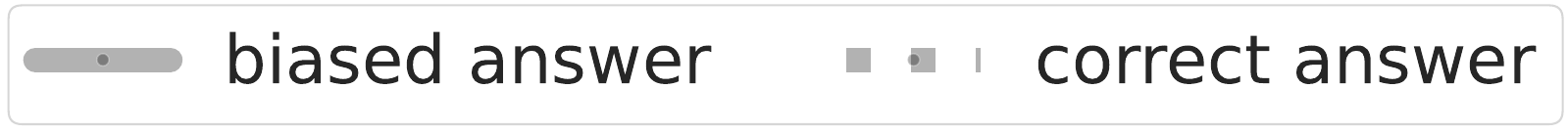} \hfill

    \vspace{0.05cm}
    \includegraphics[width=0.32\linewidth]{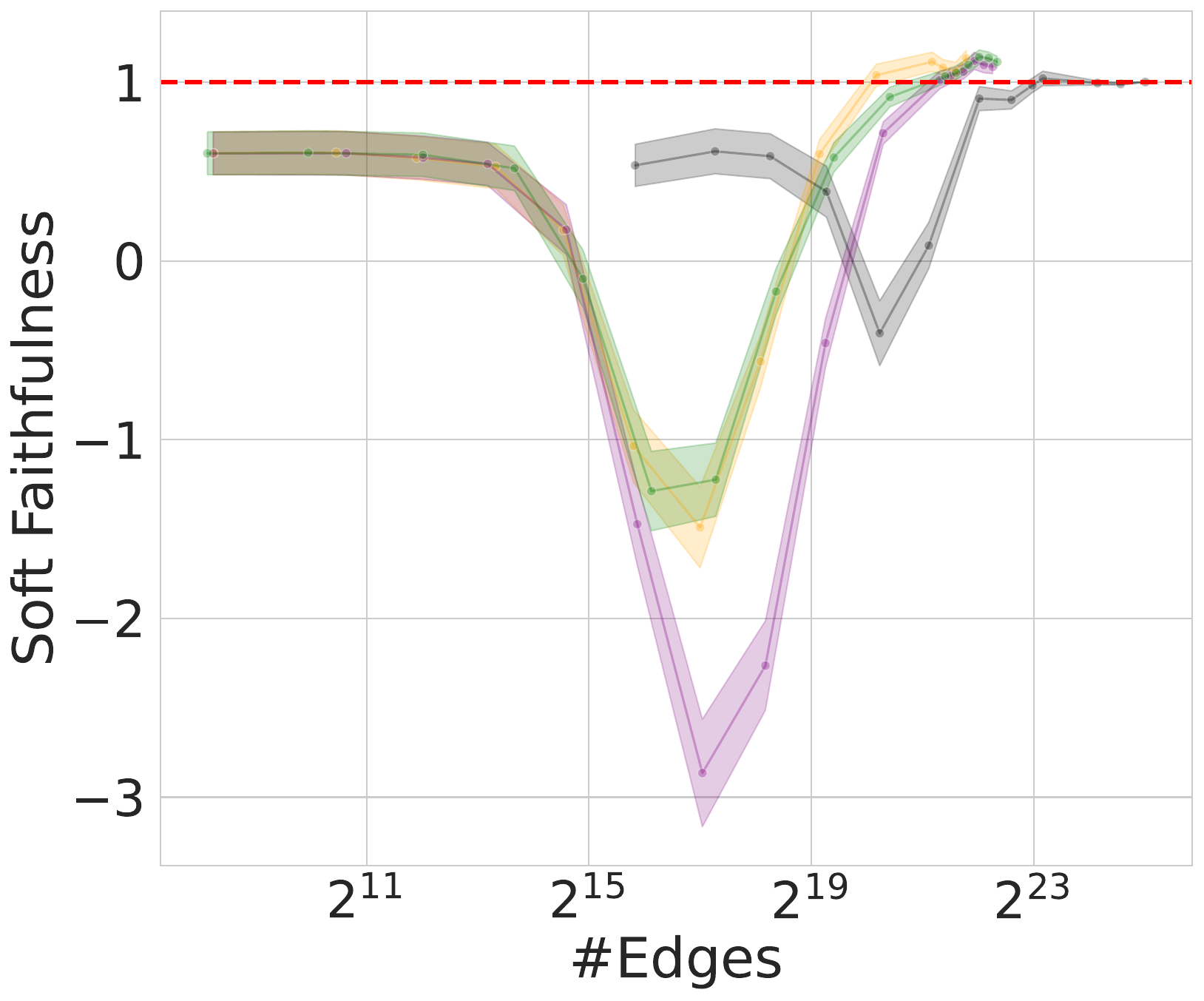} \hfill 
    \includegraphics[width=0.32\linewidth]{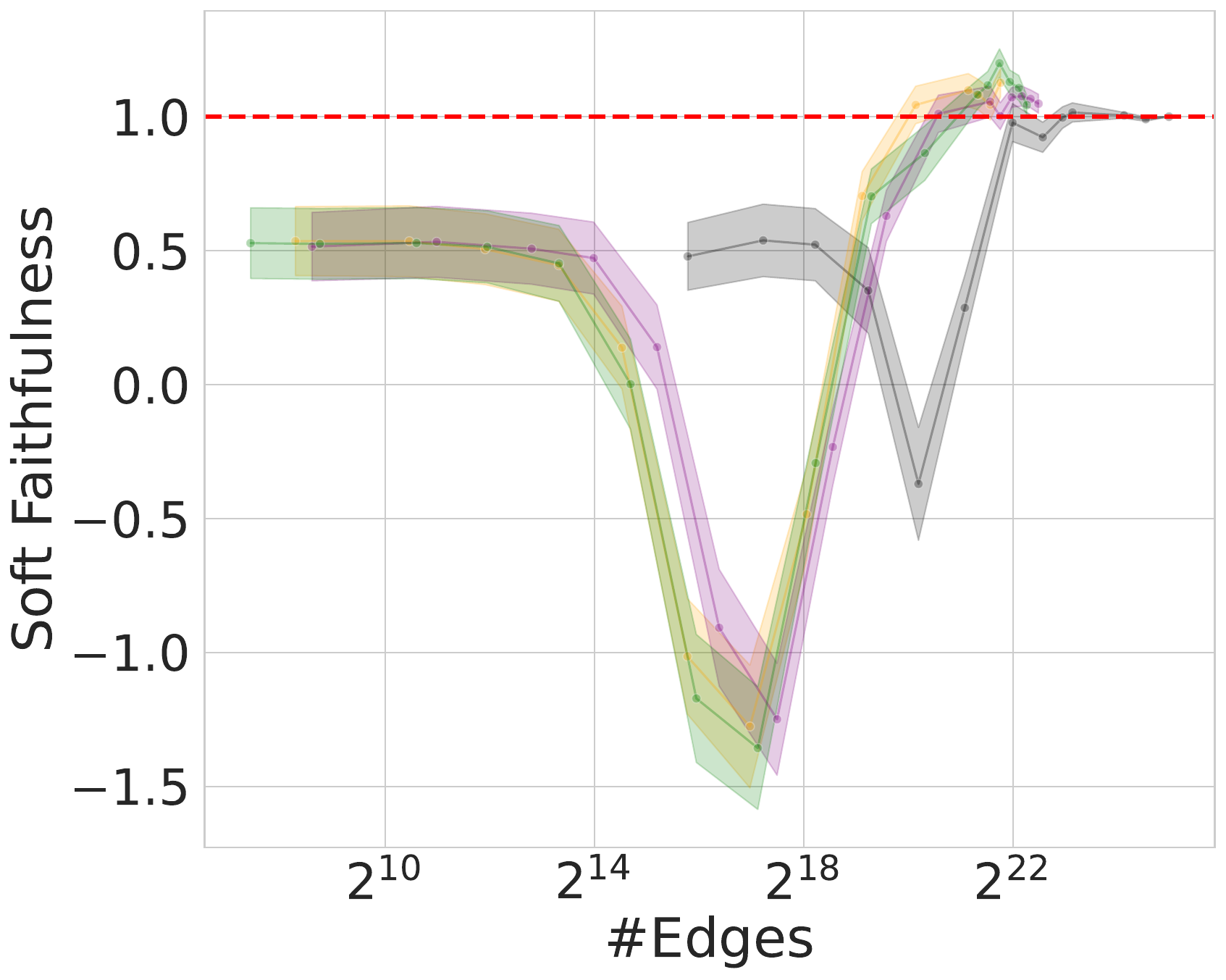} \hfill 
    \includegraphics[width=0.32\linewidth]{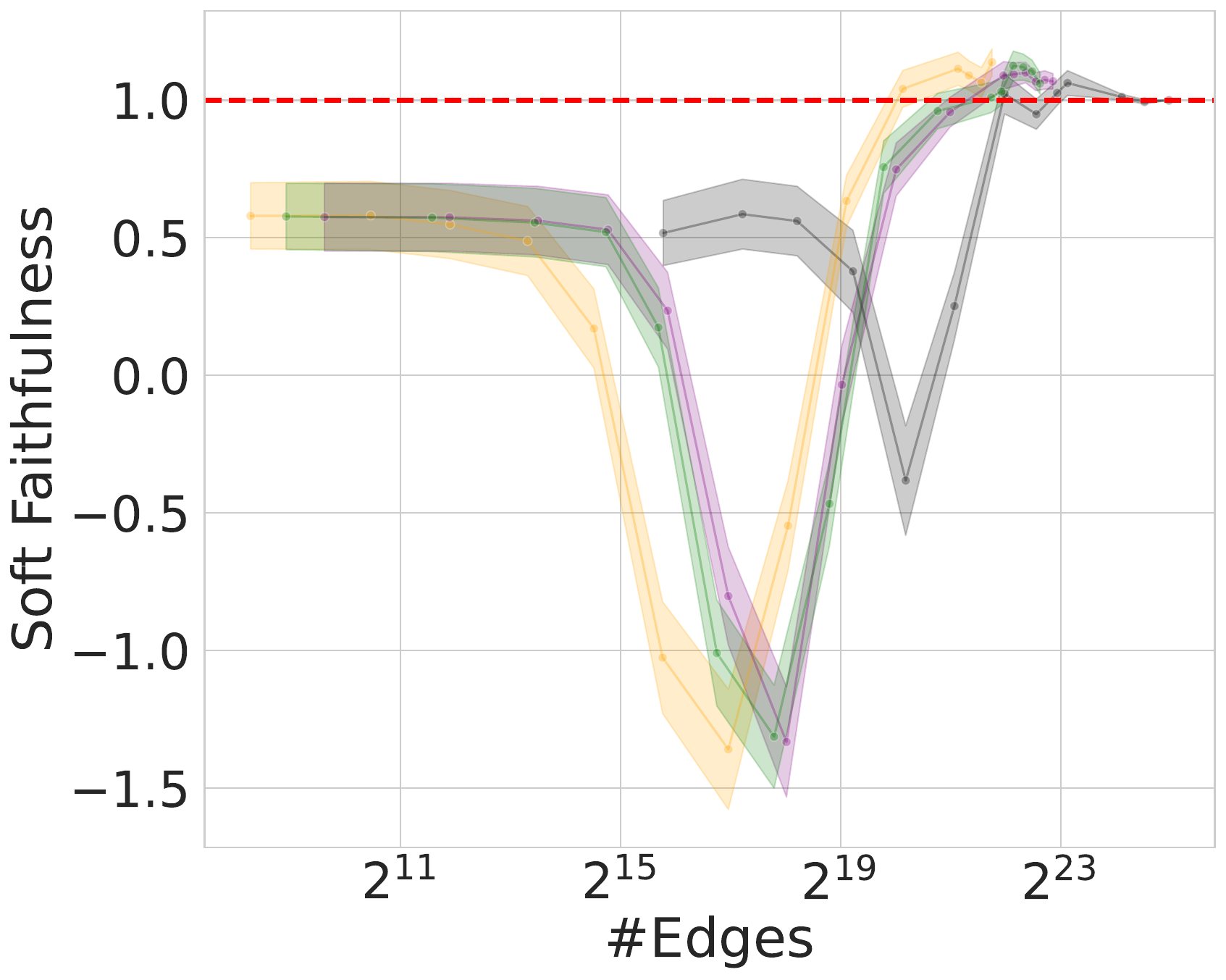} \hfill 

    \vspace{0.05cm}
     \includegraphics[width=0.5\linewidth]{graphs/legend_horizontal.pdf} \hfill 

     \textbf{Winobias II Llama-3-8b}

    \includegraphics[width=0.32\linewidth]{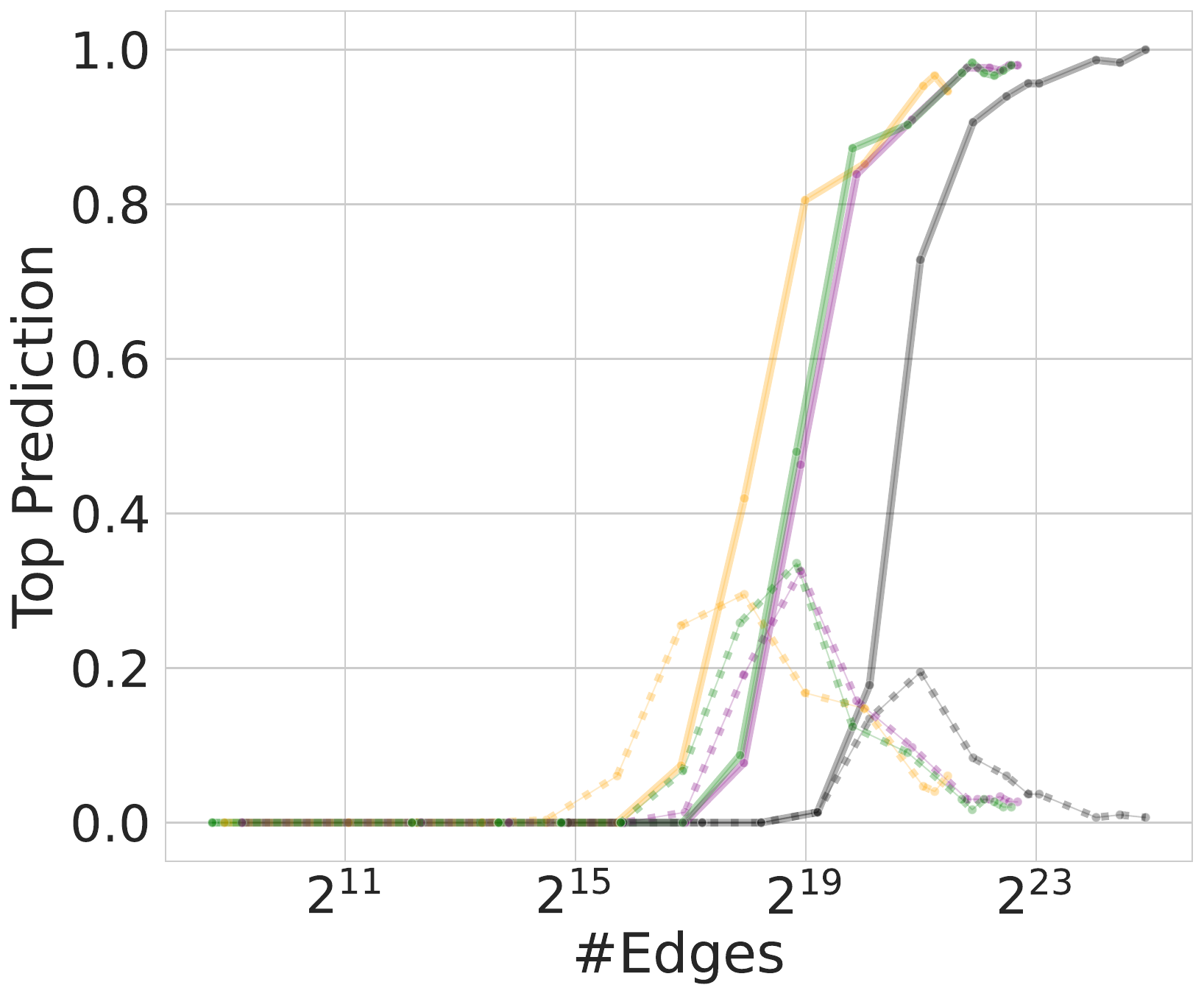} \hfill 
    \includegraphics[width=0.32\linewidth]{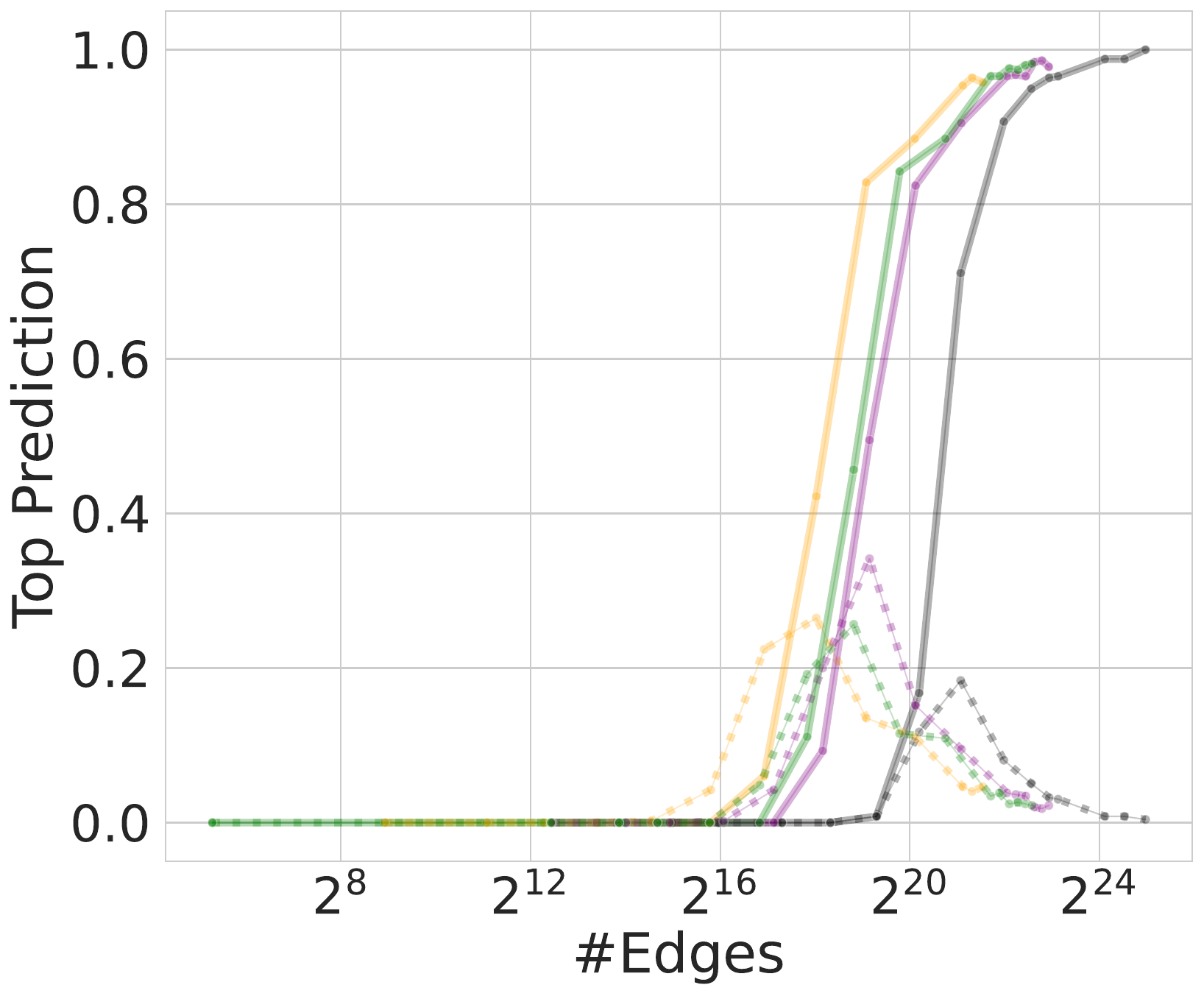} \hfill 
    \includegraphics[width=0.32\linewidth]{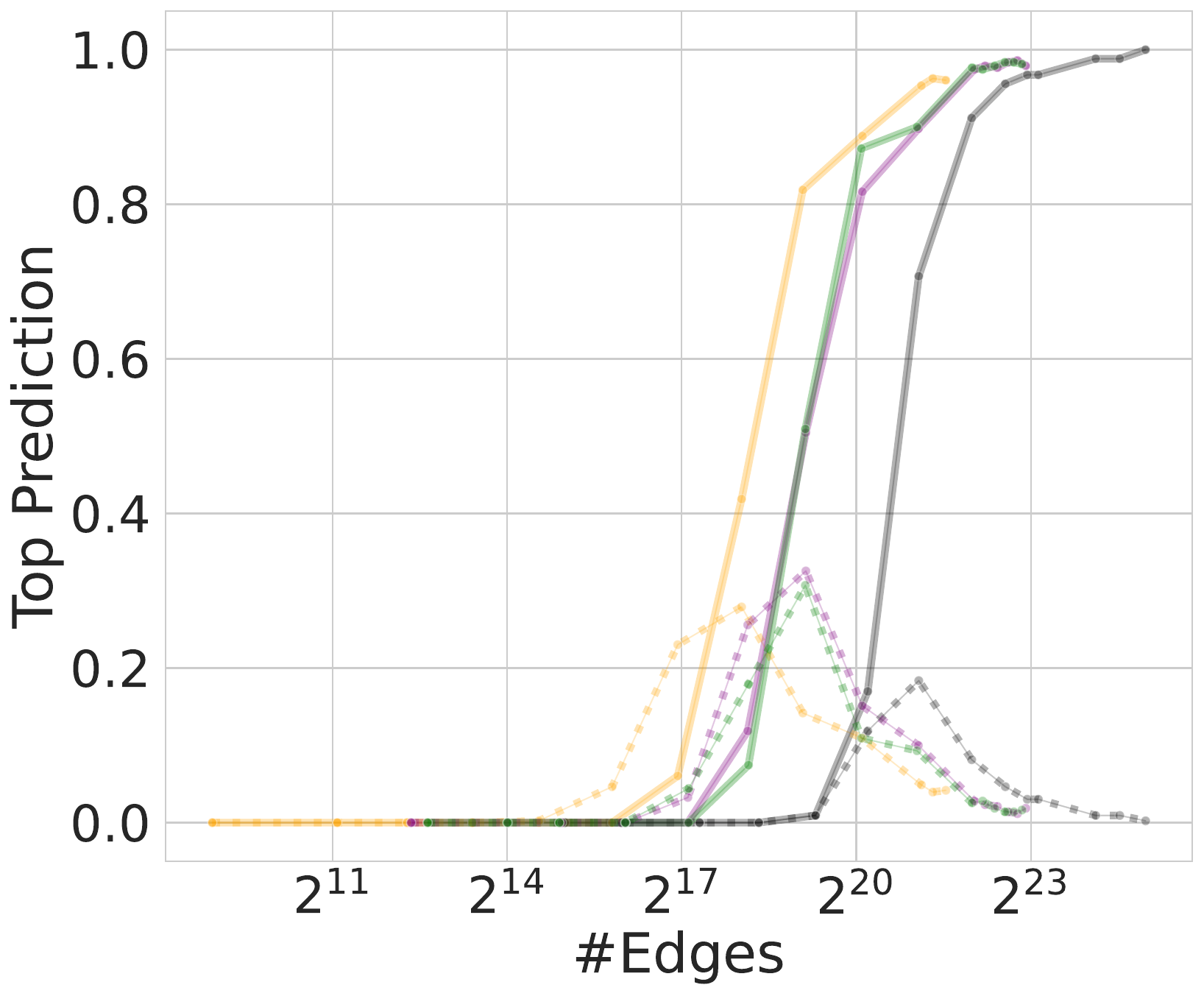} \hfill 

    \vspace{0.05cm}
     \includegraphics[width=0.5\linewidth]{graphs/legend_horizontal.pdf} \hfill

    \vspace{0.05cm}
    
    \includegraphics[width=0.32\linewidth]{graphs/legend_wb.pdf} \hfill

    \vspace{0.05cm}
    \includegraphics[width=0.32\linewidth]{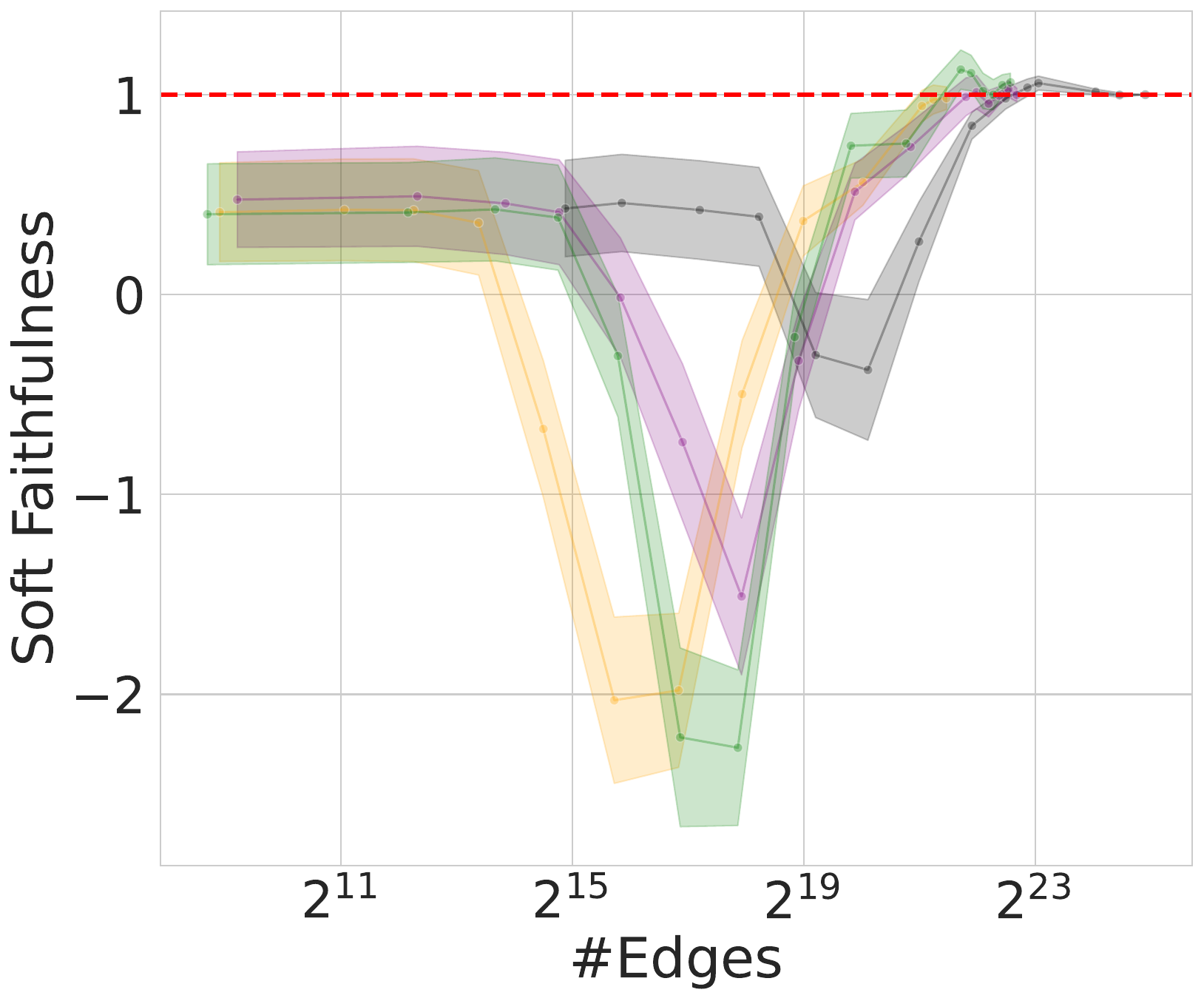} \hfill 
    \includegraphics[width=0.32\linewidth]{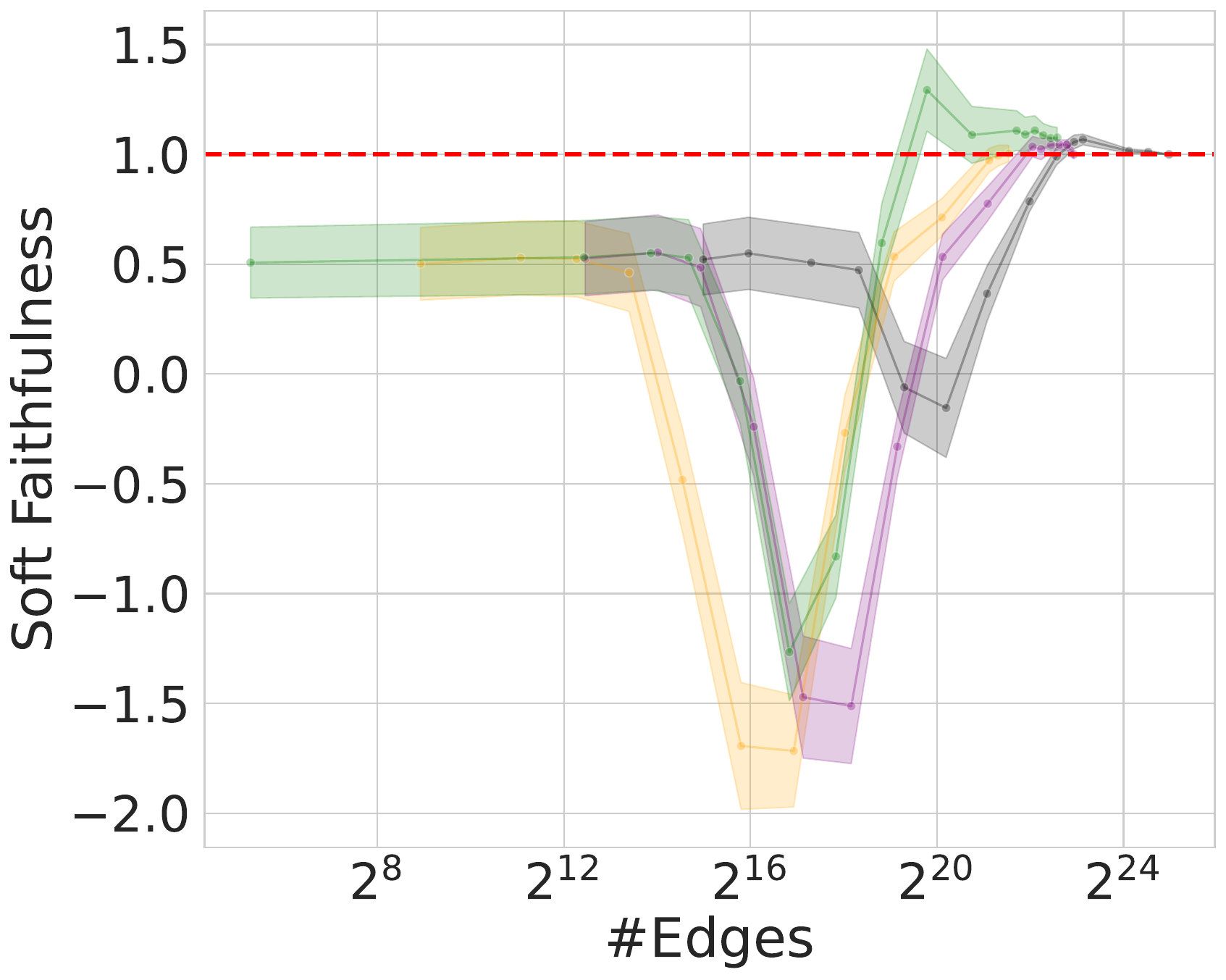} \hfill 
    \includegraphics[width=0.32\linewidth]{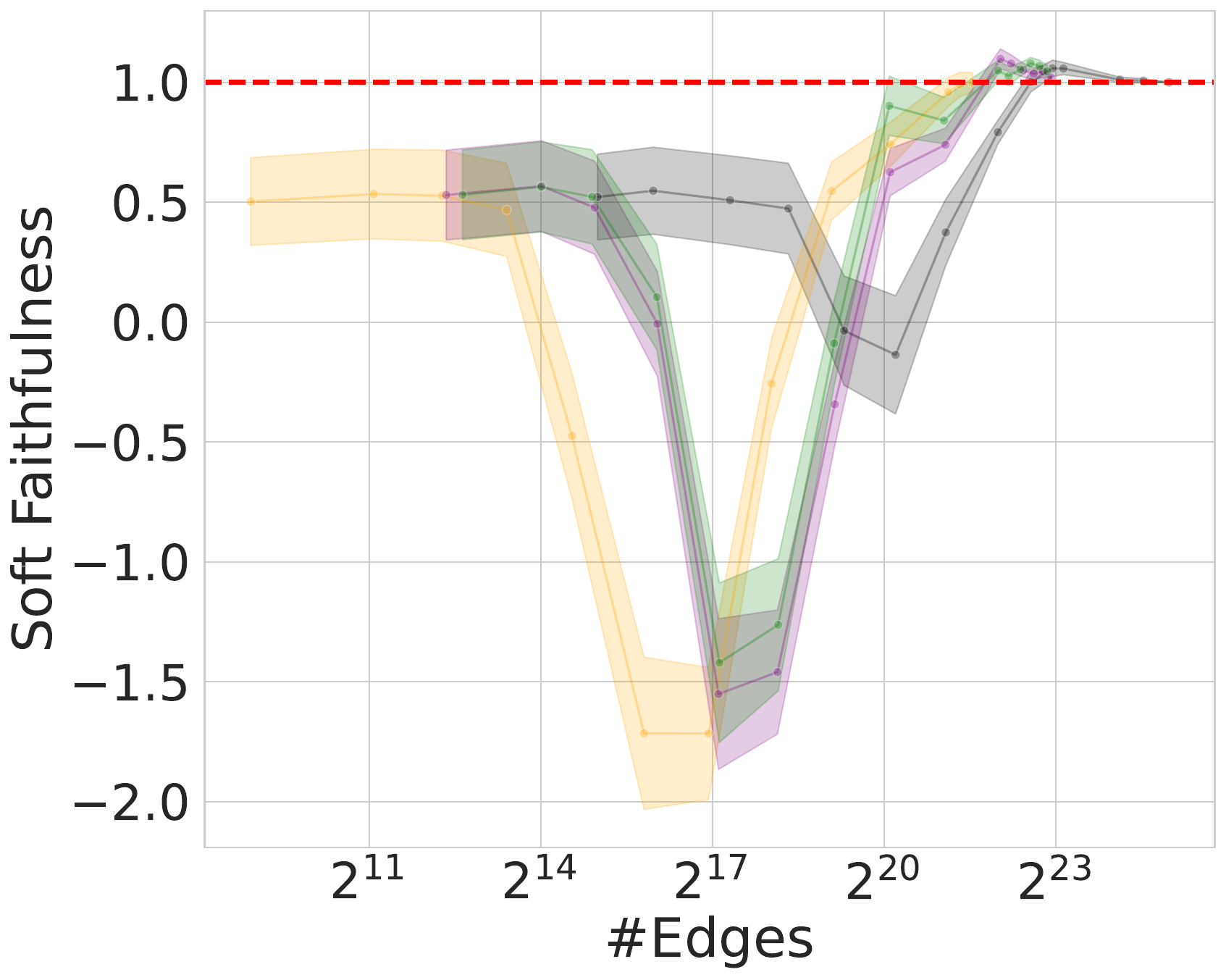} \hfill 
     \vspace{0.05cm}
     \includegraphics[width=0.5\linewidth]{graphs/legend_horizontal.pdf} \hfill

\caption{Winobias task results showing soft and hard faithfulness curves. Each column shows results for a single trial. The soft faithfulness curves initially drop significantly, suggesting the circuit assigns higher logits to the correct answer than to the incorrect, biased answer. The dotted lines in the hard faithfulness curves quantify this by showing the average percentage of cases where the circuit generates the correct answer, despite focusing on examples where the model predicts the biased answer. As the circuit size increases, the soft faithfulness curves rise, correlating with an increased percentage of biased predictions. This effect is more pronounced when token positions are differentiated.}
\label{fig:faithfulness_all_wb_llama}
\end{figure*}

In Figure \ref{fig:faithfulness_all_wb_llama}, we present the results for the Winobias task. The soft faithfulness curves across all schemas and trials initially exhibit a significant drop, suggesting that the circuit assigns higher logits to the correct answer compared to the incorrect, biased answer. To quantify this, the dotted lines in the hard faithfulness curves represent the average percentage of cases where the circuit generates the \textbf{correct answer}. This observation is non-trivial, as we specifically analyze examples where the model predicts the \textbf{biased} (and wrong) answer. Indeed, near the drop in the soft faithfulness curves, the models often predict the correct answer at a significant rate. However, as the circuit size increases, the trend reverses: the soft faithfulness curves increase, correlating with a higher percentage of biased predictions.
This effect becomes particularly pronounced when token positions are differentiated. One plausible explanation is that the circuit incorporates components that simultaneously influence both the correct and biased answers, reflecting the delicate balance between task-relevant and bias-inducing factors. As component analysis lies outside the scope of this study, future research could further investigate this phenomenon.

In general, the human schema achieve the best top-prediction scores. The shape of the faithfulness curves makes it difficult to determine a best method, but the human schema tends to produce a curve that resembles the others, but left-shifted. This suggests that it is picking up on important components before the other methods. 

\subsubsection{Indirect Object Identification}\label{ap:faithfulness_curves_ioi}
\begin{figure*}
    \centering
    \textbf{IOI-ABBA GPT2-small}

    \includegraphics[width=0.32\linewidth]{graphs/hard_faithfulness/main/ioi_ABBA_gpt2_2_452_accuracy.pdf} \hfill 
    \includegraphics[width=0.32\linewidth]{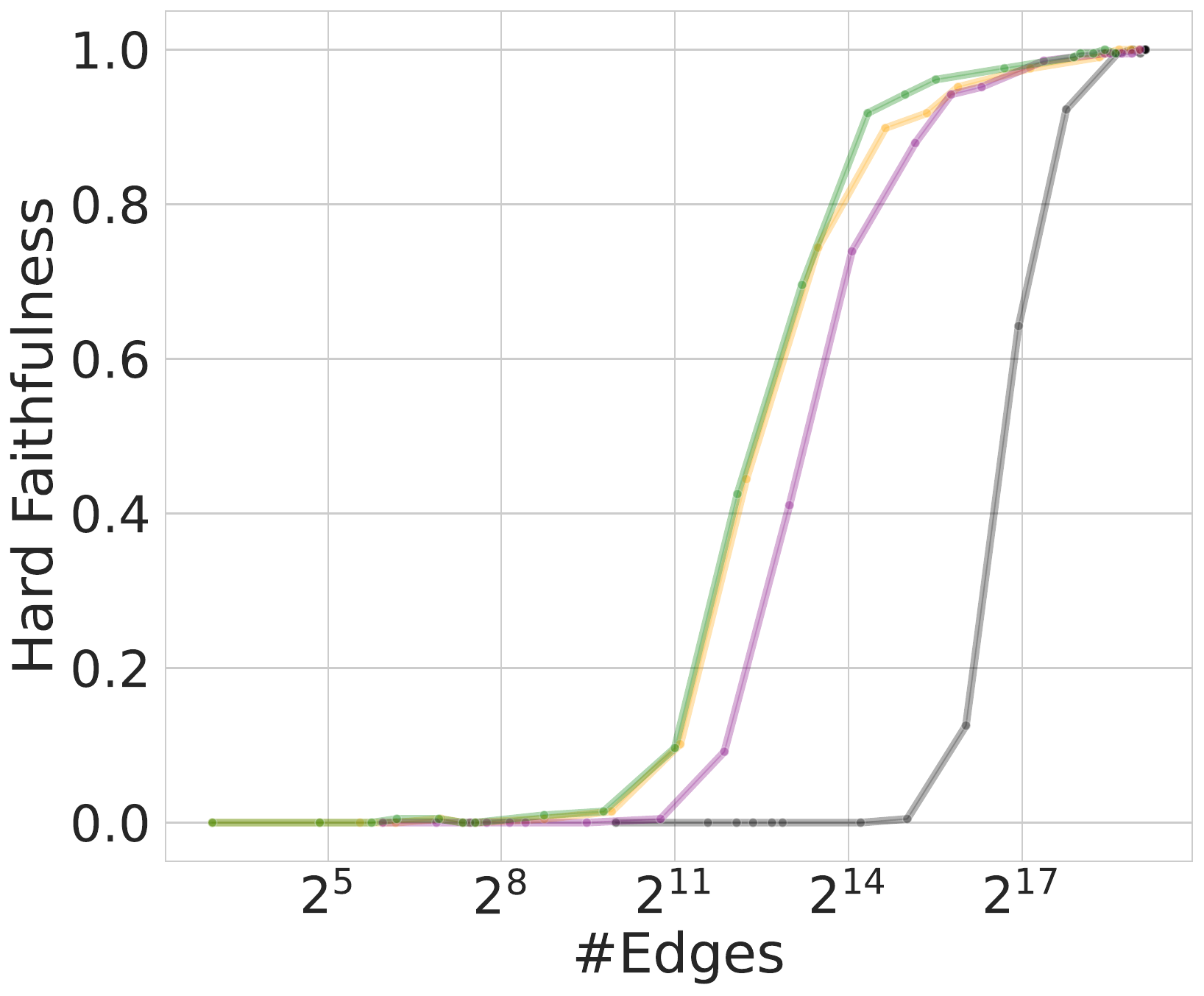} \hfill 
    \includegraphics[width=0.32\linewidth]{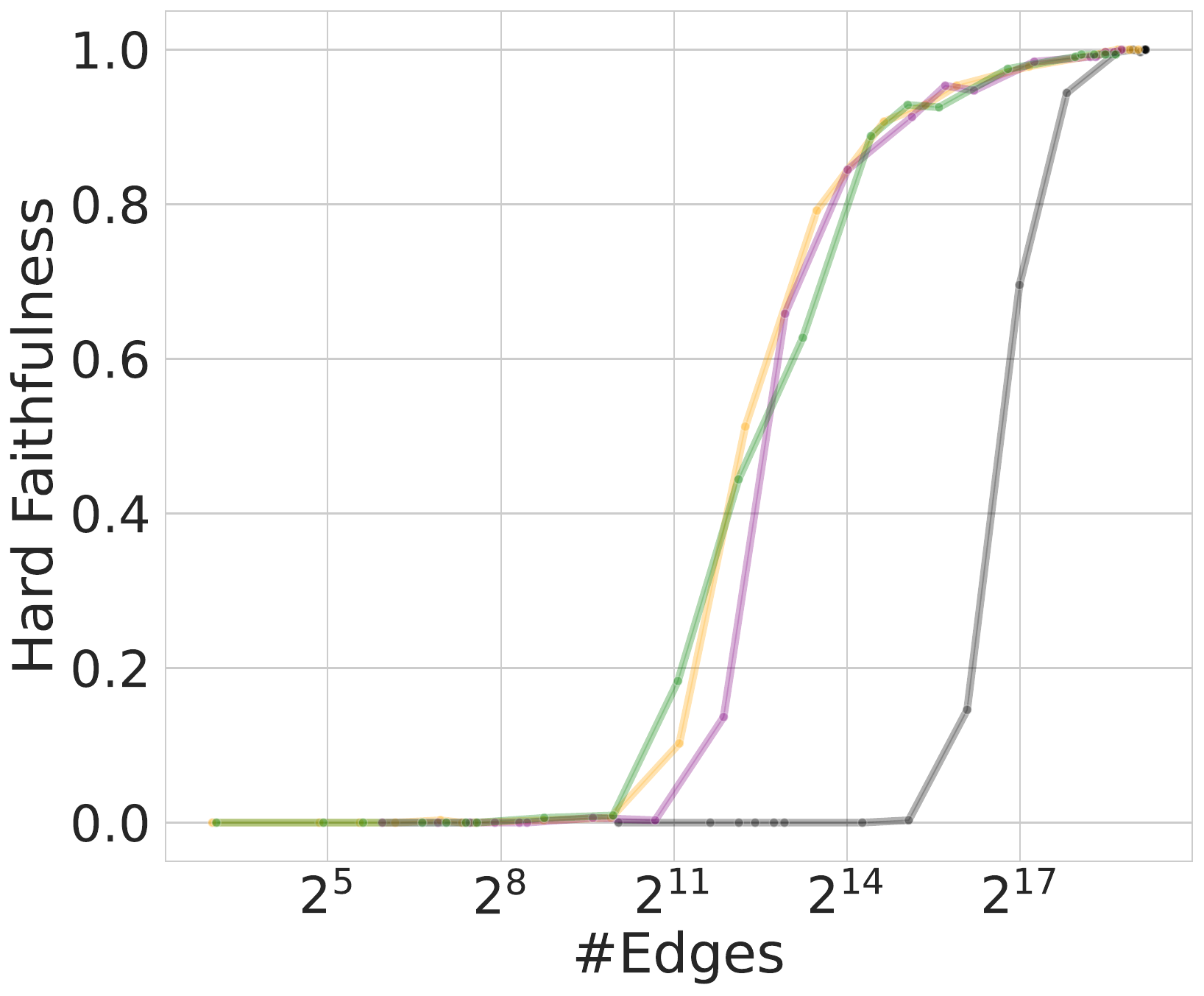} \hfill

    \vspace{0.05cm}
    \includegraphics[width=0.32\linewidth]{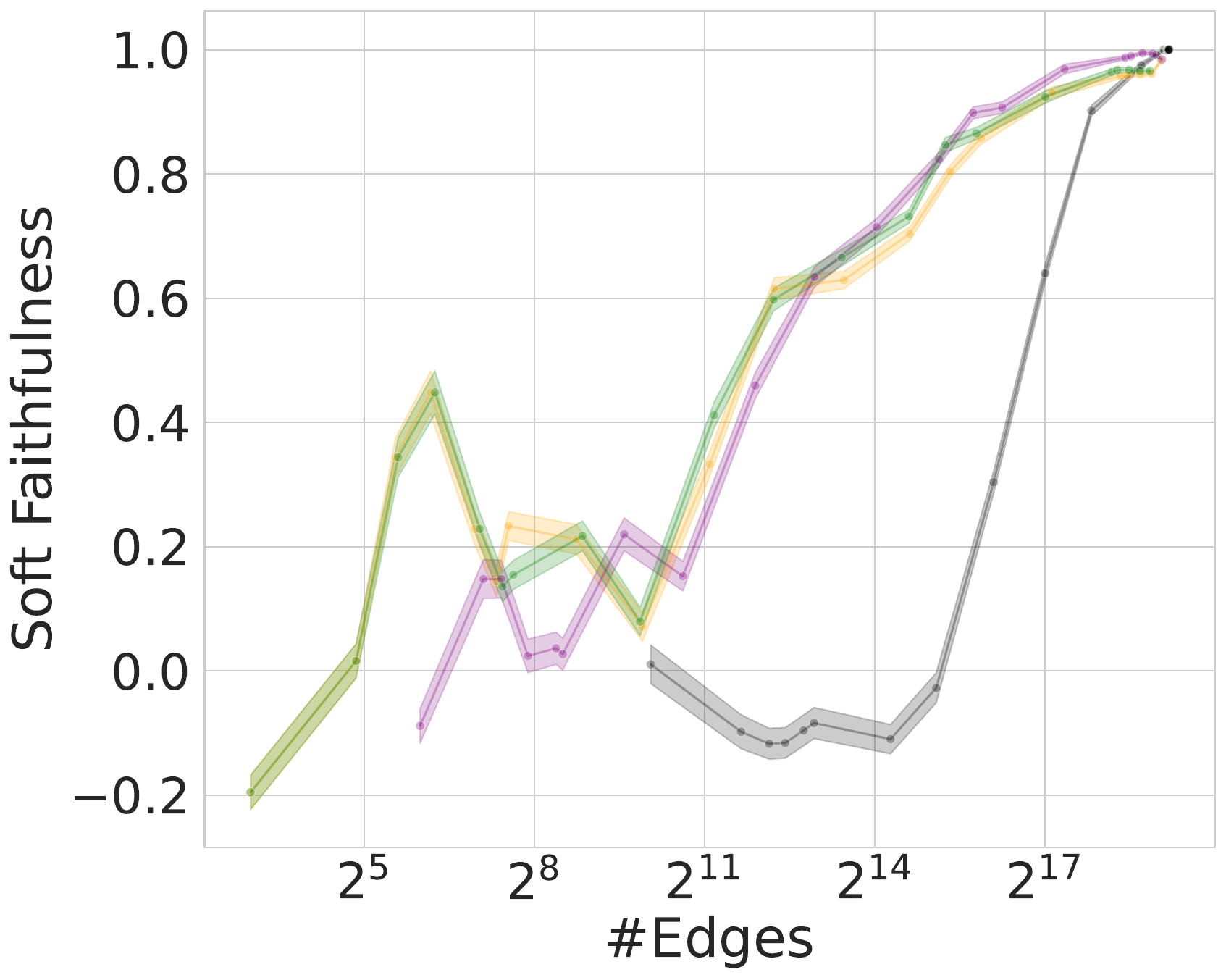} \hfill 
    \includegraphics[width=0.32\linewidth]{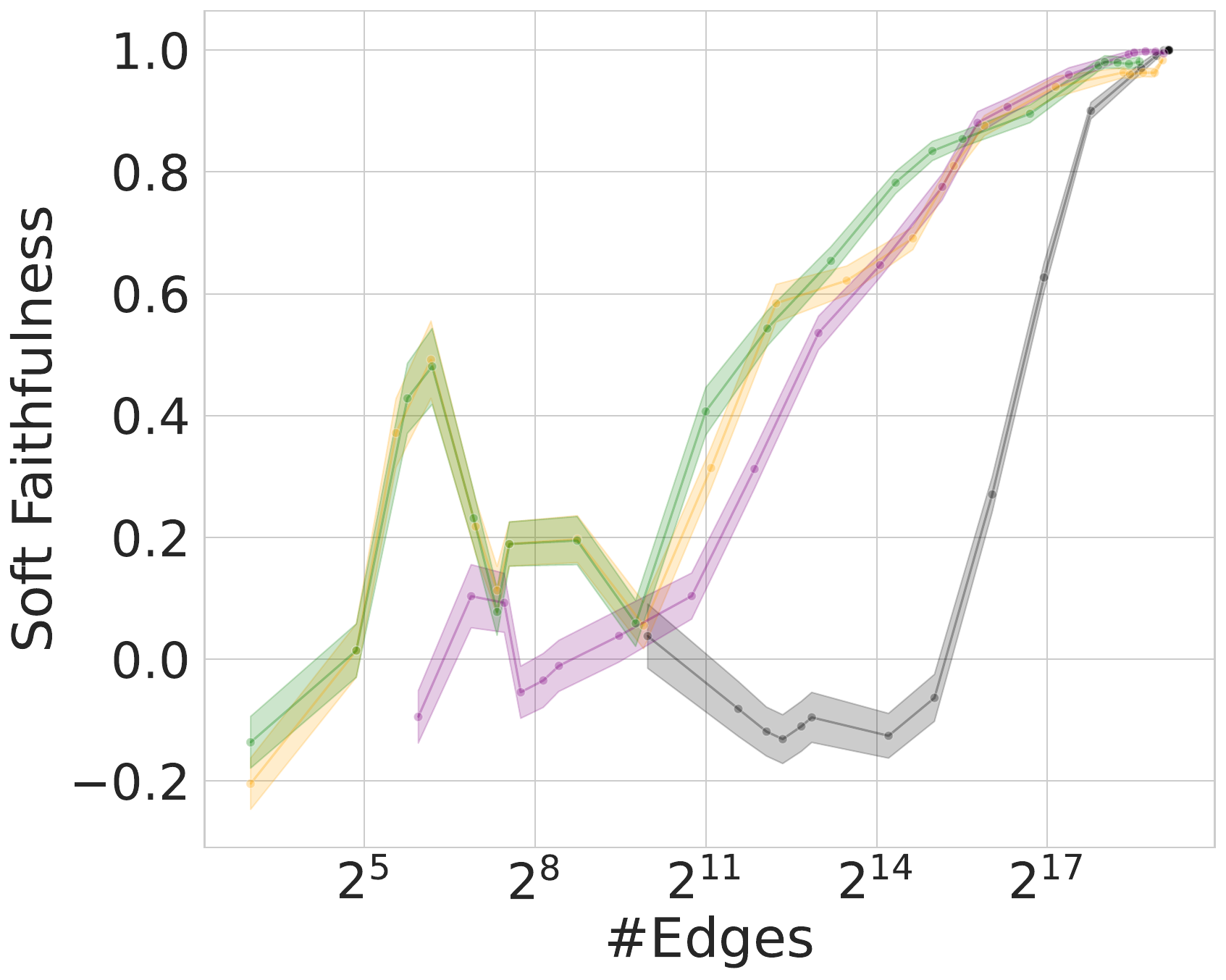} \hfill 
    \includegraphics[width=0.32\linewidth]{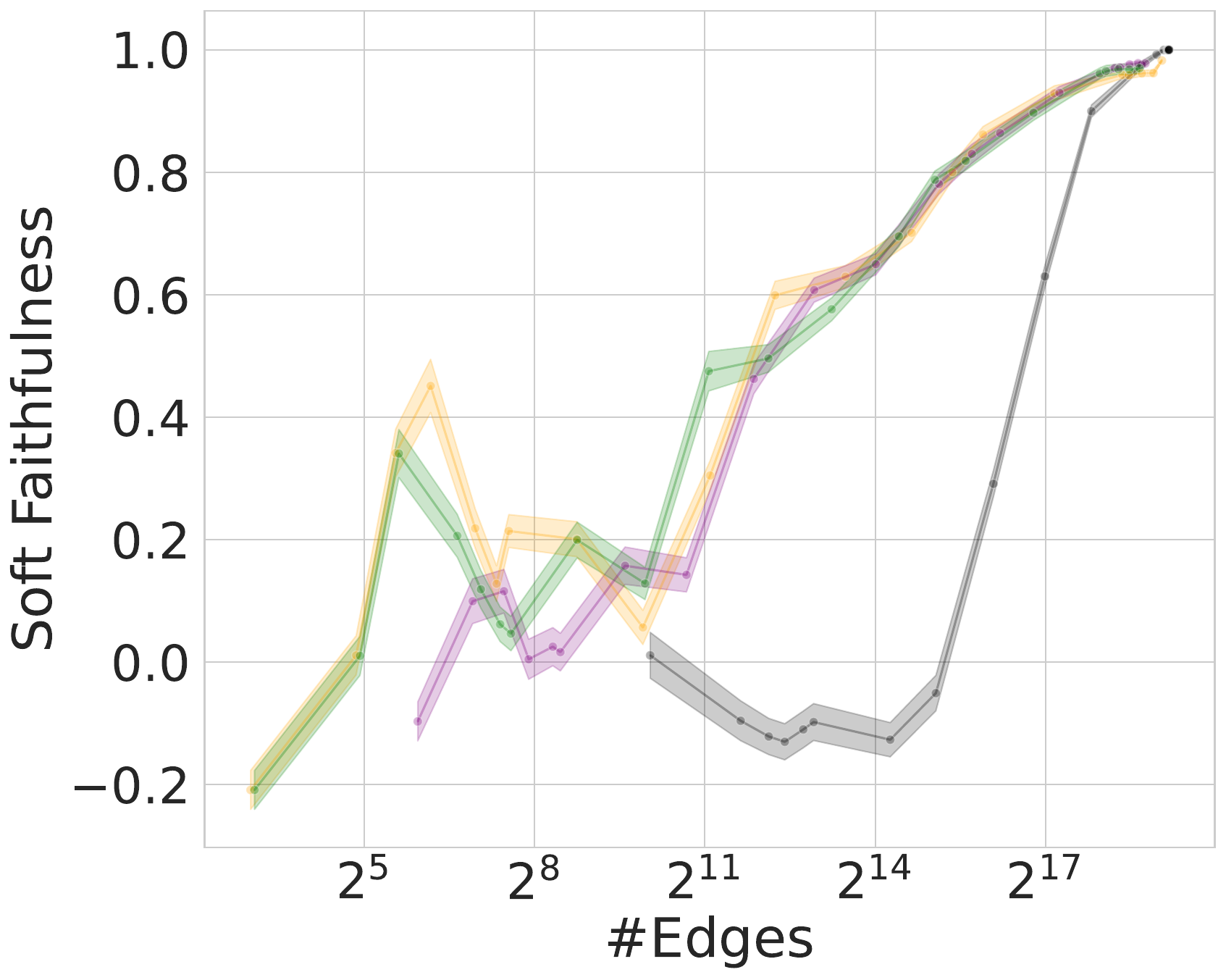} \hfill

     \textbf{IOI-BABA GPT2-small}

    \includegraphics[width=0.32\linewidth]{graphs/hard_faithfulness/appendix/ioi_ABBA_gpt2_1_207_accuracy.pdf} \hfill 
    \includegraphics[width=0.32\linewidth]{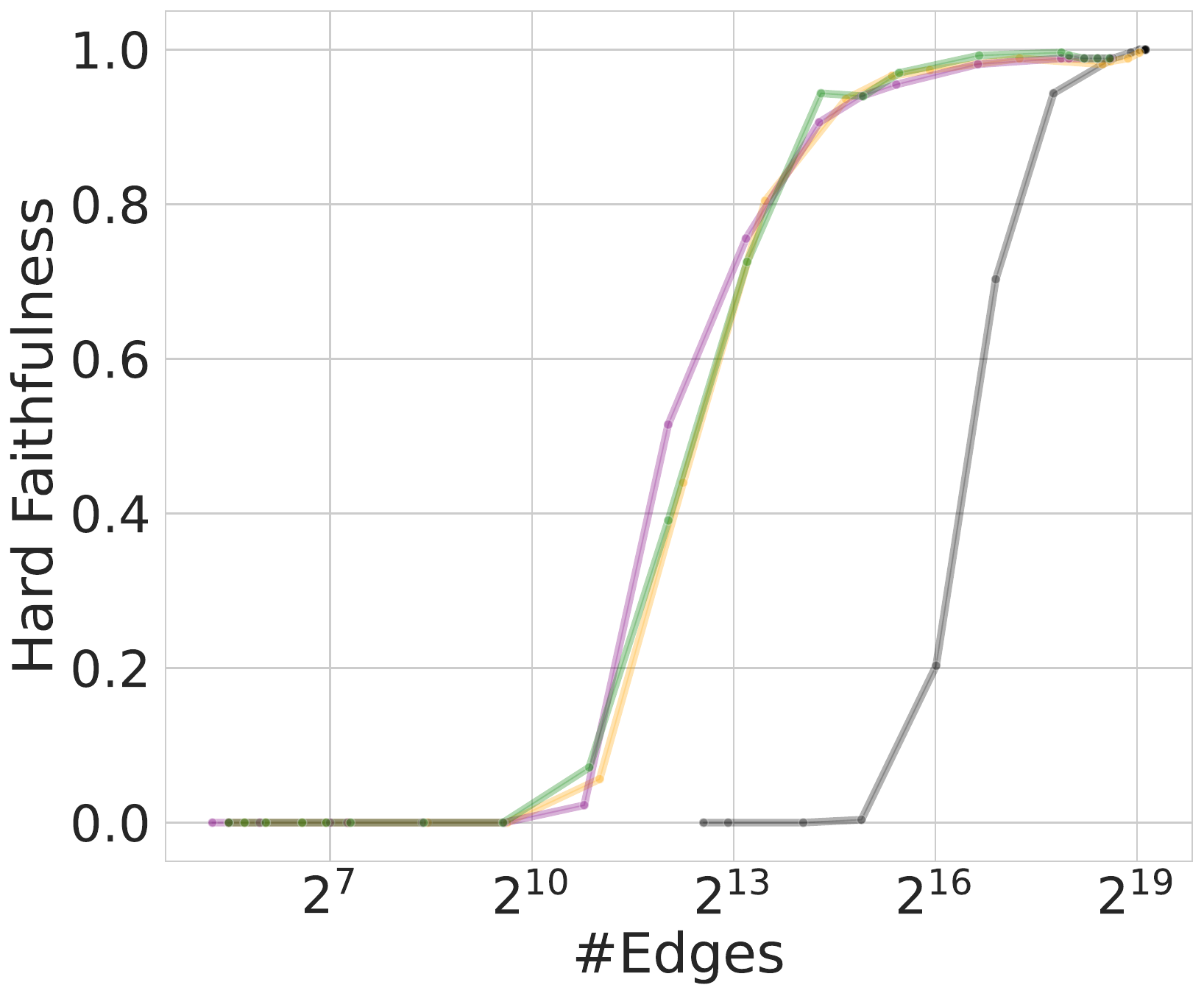} \hfill 
    \includegraphics[width=0.32\linewidth]{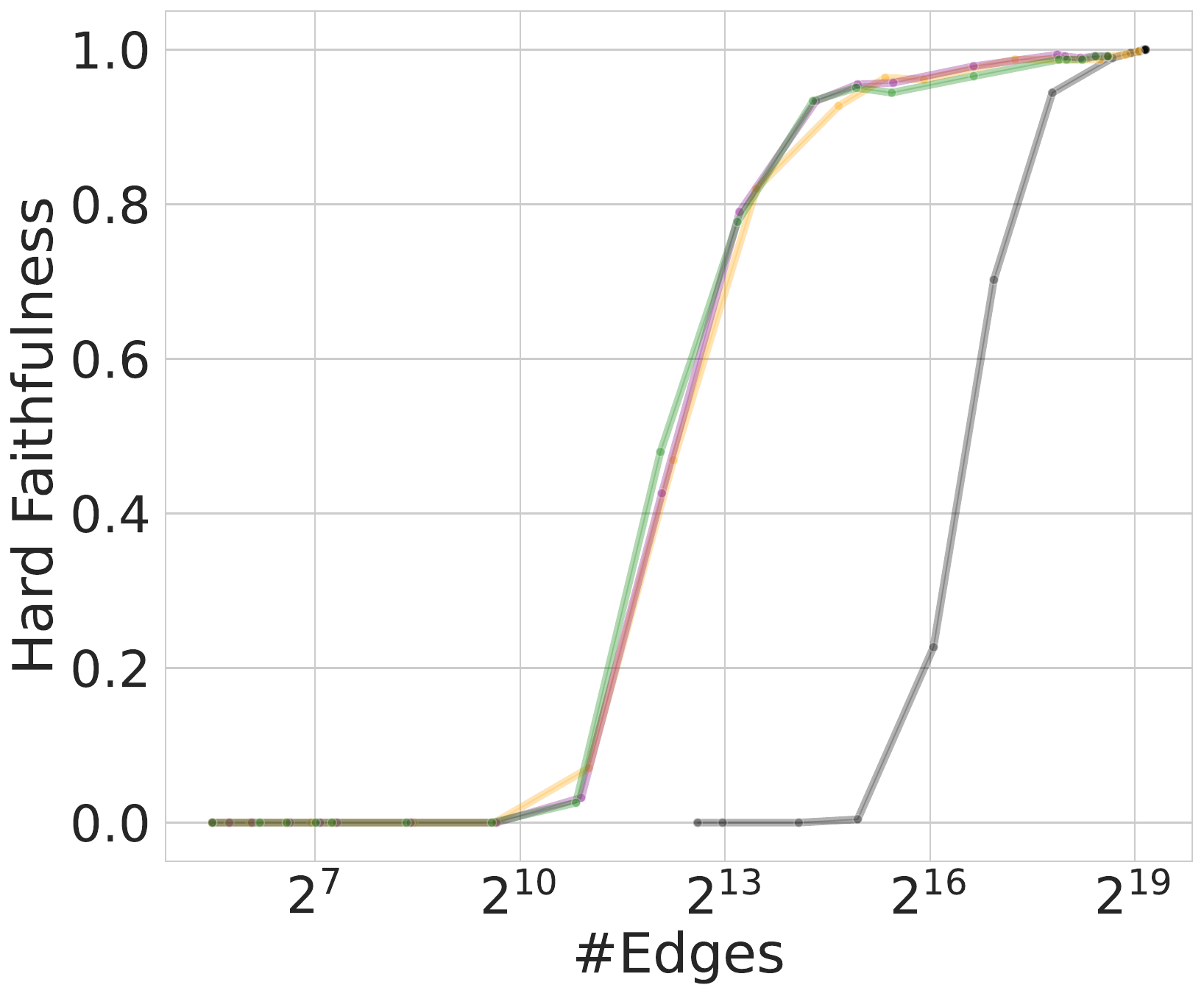} \hfill

    \vspace{0.05cm}
    \includegraphics[width=0.32\linewidth]{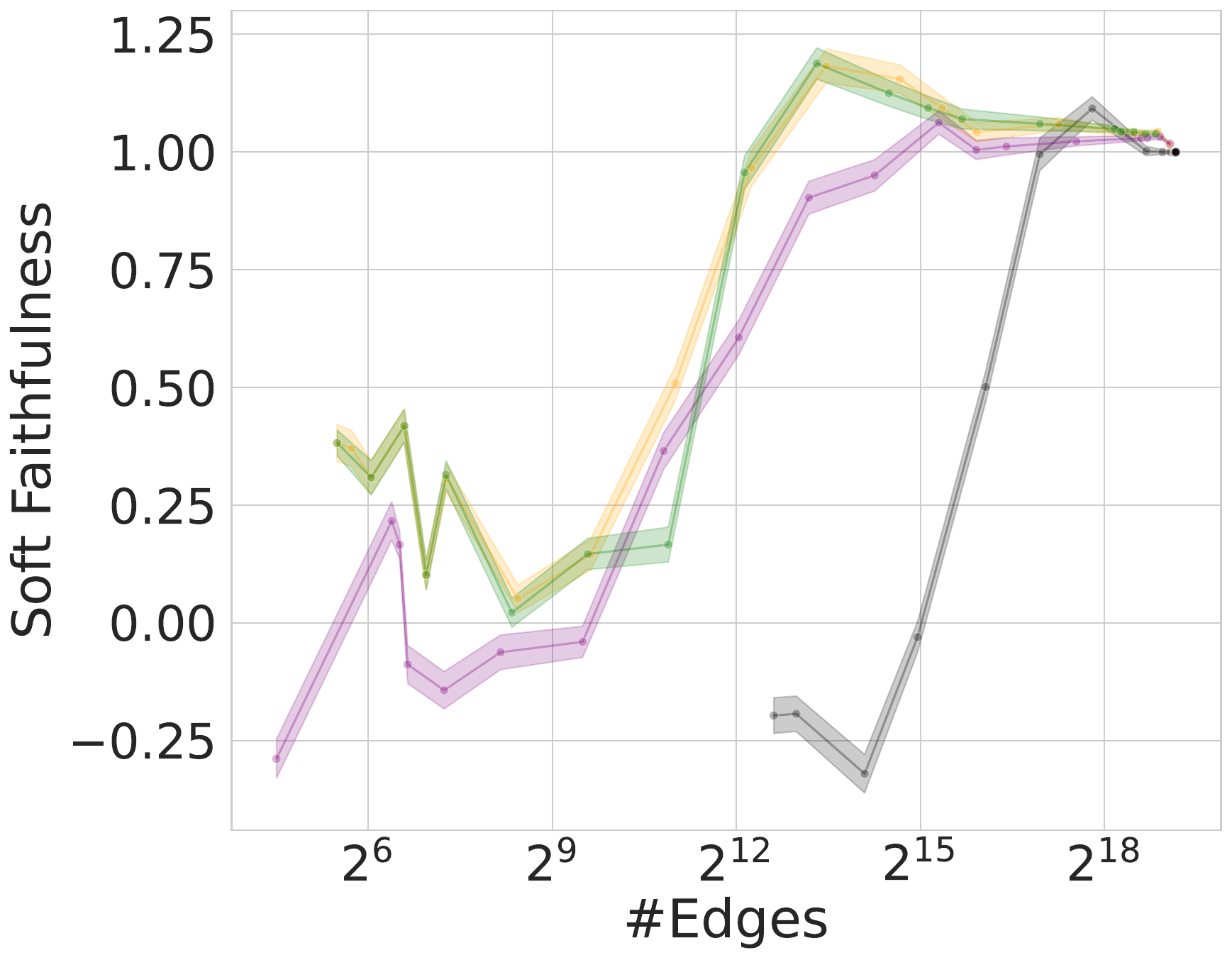} \hfill 
    \includegraphics[width=0.32\linewidth]{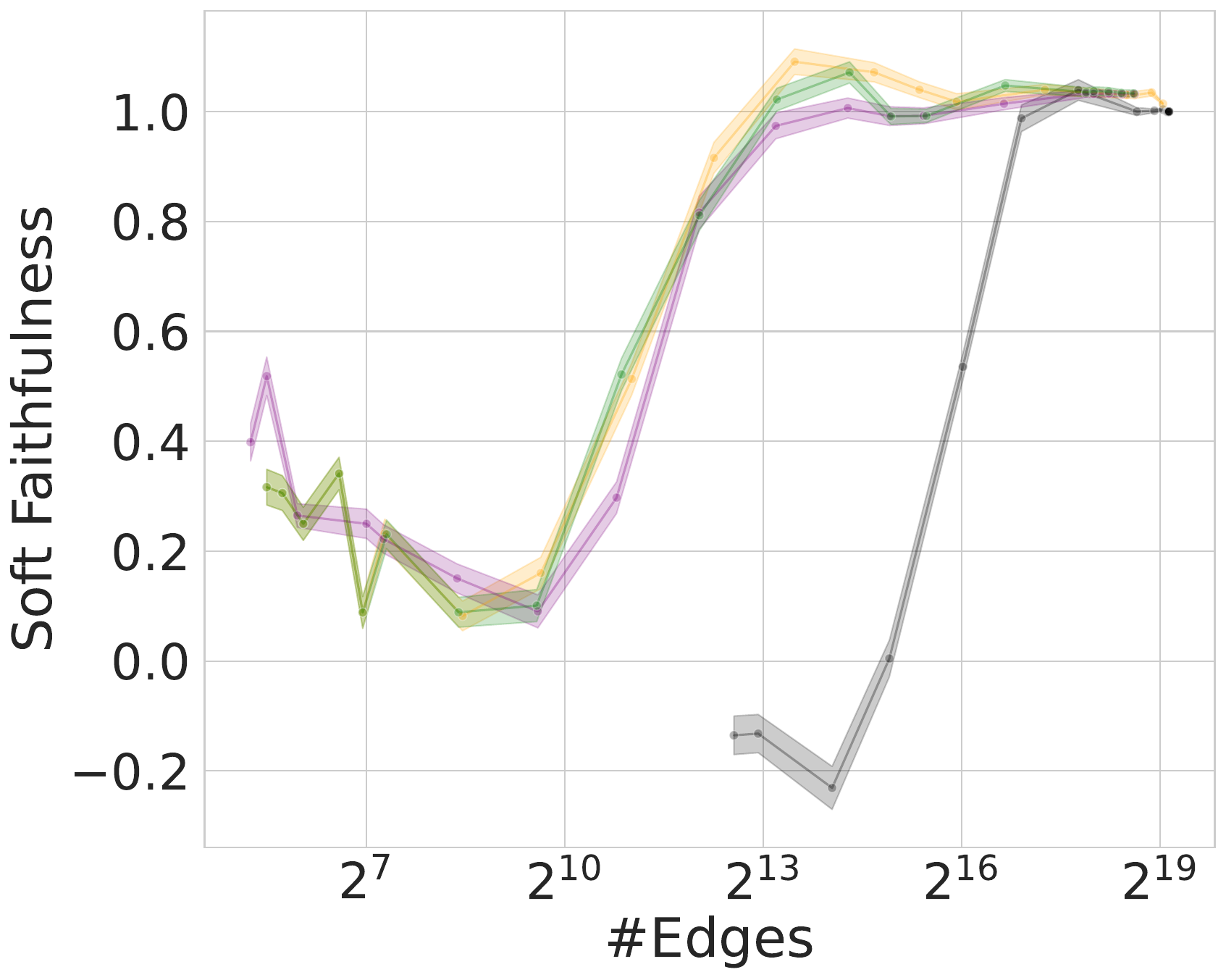} \hfill 
    \includegraphics[width=0.32\linewidth]{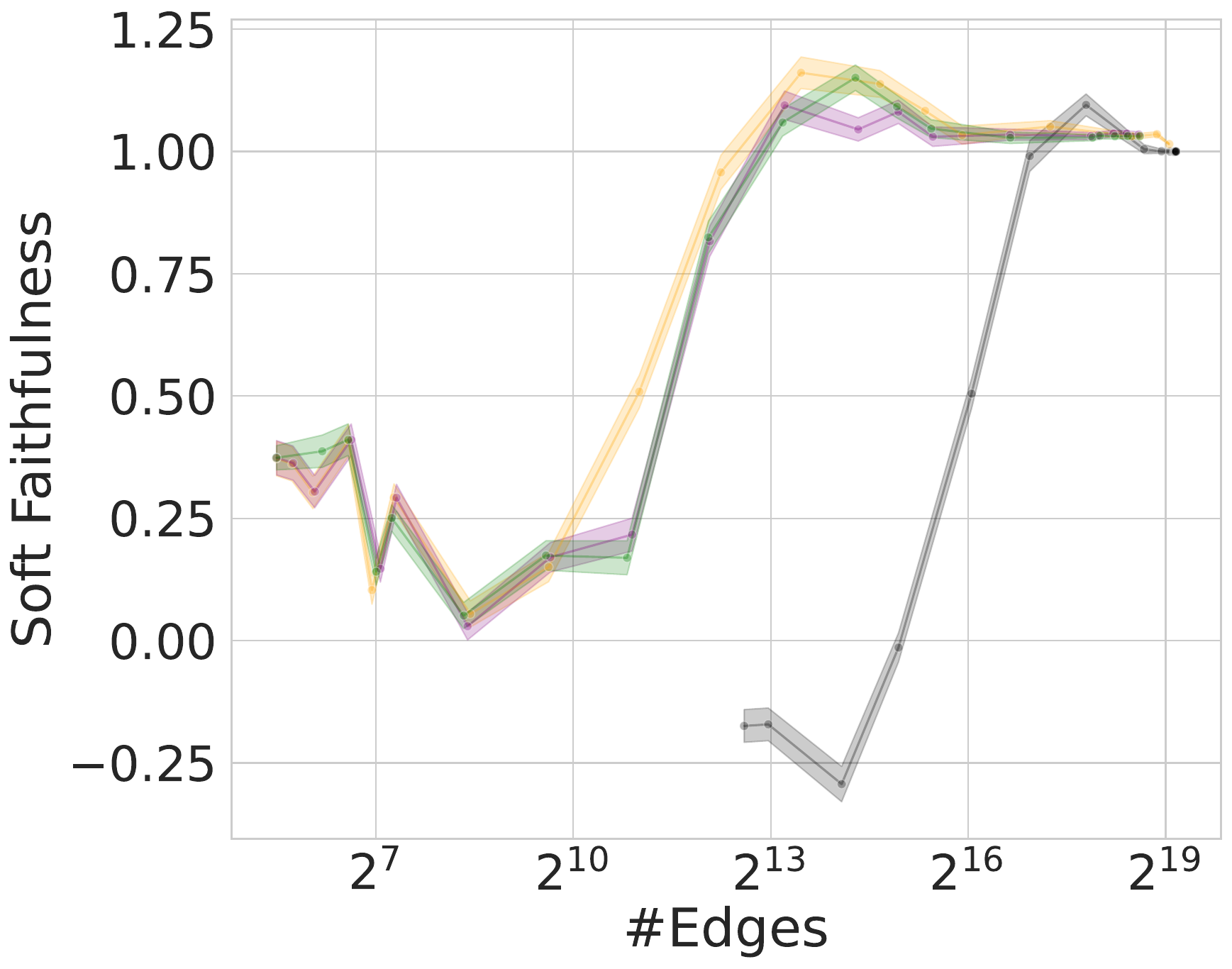} \hfill 
    \vspace{0.05cm}
    \includegraphics[width=0.6\linewidth]{graphs/legend_horizontal.pdf} \hfill
\caption{Each column shows results for a single trial.}
\label{fig:faithfulness_all_ioi_gpt2}
\end{figure*}

\begin{figure*}
    \centering
    \textbf{IOI-ABBA Llama-3-8b}

    \includegraphics[width=0.32\linewidth]{graphs/hard_faithfulness/main/ioi_ABBA_llama_3_454_accuracy.pdf} \hfill 
    \includegraphics[width=0.32\linewidth]{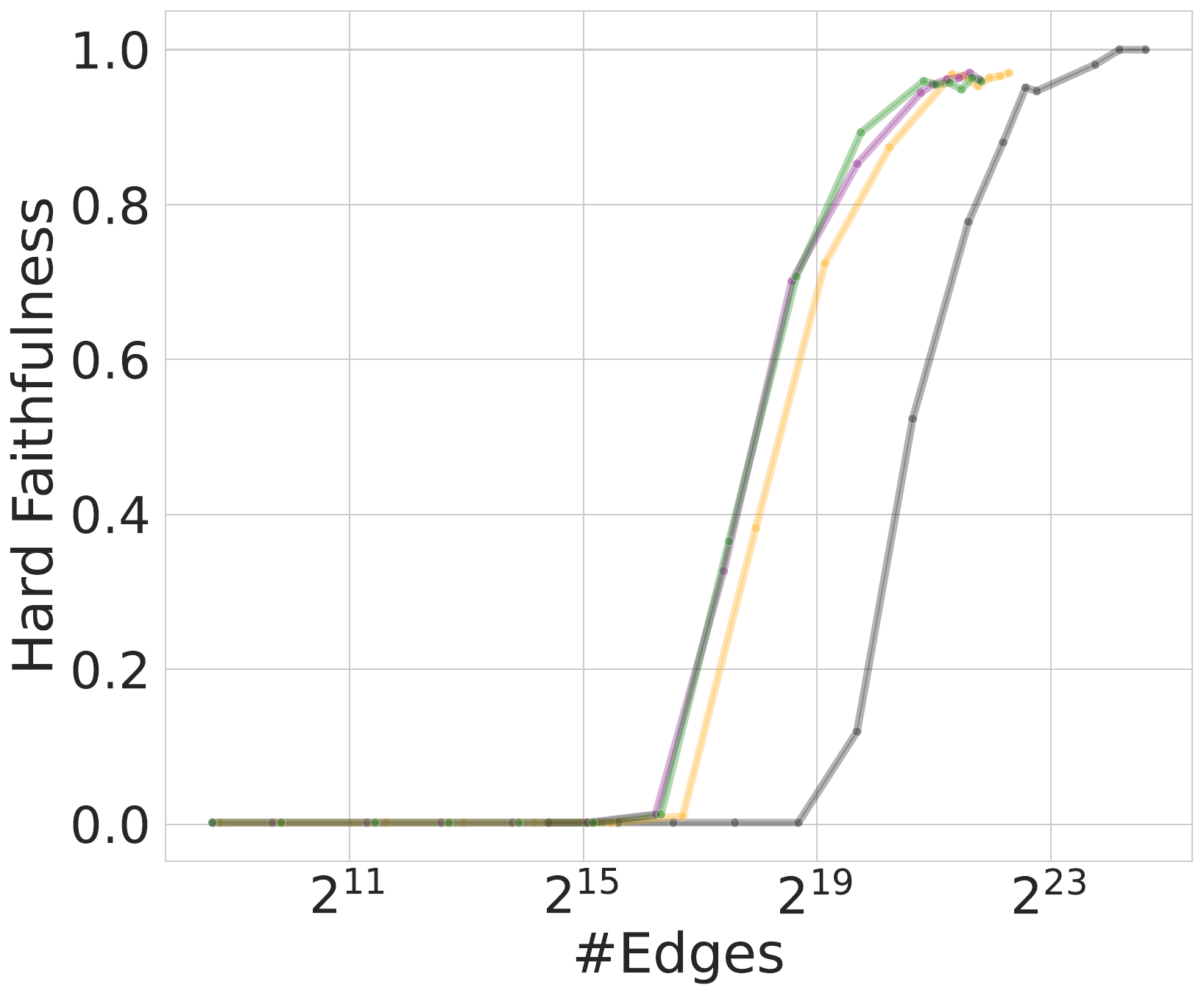} \hfill 
    \includegraphics[width=0.32\linewidth]{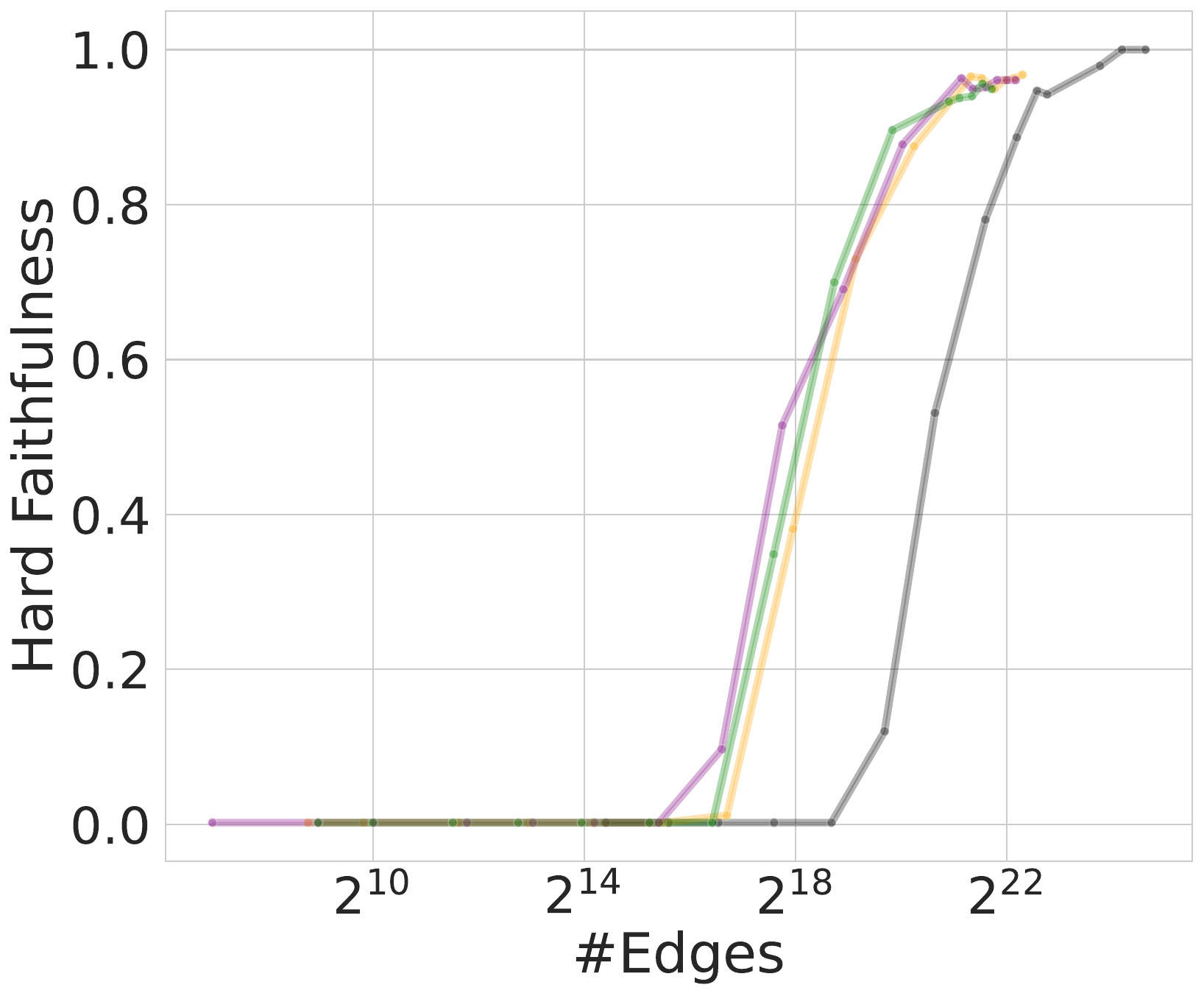} \hfill

    \vspace{0.05cm}
    \includegraphics[width=0.32\linewidth]{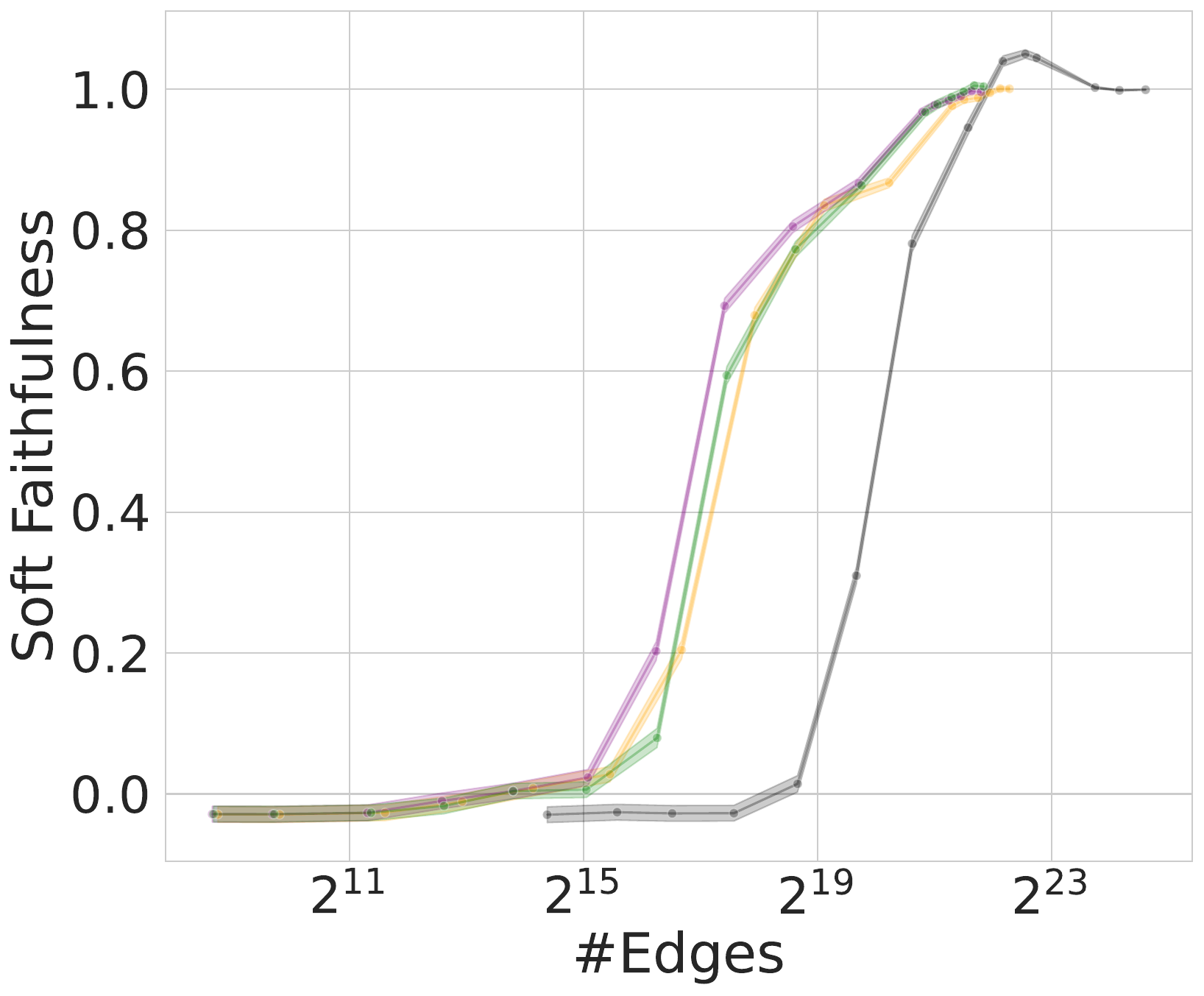} \hfill 
    \includegraphics[width=0.32\linewidth]{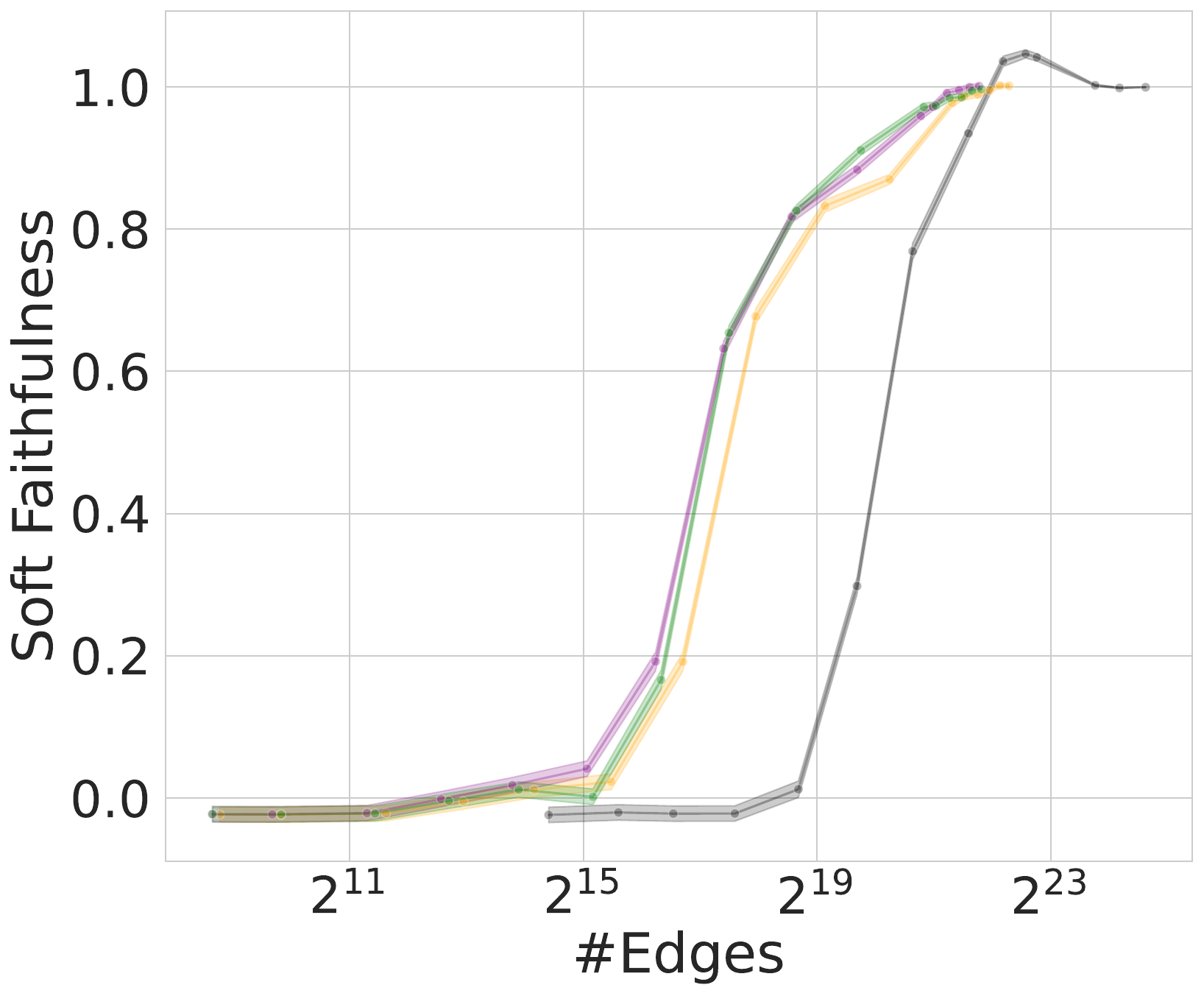} \hfill 
    \includegraphics[width=0.32\linewidth]{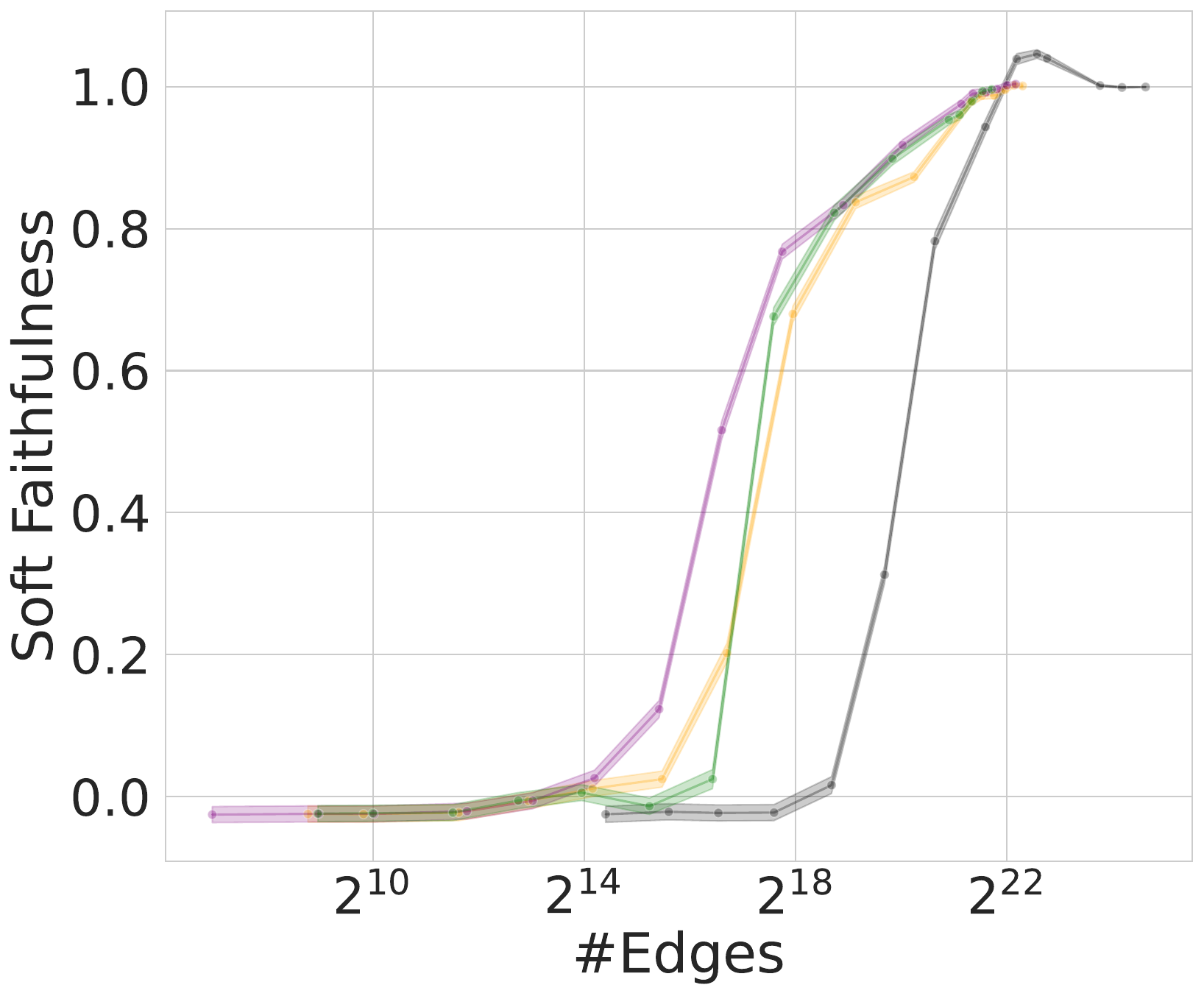} \hfill

     \textbf{IOI-BABA Llama-3-8b}

    \includegraphics[width=0.32\linewidth]{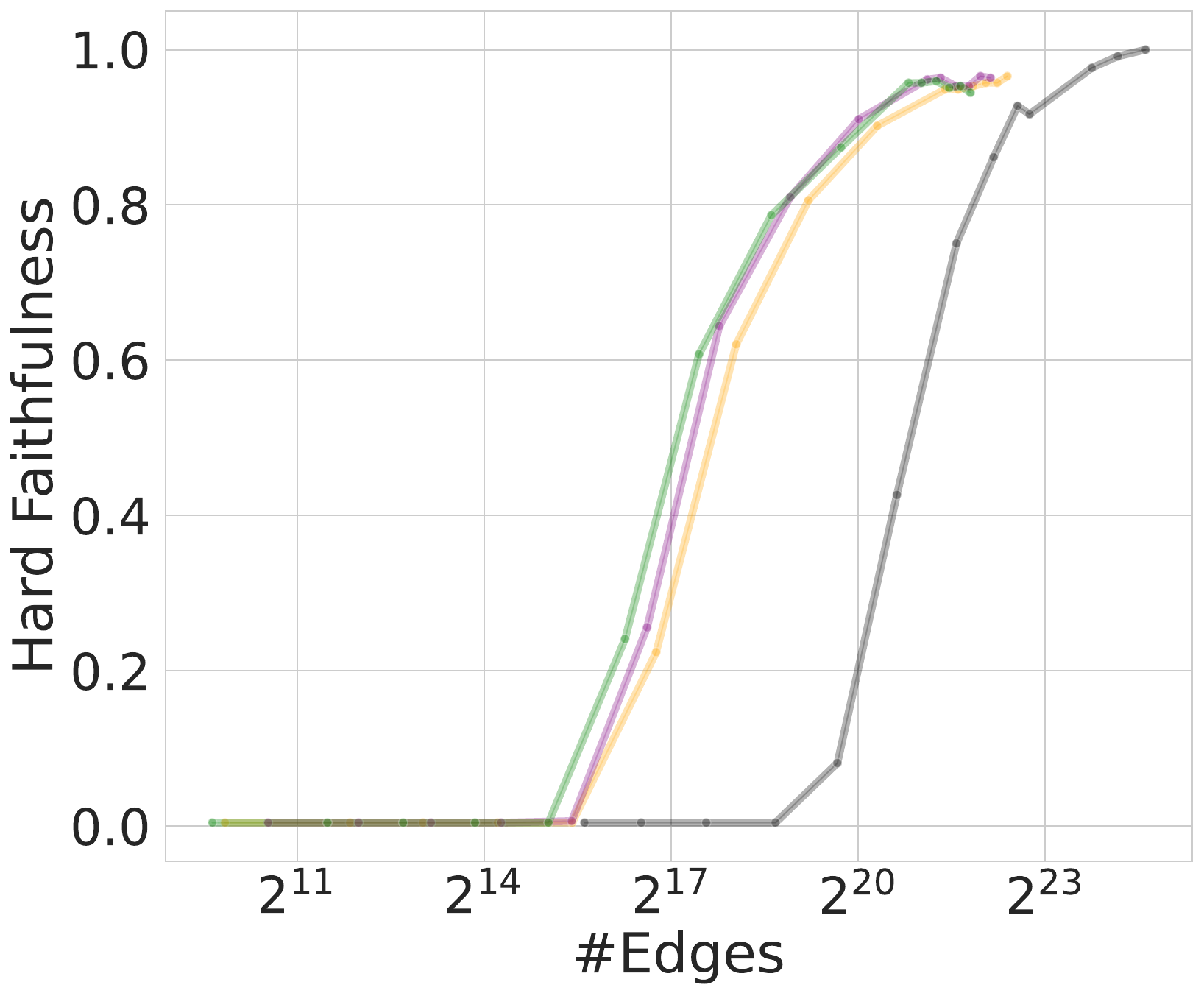} \hfill 
    \includegraphics[width=0.32\linewidth]{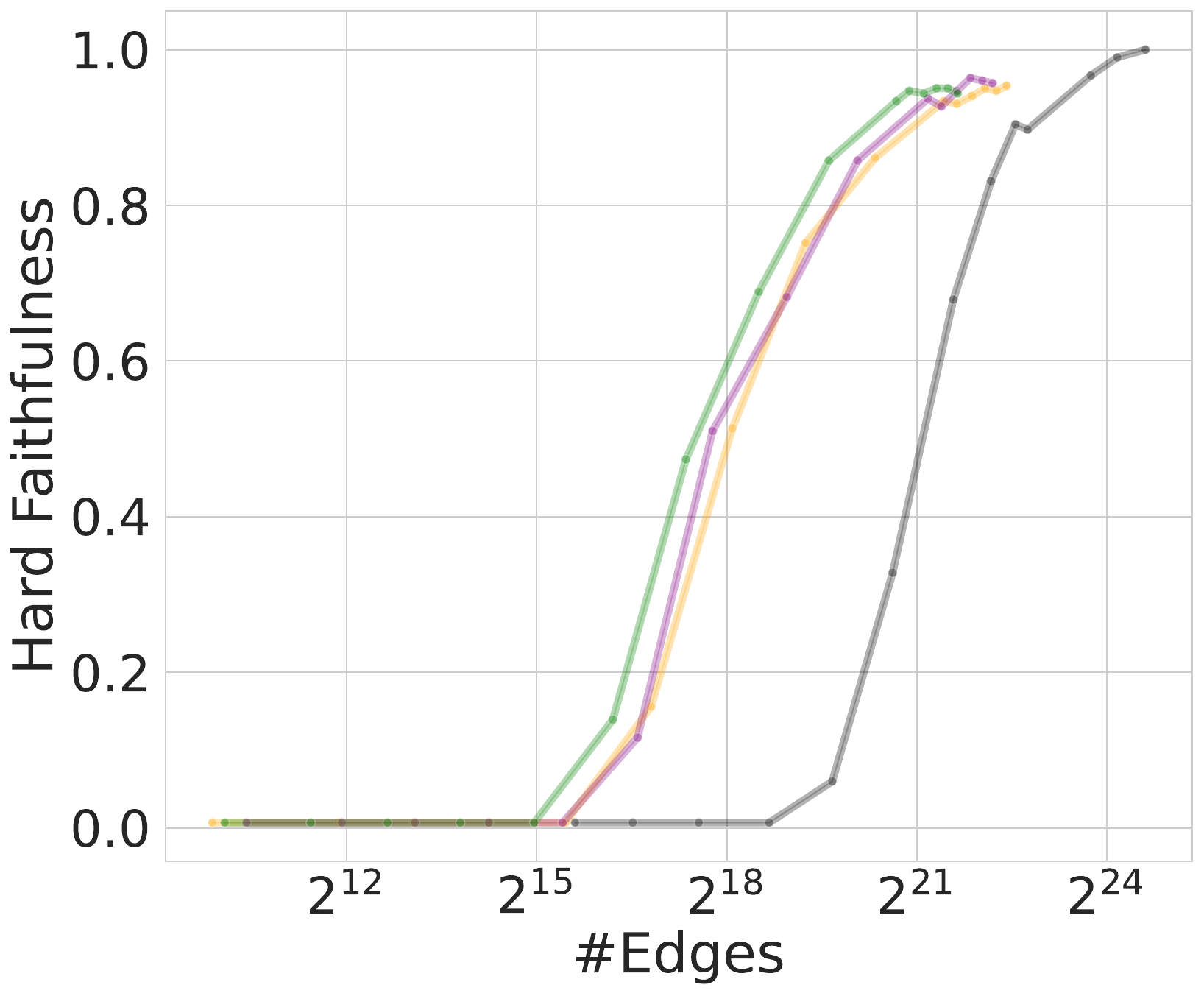} \hfill 
    \includegraphics[width=0.32\linewidth]{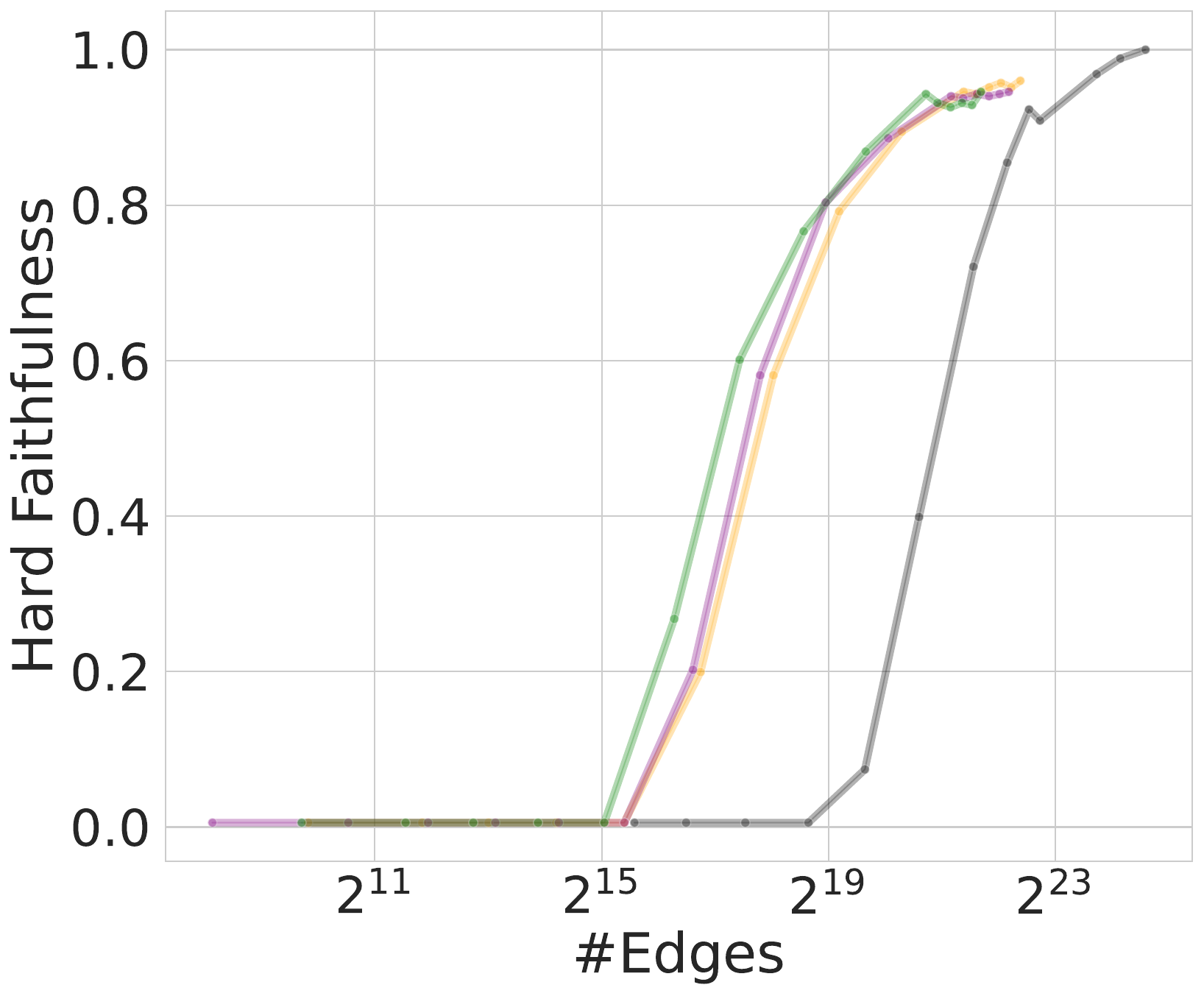} \hfill

    \vspace{0.05cm}
    \includegraphics[width=0.32\linewidth]{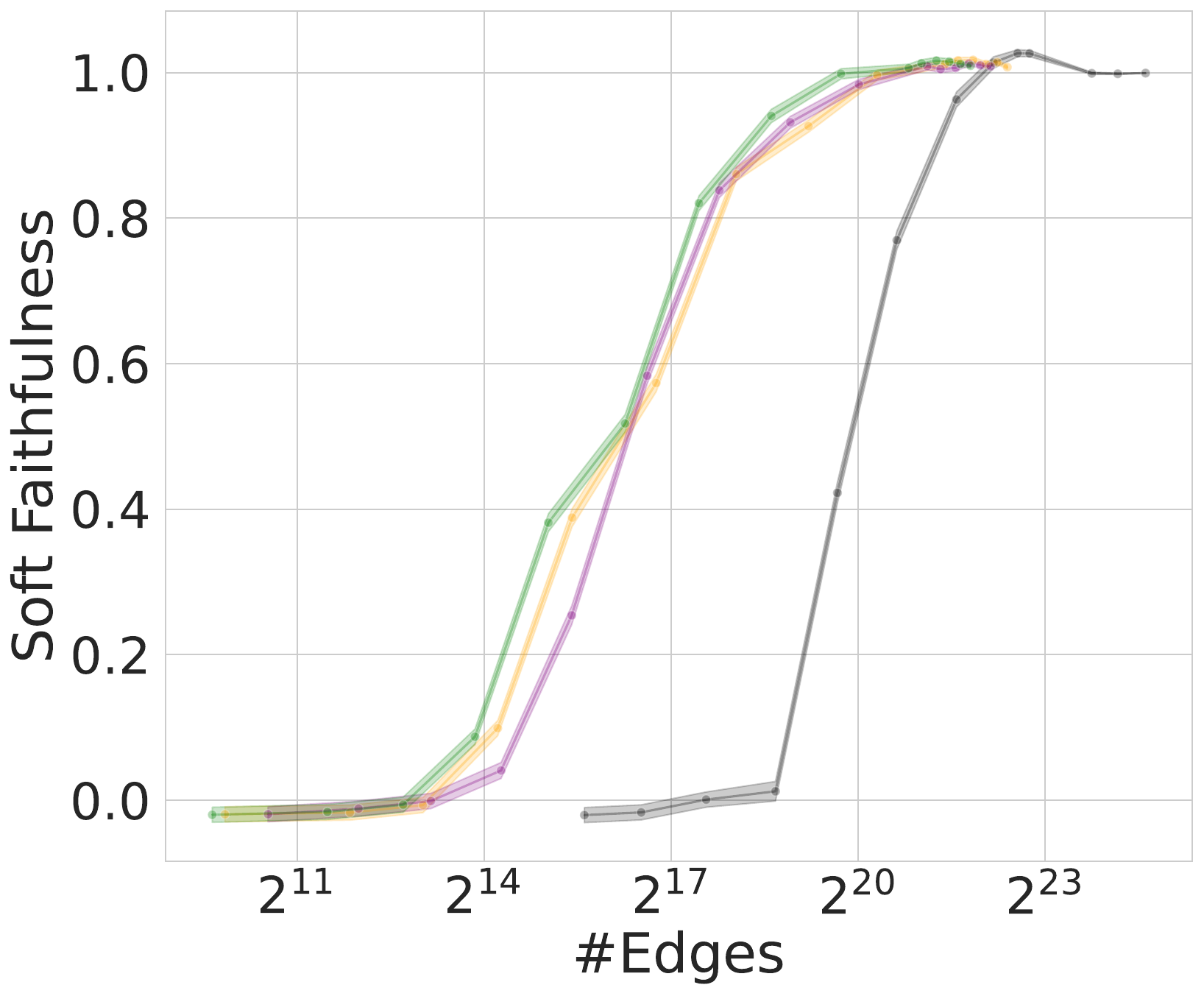} \hfill 
    \includegraphics[width=0.32\linewidth]{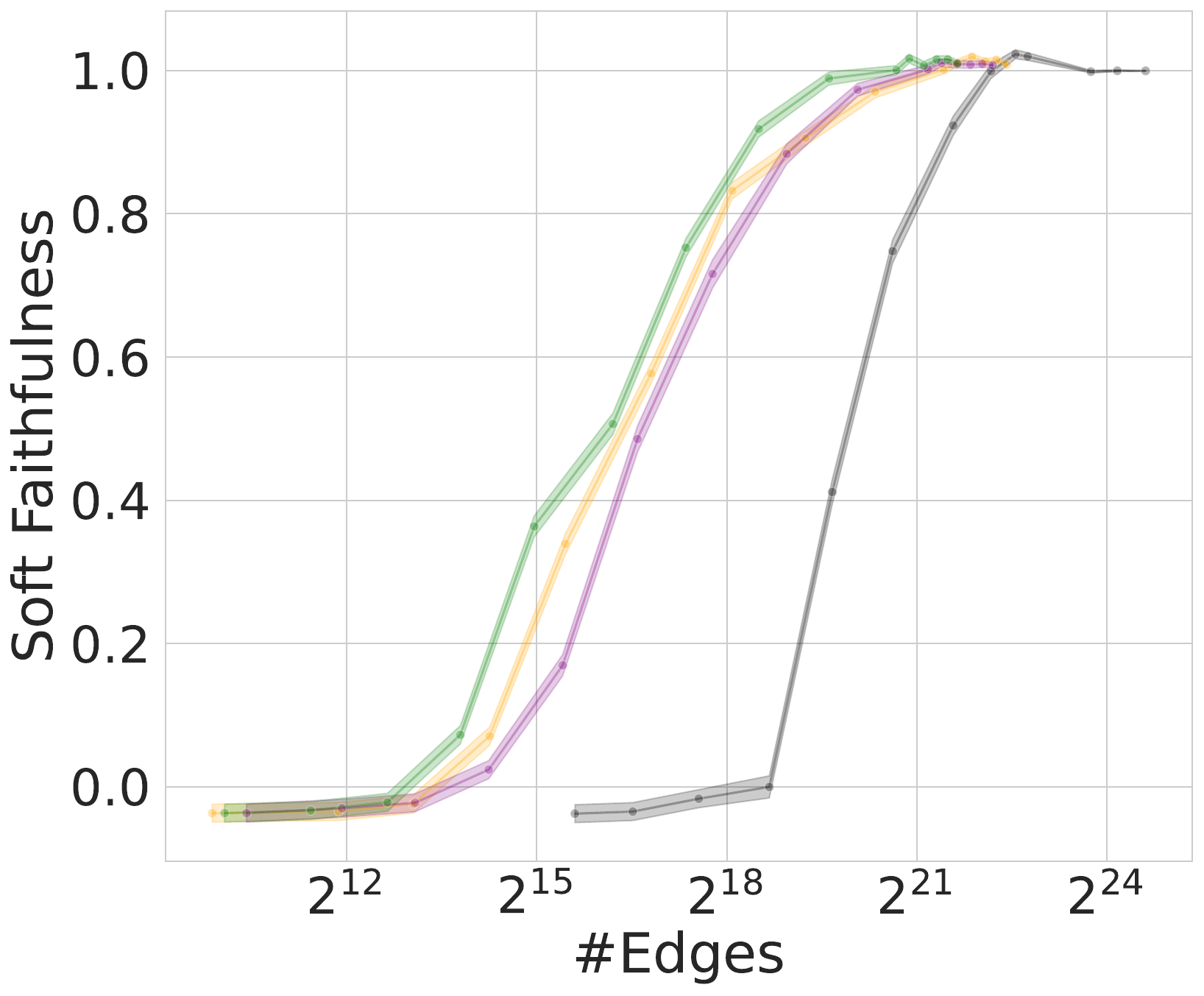} \hfill 
    \includegraphics[width=0.32\linewidth]{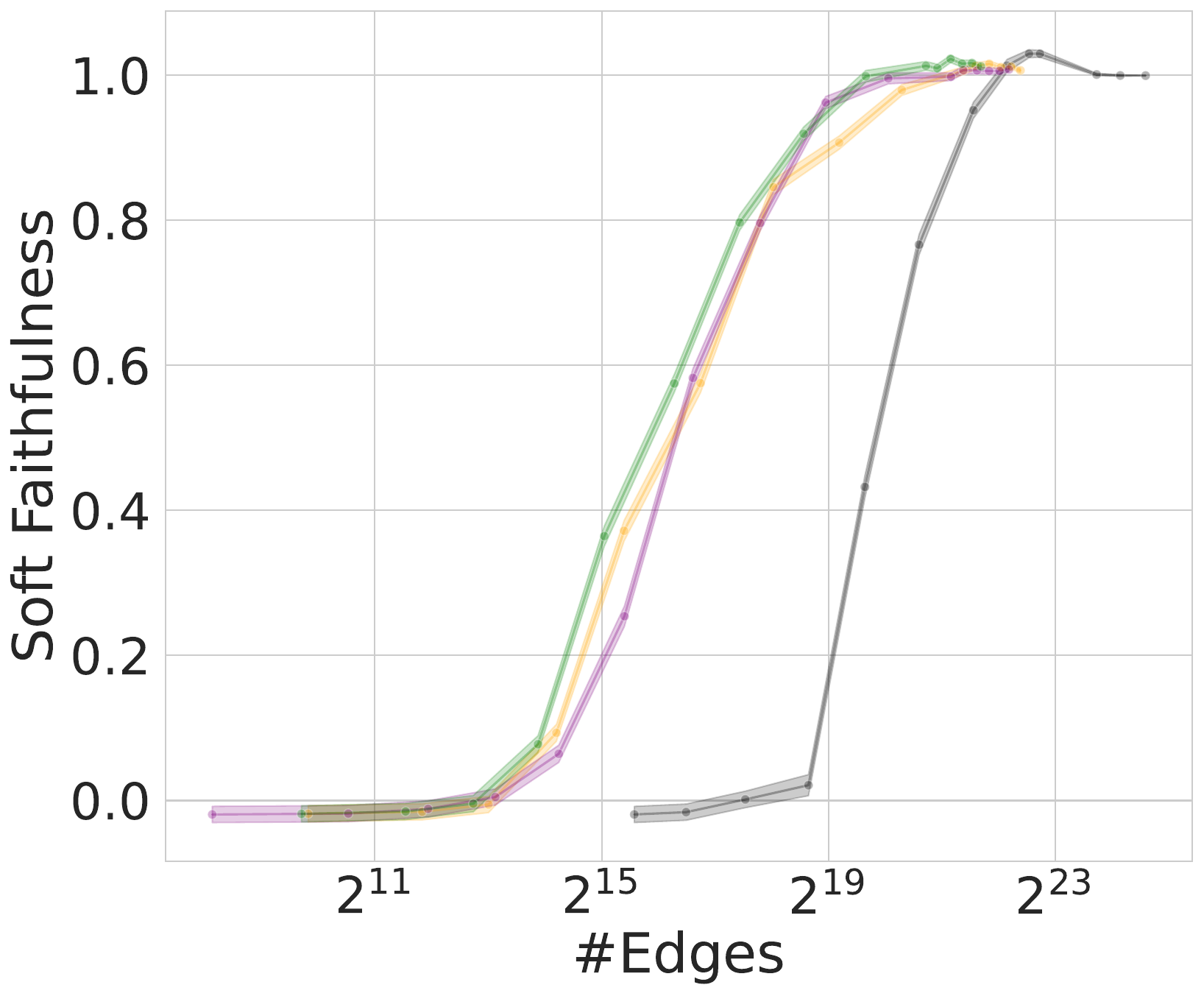} \hfill 
    \vspace{0.05cm}
    \includegraphics[width=0.6\linewidth]{graphs/legend_horizontal.pdf} \hfill
\caption{Each column shows results for a single trial.}
\label{fig:faithfulness_all_ioi_llama}
\end{figure*}

In Figures~\ref{fig:faithfulness_all_ioi_gpt2} and \ref{fig:faithfulness_all_ioi_llama}, we show faithfulness curves for GPT2-small and Llama-3-8B, respectively. When viewing hard faithfulness, results do not differ significantly across templates, nor across trials for GPT2-small. The difference between LLM, LLM+Mask, and the human schema is smaller for the third trial for the ABBA template. It is also low for the second two trials for the BABA template. When viewing soft faithfulness, similar trends are present, but the schema-based approaches generally perform similarly to each other (with human and LLM+mask's margin from LLM being much smaller).

The difference between schemas is smaller for Llama-3-8B. While each schema-based method outperforms non-positional circuits, there does not appear to be a significant difference between LLM, LLM+Mask, and human schema.

\end{document}